\documentclass[lettersize,journal]{IEEEtran}
\usepackage{amsmath,amsfonts}
\usepackage{algorithmic}
\usepackage{algorithm}
\usepackage{array}
\usepackage[caption=false,font=scriptsize,labelfont=rm,textfont=rm]{subfig}
\usepackage{textcomp}
\usepackage{stfloats}
\usepackage{url}
\usepackage{verbatim}
\usepackage{graphicx}
\usepackage{cite}
\usepackage{makecell}
\usepackage[mathscr]{eucal}
\usepackage{booktabs}
\usepackage{float}   
\usepackage{subfloat}
\usepackage{ragged2e} 
\usepackage{bbm}
\usepackage{bm}
\usepackage{xcolor}
\usepackage{hyperref}
\hypersetup{hypertex=true,
	colorlinks=true,
	linkcolor=blue,
	anchorcolor=blue,
	citecolor=blue}
\usepackage{orcidlink}
\usepackage{tikz}
\usepackage{mathrsfs}
\usepackage{amssymb}
\usepackage{dsfont}
\usepackage{multirow}
\usepackage{placeins}

\begin{document}

\title{Unified Domain Adaptive Semantic Segmentation}

\author{Zhe Zhang\hspace{-1.0mm}$^{~\orcidlink{0009-0000-0298-5497}}$, Gaochang Wu\hspace{-1.0mm}$^{~\orcidlink{0000-0002-5149-2995}}$,~\IEEEmembership{Member,~IEEE}, Jing Zhang\hspace{-1.0mm}$^{~\orcidlink{0000-0001-6595-7661}}$, Xiatian Zhu\hspace{-1.0mm}$^{~\orcidlink{0000-0002-9284-2955}}$, Dacheng~Tao\hspace{-1.0mm}$^{~\orcidlink{0000-0001-7225-5449}}$,~\IEEEmembership{Fellow,~IEEE,}\\
	Tianyou Chai\hspace{-1.0mm}$^{~\orcidlink{0000-0002-4623-1483}}$,~\IEEEmembership{Life Fellow,~IEEE}%

\thanks{
	\IEEEcompsocthanksitem Zhe Zhang, Gaochang Wu, and Tianyou Chai are with the State Key Laboratory of Synthetical Automation for Process Industries, Northeastern University, Shenyang, China. Email: zhangzhe17@stumail.neu.edu.cn, wugc@mail.neu.edu.cn, tychai@mail.neu.edu.cn.
    \IEEEcompsocthanksitem Jing Zhang is with the School of Computer Science, Wuhan University, China. E-mail: jingzhang.cv@gmail.com.
 \IEEEcompsocthanksitem Xiatian Zhu is with the Surrey Institute for People-Centred Artificial
Intelligence, and Centre for Vision, Speech and Signal Processing, University of Surrey, Guildford, UK. E-mail: Eddy.zhuxt@gmail.com.
    \IEEEcompsocthanksitem Dacheng Tao is with the College of Computing \& Data Science, Nanyang Technological University, Singapore. E-mail: dacheng.tao@gmail.com
	\IEEEcompsocthanksitem Corresponding author: Tianyou Chai, Gaochang Wu.
	}

}

\markboth{IEEE TRANSACTIONS ON PATTERN ANALYSIS AND MACHINE INTELLIGENCE, VOL. XX, NO. XX, XX 2025}%
{Zhang \MakeLowercase{\textit{et al.}}: Unified Domain Adaptive Semantic Segmentation}


\maketitle
\begin{abstract}
Unsupervised Domain Adaptive Semantic Segmentation (UDA-SS) aims to
transfer the supervision from a labeled source domain to an unlabeled and shifted target domain. The majority of existing UDA-SS works typically consider images whilst recent attempts have extended further to tackle videos by modeling the temporal dimension. Although two lines of research share the major challenges -- overcoming the underlying domain distribution shift, their studies are largely independent. It causes several issues: (1) The insights gained from each line of research remain fragmented, leading to a lack of holistic understanding of the problem and potential solutions. (2) Preventing the unification of methods and best practices across two scenarios (images and videos) will lead to redundant efforts and missed opportunities for cross-pollination of ideas. (3) Without a unified approach, the knowledge and advancements made in one scenario may not be effectively transferred to the other, leading to suboptimal performance and slower progress. Under this observation, we advocate unifying the study of UDA-SS across video and image scenarios, enabling a more comprehensive understanding, synergistic advancements, and efficient knowledge sharing. To that end, we explore the unified UDA-SS from a general domain augmentation perspective, serving as a unifying framework, enabling improved generalization, and potential for cross-pollination, ultimately contributing to the practical impact and overall progress. Specifically, we propose a Quad-directional Mixup (QuadMix) method, characterized by tackling intra-domain discontinuity, fragmented gap bridging, and feature inconsistencies through four-directional paths designed for intra- and inter-domain mixing within an explicit feature space. To deal with temporal shifts within videos, we incorporate optical flow-guided feature aggregation across spatial and temporal dimensions for fine-grained domain alignment, which is extendable to image scenarios. Extensive experiments show that QuadMix outperforms the state-of-the-art works by large margins on four challenging UDA-SS benchmarks. Our source code and models will be released at \url{https://github.com/ZHE-SAPI/UDASS}.
\end{abstract}

\begin{figure}[!t]
	\centering
	\includegraphics[width= 0.95\linewidth]{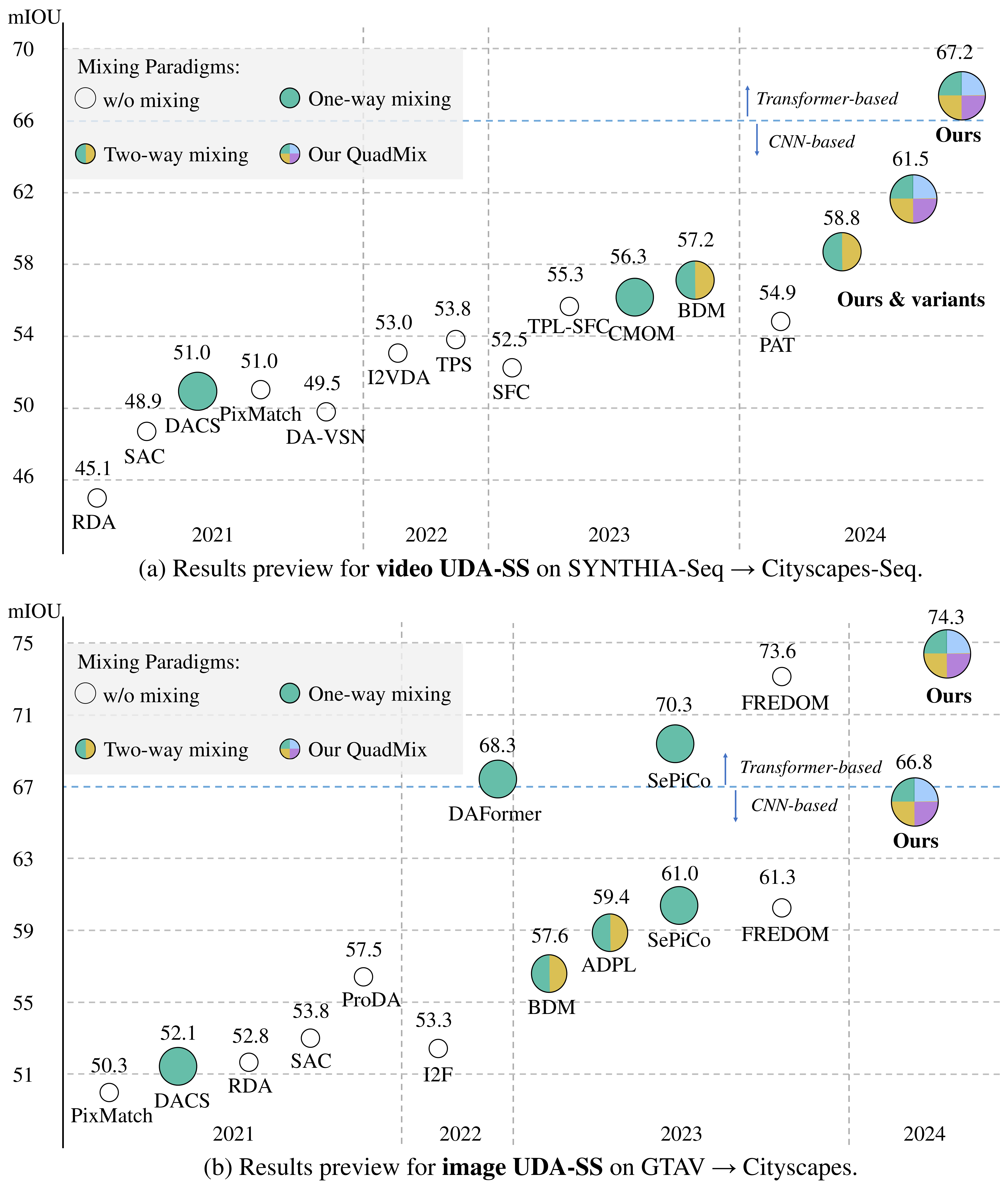} 
	\caption{
 Instead of studying UDA-SS for images and videos independently, we explore the unified study with a single approach.
 }
	\label{fig. 1}
\end{figure}
\section{Introduction}

\IEEEPARstart{U}{nsupervised} Domain Adaptive Semantic Segmentation (UDA-SS) focuses on transferring pixel-level dense knowledge from a labeled source domain to an unlabeled target domain with distribution shift \cite{daformer,cmom,dsp,unimix}. 
The key challenge is all about how to tackle intrinsic distributional gaps between the two domains without target domain supervision.
As typical, the literature is dominated by \texttt{image} UDA-SS with the aim to reduce the domain discrepancy through image alignment~\cite{bdm,cycada,intro5}, metric minimization~\cite{intro5, intro2-13, intro2-14}, self-supervised learning with pseudo-label~\cite{sac,iast,proda}, or regularization~\cite{crst, related-50.5}. 
This inspires the progress of \texttt{video} UDA-SS by additionally handling the temporal dimension via video alignment~\cite{vat,davsn}, geometric-based cues \cite{tps,moda} for domain-invariant learning.

While the two UDA-SS tasks share the congenetic challenge, i.e., feature distributional gap across domains, their investigation is clearly separated. This could cause limitations like fragmented insights over tasks, lack of holistic understanding of the challenge, preventing the unification
of techniques, and finally slowing the development progress.

Under the aforementioned insights, in this work, we advocate the unified investigation of image and video UDA-SS.  To have a more generic solution, we explore the novel domain augmentation strategy by introducing a novel method, \textbf{Quad-directional Mixup} (QuadMix), capable of generating a diversity of intermediate domains within both source and target domains and across them within an explicit feature space, facilitated by an adaptive feature aggregation module for fine-grained domain alignment. Specifically, we generate adaptive category-aware sample patch templates from either (source or target) domain, followed by fusing them within the domain to create generalized \texttt{intra-mixed} domains.
Concurrently, we further fuse those patches across the original domains interactively to construct more enhanced \texttt{inter-mixed} domains. 
In addition to common pixel-level mixing, feature-level mixing across the quad-mixed domains is also proposed to alleviate feature inconsistencies caused by domain-wise mixed-semantic discrepancies.
Furthermore, to deal with persistent category-aware domain gaps, we design a unified feature aggregation module guided by spatial cues for image UDA-SS and spatio-temporal cues for video UDA-SS, enabling fine-grained domain alignment.
Technically, our unified method goes beyond existing alternatives for image or video UDA-SS with limitations such as intra-domain discontinuity due to over distributed samples~\cite{sac,aaai,davsn,pat}, less generalizable feature distribution due to limited mixing directions~\cite{daformer,cmom,dacs,bdm,adpl}, and feature inconsistencies due to diverse mixed-semantic contexts~\cite{adpl,sepico}.
 
We make the following contributions:
(1) We \textit{for the first time} investigate image and video UDA-SS from a unified perspective due to their shared major challenges.
(2) We introduce a unifying UDA-SS framework, Quad-directional Mixup (QuadMix),
which involves pixel- and feature-level mixing to facilitate comprehensive domain gap bridging and
alleviate feature inconsistencies through four-directional paths for intra-
and inter-domain mixing. For temporal shifts in videos, we incorporate optical flow-guided spatio-temporal feature aggregation, which is well-extensible to image scenarios.
(3) As shown in Fig~\ref{fig. 1}, extensive experiments on four image and video benchmarks show that our method sets the new state-of-the-art performance in comparison to prior works.

\section{Related Works}

\subsection{Unsupervised Domain Adaption for Image Segmentation}
Image UDA-SS has garnered significant attention in both theoretical and practical research. 
At the image level, the mainstream is to transfer image styles while preserving original semantics. Various techniques including cycle translation~\cite{cycada}, patch adversarial~\cite{related-49}, dual path learning~\cite{adpl}, and diffusion models~\cite{related-50} have been proposed. At the feature level, influential metrics and the variants~\cite{intro2-13,intro2-14,related-52} are developed to address distribution discrepancies across domains. Additionally, using prototypes to represent feature centroids is proposed for pseudo-label denoising~\cite{caguda}, divergence measurement~\cite{proda},~\cite{dtst}, and contrastive learning~\cite{sepico,enthi}. At the output level, adversarial adaptation on low-dimensional logit space has been proposed in~\cite{related-47}. Other works, such as progressive confidence~\cite{related-55} and curriculum learning~\cite{related-56, reliable}, aim to refine target pseudo-labels within a structured output space~\cite{outputda}. Moreover, from the perspective of feature distribution, generating intermediate samples~\cite{dacs,daformer,sepico, bdm,adpl} to mitigate domain gaps has yielded impressive results. However, the intra-domain discontinuity, less generalizable distribution, and feature inconsistencies are significant challenges that often hinder effective knowledge transfer to the target domain.

\subsection{Unsupervised Domain Adaptive Video Segmentation}
Extending image segmentation, Video Semantic Segmentation (VSS) entails pixel-level discrimination for adjacent video frames~\cite{related-29, related-30}. Mainstream works focus on enhancing segmentation accuracy, proposing methods involving feature warping~\cite{accel,related-32}, recurrent networks~\cite{related-33}, or auxiliary networks~\cite{related-34}. To improve inference efficiency, methods like attention-based spatial feature reconstruction~\cite{related-37} and weight-sharing subnetworks~\cite{related-36} have been proposed. However, they heavily rely on laborious dense labels, posing challenges in adapting to shifted and unlabeled target videos.

\begin{figure*}[!t]
	\centering
	\includegraphics[width=0.98\linewidth]{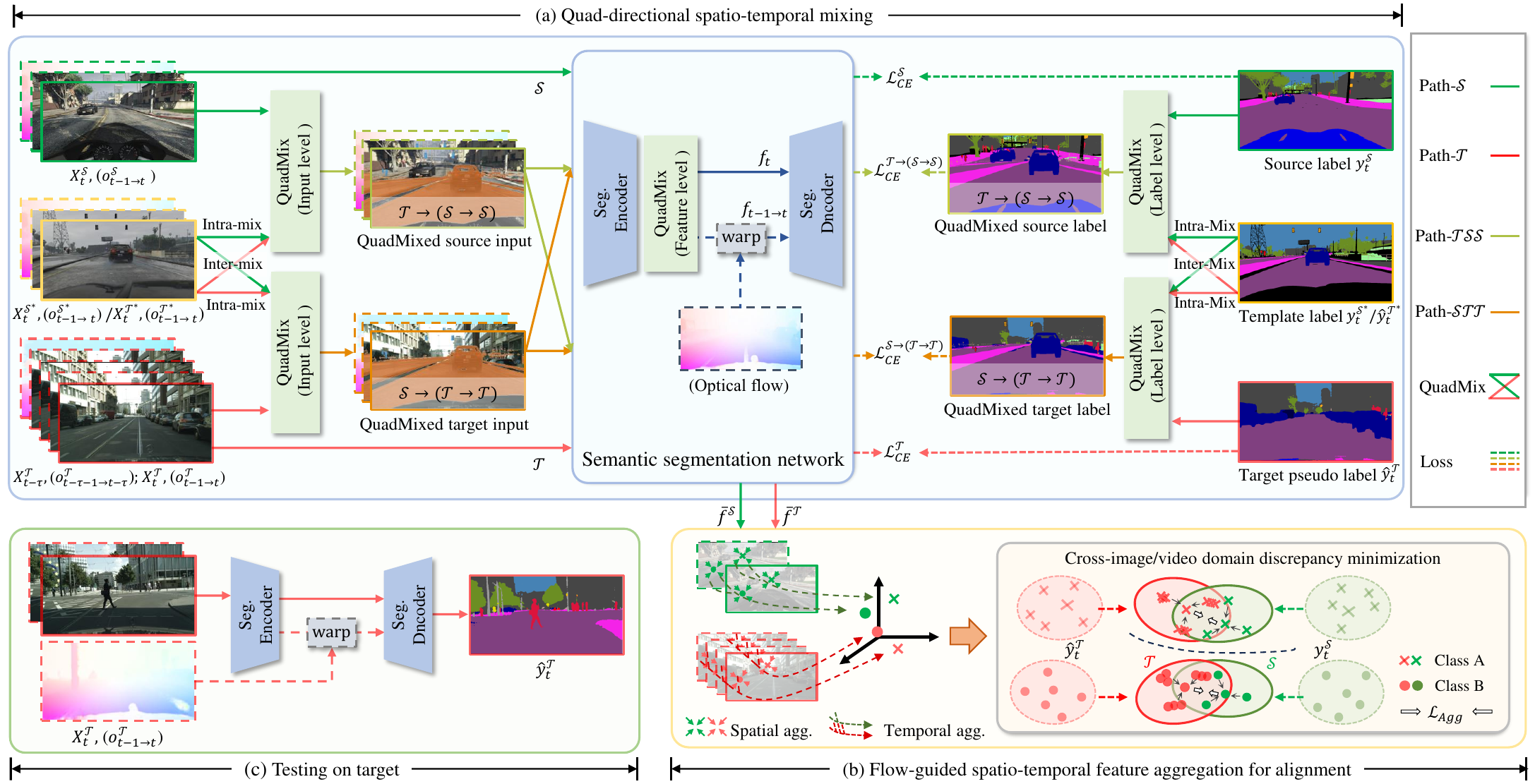} 
	\caption{Overview of the proposed QuadMix for UDA-SS. \textit{\textbf{Image UDA-SS}} follows a parallel approach, with the exception of temporal cues, as indicated by the dashed lines. (i) In part (a), QuadMix comprises four comprehensive intra/inter-domain mixing paths to bridge domain gaps at spatio-temporal pixel and feature levels. Pixel-level mixing is performed on adjacent frames, optical flow, and labels/pseudo-labels, aiming to generate two enhanced inter-mixed domains that derived from intra-mixed domains iteratively: $\mathcal{T} \rightarrow (\mathcal{S} \rightarrow \mathcal{S})$ and $\mathcal{S} \rightarrow (\mathcal{T} \rightarrow \mathcal{T})$. These intermediates overcome the \textit{intra-domain discontinuity} within $\mathcal{S}$ and $\mathcal{T}$ and exhibit more \textit{generalizable features} for gap bridging. Additionally, feature-level mixing across quad-mixed domains alleviates \textit{feature inconsistencies} caused by distinct domain-wise video contexts; (ii) In part (b), optical flow-guided spatio-temporal feature aggregation compresses video features across domains into a compacted category-aware space, minimizing intra-category discrepancies and enhancing inter-category discriminability for target domain; (iii) The training process is end-to-end. In part (c), stacked adjacent frames $X_t^\mathcal{T}$ and optical flow $o_{t - 1 \to t}^{ \mathcal{T}}$ are needed for target domain testing.}
	\label{fig. 2}
    \vspace{-10pt}
\end{figure*}

 \begin{figure}[!t]
	\centering
	\includegraphics[width=0.9\linewidth]{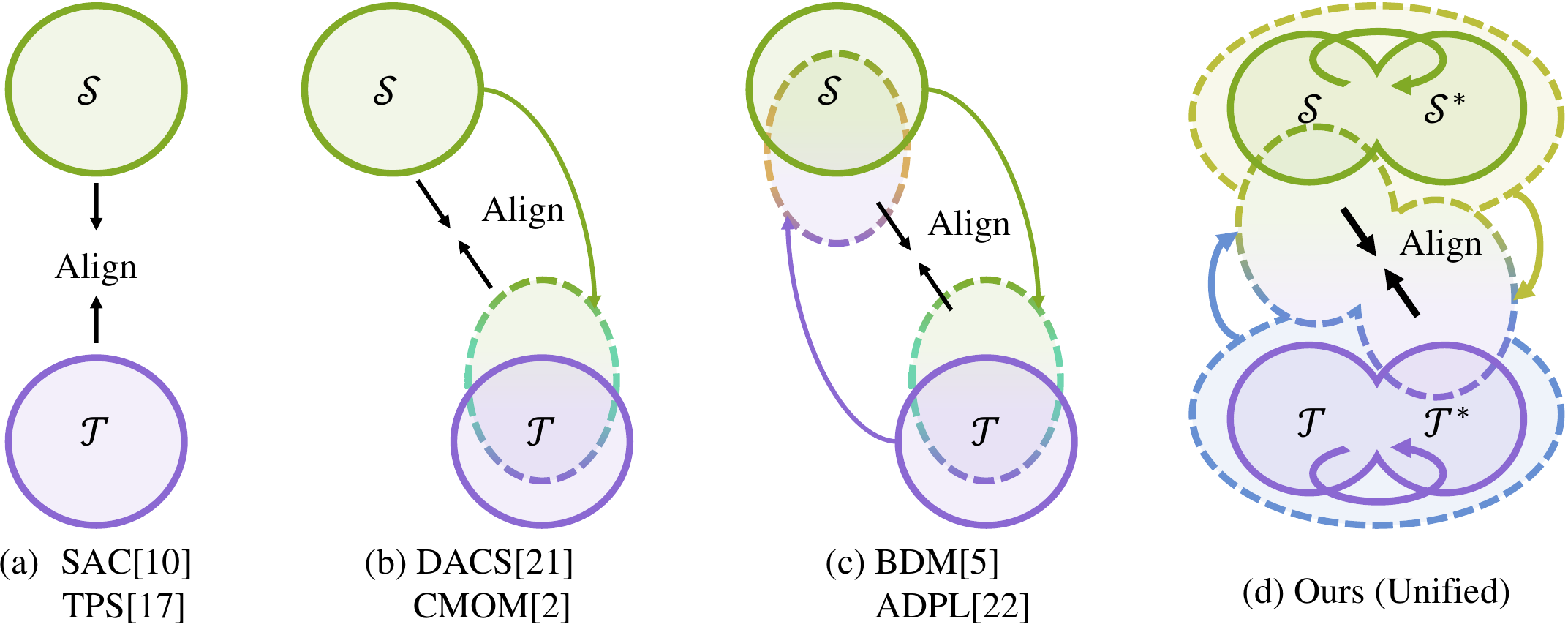}
	\caption{Comparison of domain mixing paradigms. 
    Going beyond existing ideas with the limitations such as intra-domain discontinuity~\cite{sac,tps}, less generalizable feature distribution~\cite{dacs,cmom,bdm,adpl}, and feature inconsistencies~\cite{sac,tps,dacs,cmom,bdm,adpl}, the proposed QuadMix generalizes \textbf{intra-mixed} domains and enhances \textbf{inter-mixed} domains at both spatial (temporal) pixel- and feature-levels. The symbol ``*'' denotes the sample templates.}
	\label{fig. duibi}
         \vspace{-10pt}
\end{figure}

To address this, video UDA-SS has recently gained great interest. Pioneering works~\cite{vat,davsn} employ adversarial learning with DCGAN~\cite{dcgan} to extract domain-invariant video features. Guan et al.~\cite{davsn} further improve this by aligning uncertain target predictions with more confident adjacent ones. However, asynchronous issues with generator-discriminator convergence often cause training instability. Additionally, auxiliary cues involving geometric motion~\cite{moda,wufmf} or optical flow~\cite{tps, aaai} are introduced to generate pseudo-labels. However, they train the model on samples from each domain alternately but overlook domain interactions, leading to isolated learning that hampers adaptation. Meanwhile, Wu et al.~\cite{necess} explore transferring spatial knowledge from image to video domains while overlooking temporal feature gaps. Cho et al.\cite{cmom} perform one-way pixel-level mixing from source to target but overlook the intra-domain discontinuity across domains. In addition, they obtain pseudo-labels offline using models in~\cite{davsn} and only align single-frame spatial features. In contrast, our method is trained end-to-end and holistically mitigates temporal shifts.
 
From a unified view, we consider the image to be a special case of the video without the temporal dimension. For video UDA-SS, we further leverage \textbf{video temporal cues} fully by (1) using optical flow to generate cross-frame consistent pseudo-labels; (2) integrating temporal cues to spatial features for each domain; and (3) aggregating video features of consecutive frames for fine-grained domain alignment. This is the first work applying the vision transformer \cite{segformer} to video UDA-SS.

\subsection{Intermediate Domains for Adaption}
Current works have achieved impressive results by addressing feature gaps with meticulously designed intermediate domains for images~\cite{dacs,bdm,adpl,sepico,daformer,unimix} and videos~\cite{cmom}. However, the discontinuous distribution of features within each domain (intra-domain) always results in distinct point attributes~\cite{mani2014}, which makes current intermediates derived from such distinct domains with poorly generalized distributions~\cite{mani2013}. Fig.~\ref{fig. duibi} (a)-(d) present a comparison of existing methods, distinguishing source features in green and target features in purple. Inspired by~\cite{classmix,cutmix}, strategies like one-way mixing~\cite{dacs,daformer,sepico,cmom} and two-way mixing~\cite{bdm,adpl} extensively explore the integration of source patches into the target domain, or vice versa,~aiming to bridge domain gaps. Nonetheless, t-SNE~\cite{tsne} feature visualization in Fig.~\ref{fig. 12} (a)(c)(d) reveals that the effect of existing intermediates across domains (inter-domain) is fragmented and insufficient, and persistent category-aware domain gaps continue to hinder effective knowledge generalization. Moreover, pure pixel-level mixing may lead to model confusion because of the feature inconsistencies caused by distinct domain-wise contexts.

\section{Method}
In this section, we start with an overview in Section~\ref{subsec:overview}. Then, we elaborate on pixel-level mixing for intra- and inter-domains, as well as feature-level mixing across the overall quad-mixed domains in Section~\ref{subsec:322} to Section~\ref{subsec:inter}. Next, Section~\ref{subsec:castfa} presents the flow-guided spatio-temporal feature aggregation module, and Section~\ref{subsec:loss} covers the training loss.
\subsection{Method Overview}
\label{subsec:overview}

Generally, current studies tend to address image and video UDA-SS tasks independently despite the shared challenges, leading to redundant efforts and missed opportunities for cross-pollination. In this paper, we tackle the more challenging task of unified UDA-SS. As depicted in Fig.~\ref{fig. 2}, we propose a QuadMix method that integrates four-directional paths designed for both intra- and inter-domains, operating at spatial (frames and labels), temporal (optical flow of videos), and feature levels. Technically, the online generated category-aware video/image patch templates from two domains are internally fused with their original domains to create intra-mixed source $(\mathcal{S}$$\rightarrow$$\mathcal{S})$ and target $(\mathcal{T}$$\rightarrow$$\mathcal{T})$ domains with smoother and generalized feature embeddings. Additionally, both pixel- and feature-level patch templates are interactively fused with the intra-mixed domains to generate more enhanced inter-mixed source $(\mathcal{T}$$\rightarrow$$(\mathcal{S}$$\rightarrow$$\mathcal{S}))$ and target $(\mathcal{S}$$\rightarrow$$(\mathcal{T}$$\rightarrow$$\mathcal{T}))$ domains, serving as comprehensive and efficient intermediates for gap bridging (see Fig.~\ref{fig. 12} (e)). Our QuadMix is fairly practical and effective. A detailed comparison with previous works is shown in Fig.~\ref{fig. duibi}.

Video is uniquely characterized by the temporal dimension as compared to image.
To account for this intrinsic information,
we propose an adaptive flow-guided spatial-temporal feature aggregation module for fine-grained domain alignment (Fig.~\ref{fig. 2} (b)).
The spatial feature of each frame is enriched with temporal knowledge of previous frames via optical flow propagation. Based on this, spatial-level aggregation is guided by pseudo masks from the source and target domain adaptively. While at the temporal level, features are further integrated based on logit entropy to enhance overall temporal coherence. The aggregated features across domains are finely aligned in a fine-grained manner. It can be extended to image UDA-SS using spatial feature aggregation, as an image is essentially a special case of a video without the temporal dimension.

The main notations are listed as follows. For video UDA-SS, let $X_t^\mathcal{S} = \mathbb{S}\{ x_{t - 1}^\mathcal{S},x_t^\mathcal{S}\} $ with $o_{t - 1 \to t}^{\mathcal{S}}$ representing source frames alongside optical flow, and $y_{t}^\mathcal{S}$ to be the truth label of $x_t^\mathcal{S}$, where $\mathbb{S}$ denotes a stack operation along time dimension. The target frames and optical flow are $X_t^\mathcal{T} = \mathbb{S}\{ x_{t - 1}^\mathcal{T},x_t^\mathcal{T}\}$ and $o_{t - 1 \to t}^{ \mathcal{T}}$ but without label $y_{t}^\mathcal{T}$. For image UDA-SS, the source image and label are denoted as $x^\mathcal{S}$ and $y^\mathcal{S}$ while the target image is $x^\mathcal{T}$ without label $y^\mathcal{T}$. As for the pseudo-label generation, for the video scenario, $X_{t - \tau }^\mathcal{T} = \mathbb{S}\{ x_{t - \tau - 1}^\mathcal{T},x_{t - \tau }^\mathcal{T}\}$ and $o_{t-\tau-1  \to t-\tau}^{ \mathcal{T}}$ from previous timestamp are input into the model to obtain $\hat y_{t - \tau }^{\mathcal{T}}$, which is then warped using $o_{t-\tau \to t}^{ \mathcal{T}}$ to generate temporally consistent $\hat y_t^{\mathcal{T}}$, where $\tau$ denotes the video temporal deviation. For the image scenario, pseudo-labels are generated by a momentum network~\cite{mean} for each single image.

  \begin{figure*}[!t]
	\centering
	\includegraphics[width=0.98\linewidth]{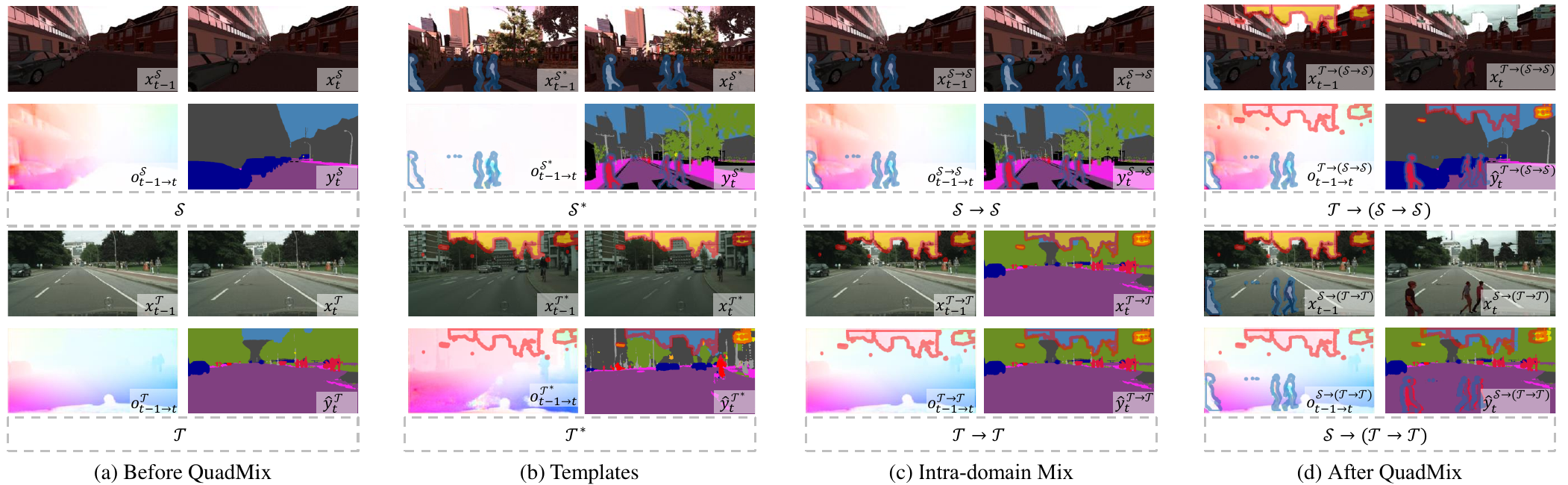}
    \vspace{-5pt}
	\caption{Examples of various mixing strategies in QuadMix for \textit{\textbf{video UDA-SS}}: (a) $\mathcal{S}$ and $\mathcal{T}$ (before QuadMix), (b) $\mathcal{S^*}$ and $\mathcal{T^*}$ (source templates: $\textit{person}$ and $\textit{rider}$, target templates: $\textit{sign}$ and $\textit{sky}$), 
    (c) $\mathcal{S} \rightarrow \mathcal{S}$ and $\mathcal{T} \rightarrow \mathcal{T}$ (intra-domain mixing), (d) $\mathcal{S} \rightarrow (\mathcal{T} \rightarrow \mathcal{T})$ and $\mathcal{T} \rightarrow (\mathcal{S} \rightarrow \mathcal{S})$ (further with inter-domain mixing, i.e. after QuadMix). The effects of these strategies on video frames, optical flow, and labels/pseudo-labels are illustrated. We present $x_t^{\mathcal{T} \rightarrow (\mathcal{S}\rightarrow \mathcal{S})}$ and $x_t^{\mathcal{S} \rightarrow (\mathcal{T}\rightarrow \mathcal{T})}$ without masks for better understanding. Notably, the patch templates required in training iteration $n$ are generated online adaptively from iteration $n-1$. Please zoom in for details.
    }
	\label{fig. 4}
    \vspace{-10pt}
\end{figure*}

\subsection{Video/Image Patch Templates Generation Across Domains}
\label{subsec:322}
The domain gap becomes more evident when we observe the motion of diverse objects at the category-aware level. To mitigate the intrinsic distributional gaps caused by \textit{intra-domain discontinuity} and \textit{less generalizable feature distribution} within both video and image scenarios, we propose quad-directional mixing, which generates intermediate intra- and inter-domain representations within a unified feature space, thereby generalizing features and bridging domain gaps. This design necessitates the adaptive generation of diverse and reliable video/image patch sequences as templates, as denoted in Fig.~\ref{fig. 4}. In this paper, we adaptively generate cross-domain patch templates online. Specifically, for training iteration $n$, the source domain patch templates $Z_{t,n}^{\mathcal{S^*},k}$, $k\in K_{\mathcal{S}}$ and target domain patch templates $Z_{t,n}^{\mathcal{T^*},k}$, $k\in K_{\mathcal{T}}$ required for quad-directional mixing are derived from the previous iteration $n-1$, where the symbol ``*'' denotes the templates. These patch templates contain pixel regions belonging to specific categories $k$, which are randomly selected from segmentation label spaces $K_{\mathcal{S}}$ and $K_{\mathcal{T}}$, respectively.

For \textbf{video UDA-SS}, patch templates form a unified set of category-aware adjacent frames, labels/pseudo-labels, and optical flow. The target video patch template is defined as:
\begin{equation}
	Z_{t,n}^{\mathcal{T^*},k} = \left\{X_{t,n-1}^{\mathcal{T^*},k}, \hat y_{t,n-1}^{\mathcal{T^*},k}, o_{t - 1 \to t,n-1}^{\mathcal{T^*},k}  \right\},
	\label{eq:z_tmpl}
\end{equation}
specifically, 
\begin{equation}
	X_{t,n-1}^{\mathcal{T^*},k}=X_{t,n-1}^{\mathcal{T^*}}\odot\mathds{1}(\mathcal{F}(\hat Y_{t,n-1}^{\mathcal{T}^*})==k), 
	\label{eq:x_tmpl}
\end{equation}
\begin{equation}
	\hat y_{t,n-1}^{\mathcal{T^*},k}=\hat y_{t,n-1}^{\mathcal{T^*}}\odot\mathds{1}(\mathcal{F}(\hat y_{t,n-1}^{\mathcal{T}^*})==k), 
	\label{eq:y_tmpl}
\end{equation}
\begin{equation}
	o_{t-1\to t,n-1}^{\mathcal{T^*},k}=o_{t-1\to t,n-1}^{\mathcal{T^*}}\odot\mathds{1}(\mathcal{F}(\hat y_{t-1,n-1}^{\mathcal{T}^*})==k),
	\label{eq:o_tmpl}
\end{equation}
where $k\in K_{\mathcal{T}^*}$ and corresponding to two selected categories, $\mathds{1}(\cdot)$ is the indicator function, $\odot$ refers to the pixel mapping operation using Hadamard product, and $\mathcal{F}$ stands for a target filtering operation based on the confidence level of segmentation logits. Detailed depictions are shown in Fig.~\ref{fig. 3}(a)(c).

Notably, the model takes $X_t=\mathbb{S}\{x_{t-1}, x_t\}$ with $o_{t-1 \to t}$ (indices omitted) as inputs and generates a segmentation map for $x_t$. Therefore, the bottom-right time subscripts of pseudo-labels are $t$ in Eq.~\eqref{eq:y_tmpl} and $t-1$ in Eq.~\eqref{eq:o_tmpl} for corresponding category-aware masks. Additionally, $\hat Y_{t}^\mathcal{T}=\mathbb{S}\{ \hat y_{t - 1}^\mathcal{T},\hat y_t^\mathcal{T}\}$, the generation of which will be shown in Eq.~\eqref{eq:pseul}. In this paper, masks for pixel mapping evolve with pseudo labels rather than being pre-generated offline as in~\cite{cmom,bdm}, avoiding the accuracy limitation imposed by a frozen warm-up model.

The source video patch template $Z_{t,n}^{\mathcal{S^*},k}$, $k\in K_{\mathcal{S}}$ follows a pattern parallel to Eq.~\eqref{eq:z_tmpl}, just needing to replace $\mathcal{T}$ with $\mathcal{S}$ and employ source labels instead of target pseudo-labels in Eq.~\eqref{eq:z_tmpl} to Eq.~\eqref{eq:o_tmpl}.
To promote feature diversity within quad-mixed domains, we ensure that the categories $k$ of video patch templates differ across domains at each training iteration. 

For \textbf{image UDA-SS}, the image patch template also forms a set but excludes temporal cues, which is denoted as $Z_{t,n}^{\mathcal{T^*},k}=\left\{x_{t,n-1}^{\mathcal{T^*},k}, \hat y_{t,n-1}^{\mathcal{T^*},k}\right\}$ for the target domain, and $Z_{t,n}^{\mathcal{S^*},k}$ similarly for the source domain. For convenience, we retain the subscript $t$ for images, considering them as a special case of videos.

In this way, we generate category-aware patch templates online as templates, which are then utilized for quad-directional mixing to bridge domain gaps in video or image scenarios.
\vspace{-5pt}

\subsection{Intra-Mixed Video/Image Domain Adaptation}
\label{subsec:intra}
\begin{figure*}[!t]
	\centering
	\includegraphics[width=0.95\linewidth]{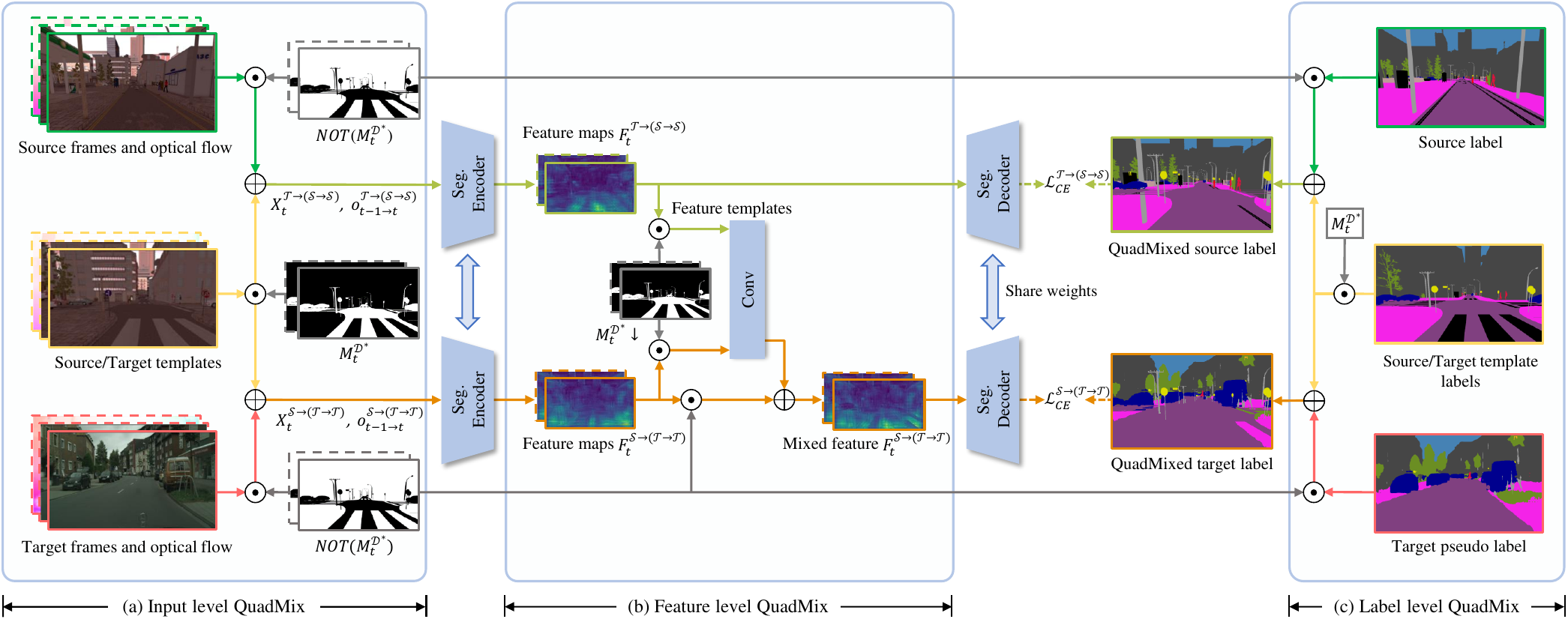}
        \vspace{-5pt}
	\caption{Details of the quad-directional mixing for UDA-SS. To alleviate intra-domain discontinuity for comprehensive domain gap bridging, we conduct QuadMix at the spatial level (video adjacent frames and label), temporal level (optical flow), and spatio-temporal feature level, constructing more enhanced inter-mixed source $(\mathcal{T}$$\rightarrow$$(\mathcal{S}$$\rightarrow$$\mathcal{S}))$ and inter-mixed target $(\mathcal{S}$$\rightarrow$$(\mathcal{T}$$\rightarrow$$\mathcal{T}))$ domains that derived from intra-mixed domains $(\mathcal{S}$$\rightarrow$$\mathcal{S})$ and $(\mathcal{T}$$\rightarrow$$\mathcal{T})$ with generalized feature spaces. The mask $M_t^{\mathcal{D}^*}$ denotes the union of source patch template mask $M_t^{\mathcal{S}^*}$ and target patch template mask $M_t^{\mathcal{T}^*}$.}
	\label{fig. 3}
    \vspace{-10pt}
\end{figure*}

According to Fig.~\ref{fig. duibi}(a) and feature visualization in Fig.~\ref{fig. 12}(a), despite the semantic similarity between source and target domains, a notable feature gap often persists in spatial structure and temporal coherence. Fig.~\ref{fig. 4} presents a depiction of four mixed paths of QuadMix. Rather than purely alternative training with cross-domain data~\cite{sac,tps}, or one way~\cite{dacs,cmom,sepico,daformer}/bidirectional~\cite{bdm,adpl} data pasting guided by offline-generated pseudo-labels via warm-up models stage-wisely~\cite{cmom,bdm}, we propose a comprehensive QuadMix trained end-to-end. A more detailed process is shown in Fig.~\ref{fig. 3}. In this section, we present the intra-mixed domain adaptation.

\subsubsection{Target to Target Mixing $(\mathcal{T}$$\rightarrow$$\mathcal{T})$}
\label{subsec:321}
From the view of feature distribution, we consider source and target video/image domains as two distinct points featured with discontinuous intra-domain feature distribution, and significant inter-domain feature gap. To address the feature discontinuity within each domain for better gap bridging, we propose smoothing and generalizing intra-domains within an explicit feature space. In this part, we focus on the target domain.

For \textbf{video UDA-SS}, adaptive target video patch templates are iteratively integrated into target domain samples, guided by pseudo mask sequences. The process unfolds as follows:
\begin{equation}
	X_{t,n}^{\mathcal{T} \rightarrow \mathcal{T}} = X_{t,n-1}^{\mathcal{T^*},k} + X_{t,n}^{\mathcal{T}} \odot (1 - M_{t,n-1}^{\mathcal{T^*},k}),
	\label{eq:xt2tfusion}
\end{equation}
\begin{equation}
	\hat y_{t,n}^{\mathcal{T} \rightarrow \mathcal{T}} = \hat y_{t,n-1}^{\mathcal{T^*},k} + \hat y_{t,n}^{\mathcal{T}} \odot (1 - m_{t,n-1}^{\mathcal{T^*},k}),
	\label{eq:yt2tfusion}
\end{equation}
\begin{equation}
	o_{t-1\to t,n}^{\mathcal{T} \rightarrow \mathcal{T}} = o_{t-1\to t,n-1}^{\mathcal{T^*},k} + o_{t-1\to t,n}^{\mathcal{T}} \odot (1 - m_{t-1,n-1}^{\mathcal{T^*},k}),
	\label{eq:ot2tfusion}
\end{equation}
for simplicity, Eq.~\eqref{eq:xt2tfusion} to Eq.~\eqref{eq:ot2tfusion} can be unified as follows:
\begin{equation}
	Z_{t,n}^{\mathcal{T} \rightarrow \mathcal{T}} = Z_{t,n}^{\mathcal{T^*},k} + Z_{t,n}^{\mathcal{T}} \odot (1 - M_{t,n-1}^{\mathcal{T^*},k}),
	\label{eq:zt2tfusion}
\end{equation}
where $M_{t,n-1}^{\mathcal{T^*},k}=\mathbb{S}\{m_{t-1,n-1}^{\mathcal{T^*},k}, m_{t,n-1}^{\mathcal{T^*},k}\}$ is the category-aware binary pseudo mask sequence corresponding to video patch template $Z_{t,n}^{\mathcal{T^*},k}$, $k\in K_{\mathcal{T}}$, with a resolution of $B\times H \times W$, where $B$ is the batch size. It can be denoted as follows:
\begin{equation}
	M_{t,n-1}^{\mathcal{T^*},k}= \mathds{1}(\mathcal{F}(\hat Y_{t,n-1}^{\mathcal{T}^*})==k), k \in K_{\mathcal{T}}. 
	\label{eq:m_tmpl}
\end{equation}
In this way, we iteratively integrate category-aware target video templates alongside target video samples mapped by $1-M$ to generate data for the intra-mixed target domain.

For \textbf{image UDA-SS}, the generation of \textit{target to target mixing} samples $Z_{t,n}^{\mathcal{T} \rightarrow \mathcal{T}}$ can be performed following Eq.~\eqref{eq:zt2tfusion} using $M_{t,n-1}^{\mathcal{T^*},k}= \mathds{1}(\mathcal{F}(\hat y_{t,n-1}^{\mathcal{T}^*})==k)$, without the optical flow-level mixing as in Eq.~\eqref{eq:ot2tfusion}.
 
Generally, spatial-level intra-domain mixing is detailed in Eq.~\eqref{eq:xt2tfusion} and Eq.~\eqref{eq:yt2tfusion}, while temporal-level mixing is specified in Eq.~\eqref{eq:ot2tfusion}, where $n$ denotes the training iteration, and $t$ represents the timestamp of adjacent frames within each iteration.
 
\subsubsection{Source to Source Mixing $(\mathcal{S}$$\rightarrow$$\mathcal{S})$}
As illustrated in Fig.~\ref{fig. 4}, the generation of the intra-mixed source domain follows a similar pattern to that of the target domain. To achieve this, we substitute $\mathcal{T}$ with $\mathcal{S}$ in Eq.~\eqref{eq:zt2tfusion} and replace $\mathcal{F}(\hat Y_{t,n-1}^{\mathcal{T}^*})$ with the source domain labels $Y_{t,n-1}^{\mathcal{S}^*}$ in Eq.~\eqref{eq:m_tmpl}, which leverage rich semantic segmentation annotations available in the source domain. It can be denoted as follows:
\begin{equation}
	Z_{t,n}^{\mathcal{S} \rightarrow \mathcal{S}} = Z_{t,n}^{\mathcal{S^*},k} + Z_{t,n}^{\mathcal{S}} \odot (1 - M_{t,n-1}^{\mathcal{S^*},k}),
	\label{eq:zs2sfusion}
\end{equation}
where $Z_{t,n} $ denotes \textit{a unified concept for both video and image scenarios}, and similarly hereinafter.

\textbf{Discussion.} Through intra-domain mixing, we adaptively generate intra-mixed source $(\mathcal{S}$$\rightarrow$$\mathcal{S})$ and target $(\mathcal{T}$$\rightarrow$$\mathcal{T})$ domains. As feature visualization in Fig.~\ref{fig. 12}(b) for video examples, the feature distribution of the latter one is closer to the source domain, and both intra-mixed domains exhibit more generalized feature embeddings. Ablation studies in Table~\ref{tab:table6} demonstrate that our intra-domain mixing (Exp. ID~6) outperforms purely alternating training, one-way or two-way mixing across domains. We attribute this to the generalized and continuous distribution of intra-mixed domains, which diminishes the feature discontinuity within naive domains, thus effectively facilitating knowledge transfer.

\subsection{Inter-Mixed Video/Image Domain Adaptation}
\label{subsec:inter}
The feature gap bridging facilitated by intra-mixed domains prompts us to delve deeper into domain-wise interaction. In addition, existing methods~\cite{dacs,bdm,adpl,sepico,dsp,daformer,cmom} often focus purely on image-level mixing, neglecting the resulting semantic feature inconsistencies, which will impact precise dense prediction on the target domain. Given our design of video or image patch templates at the pixel level, could we further enhance feature consistency across quad-mixed domains at the feature level? Driven by this, building upon our intra-mixed domains, we propose more enhanced inter-mixed domains and feature-level mixing, thereby bridging domain gaps comprehensively and alleviating feature inconsistencies.

Formally, we term the outcome of \textit{target to intra-mixed source mixing} as inter-mixed source domain, and vice versa.

\subsubsection{Target to Intra-Mixed Source Mixing $(\mathcal{T}$$\rightarrow$$(\mathcal{S}$$\rightarrow$$\mathcal{S}))$}
Following our design, the mixed domains should undergo feature-level interaction rather than being trained in isolation. To achieve this, we also utilize source and target patch templates to create more enhanced inter-mixed domains derived from our intra-domains, enabling the model to learn rich and invariant features across domains. The formula for the target to intra-mixed source domain mixing is as follows:
\begin{equation}
	Z_{t,n}^{\mathcal{T} \rightarrow (\mathcal{S}\rightarrow \mathcal{S})} = Z_{t,n}^{\mathcal{T^*},k} + Z_{t,n}^{\mathcal{S}\rightarrow \mathcal{S}} \odot (1 - M_{t,n-1}^{\mathcal{T^*},k}).
	\label{eq:zt2ssfusion}
\end{equation}

For \textbf{image UDA-SS}, the generated inter-mixed source domain sample is a set that integrates spatial elements. While for \textbf{video UDA-SS}, it further involves temporal elements, akin to Eq.~\eqref{eq:z_tmpl}, which encompasses video frames, labels, and inter-frame optical flow. Notably, the inter-mixed source domain integrates semantics from both source and target domains, featured with a more diverse and enhanced feature embedding space, which can facilitate the domain gap bridging.

\subsubsection{Source to Intra-Mixed Target Mixing $(\mathcal{S}$$\rightarrow$$(\mathcal{T}$$\rightarrow$$\mathcal{T}))$ }
The generation of the inter-mixed target domain follows a similar approach to Eq.~\eqref{eq:zt2ssfusion}:
\begin{equation}
	Z_{t,n}^{\mathcal{S} \rightarrow (\mathcal{T}\rightarrow \mathcal{T})} = Z_{t,n}^{\mathcal{S^*},k} + Z_{t,n}^{\mathcal{T}\rightarrow \mathcal{T}} \odot (1 - M_{t,n-1}^{\mathcal{S^*},k}).
	\label{eq:zs2ttfusion}
\end{equation}

Compared to the large feature gap between naive source and target domains, the gap between the inter-mixed domains is narrowed effectively, as depicted in Fig.~\ref{fig. 12}(a)(e). This facilitates the transfer of rich knowledge learned from the source domain to the target domain, promoting effective domain adaptation. The partial labels from source patch templates within the inter-mixed target domain could aid in the learning of global features. Notably, to enhance the diversity of patch templates, we ensure that categories of cross-domain templates are exclusive.

In this paper, we collectively refer to the two \textit{inter-mixed} domains as \textit{quad-mixed} domains because they are derived from intra-mixed domains and undergo quad-directional mixing.

\subsubsection{Feature-Level Template Mixing}
\label{subsubsec:bfllf}
Based on the design that the generated quad-mixed domains encompass shared category-aware sample templates from both the source and target domains, we encourage the model to alleviate \textit{feature inconsistencies} arising from domain-wise mixed-semantic contexts and prioritize learning domain-invariant features of them.

For \textbf{video UDA-SS}, to enhance the consistency learning of diverse spatio-temporal features, we perform feature-level template mixing across the generated quad-mixed domains adaptively, as depicted in Fig.~\ref{fig. 3}(b). Specifically, feature-level mask sequence $M_{t,n-1}^{\mathcal{D}^*}\downarrow$ is generated to map spatio-temporal features from inter-mixed source domain samples, the mapping results are then fused with inter-mixed target domain features. The process is as follows:
\begin{equation}
	\begin{aligned}
	F_{t,n}^{\mathcal{S} \rightarrow (\mathcal{T}\rightarrow \mathcal{T})} = & \psi(F_{t,n}^{\mathcal{T} \rightarrow (\mathcal{S}\rightarrow \mathcal{S})} \odot M_{t,n-1}^{\mathcal{D}^*}\downarrow,  F_{t,n}^{\mathcal{S} \rightarrow (\mathcal{T}\rightarrow \mathcal{T})}  \\  & \hspace{-3em} \odot  M_{t,n-1}^{\mathcal{D}^*}\downarrow)   + F_{t,n}^{\mathcal{S} \rightarrow (\mathcal{T}\rightarrow \mathcal{T})} \odot (1 - M_{t,n-1}^{\mathcal{D}^*}\downarrow),
	\end{aligned}
	\label{eq:featurefusion}
\end{equation}
where $F_{t,n}$ denotes the feature stack of adjacent frames, $\psi$ denotes a $1 \times 1$ convolution layer, and the category-aware feature binary mask $M_{t,n-1}^{\mathcal{D}^*}\downarrow$ is downsampled from the integrated $M_{t,n-1}^{\mathcal{D}^*}$ using bilinear interpolation. The resolution of $M_{t,n-1}^{\mathcal{D}^*}\downarrow$ is same as $F_t$, and $M_{t,n-1}^{\mathcal{D}^*}$ is obtained as follows:
\begin{equation}
	M_{t,n-1}^{\mathcal{D}^*} = M_{t,n-1}^{\mathcal{S}^*} \cup M_{t,n-1}^{\mathcal{T}^*},
	\label{eq:masku}
\end{equation}
where $\mathcal{D}$ denotes the union of source $\mathcal{S}$ and target $\mathcal{T}$ domain.

For \textbf{image UDA-SS}, $F_{t,n}$ corresponding to spatial image features, following a parallel manner.

\textbf{Discussion.} From a unified view, we propose a fairly simple, efficient, and practical solution to bridge video/image feature gaps and alleviate feature inconsistencies, facilitated by quad-directional mixing at pixel and feature levels. The diverse and rich mixed-domain contexts aid in the domain gap mitigation, as illustrated in Fig.~\ref{fig. duibi}(d) and feature visualization in Fig.~\ref{fig. 12}(e). 

Notably, we only utilize quad-mixed data $(\mathcal{T}$$\rightarrow$$(\mathcal{S}$$\rightarrow$$\mathcal{S}))$ and $(\mathcal{S}$$\rightarrow$$(\mathcal{T}$$\rightarrow$$\mathcal{T}))$ for training. Compared to incrementally using $(\mathcal{S}$$\rightarrow$$\mathcal{S}+\mathcal{T}$$\rightarrow$$\mathcal{T}+\mathcal{S}$$\rightarrow$$\mathcal{T}+\mathcal{T}$$\rightarrow$$\mathcal{S})$, our method yields superior results while reducing computational costs. Detailed ablation analysis is provided in Table~\ref{tab:table6} and Table~\ref{tab:table10}.

\subsection{Flow-Guided Feature Aggregation for Domain Alignment}
\label{subsec:castfa}
To further mitigate domain gaps for \textbf{video UDA-SS}, we propose compressing video features into compact distribution for fine-grained alignment. Unlike only aligning spatial features of single frames~\cite{cmom}, we fully leverage temporal cues to generate cross-frame consistent target pseudo-labels, maintain temporal consistency of spatial features for each domain, and aggregate spatio-temporal features guided by optical flow for precise domain alignment, thus enhancing segmentation for the target domain. Note that it is extendable to \textbf{image UDA-SS}.

\subsubsection{Flow-Guided Spatio-Temporal Feature Extraction}
\label{subsubsec:stfe}
In this paper, we employ a variant of ACCEL~\cite{accel} as the video segmentation network, which involves two weight-shared CNN/transformer-based segmentation branches and an optical flow branch. The segmentation branches capture the spatial feature, taking stacked adjacent frames $X_t=\mathbb{S}\{ x_{t - 1},x_t\}$ as the input. The optical flow $o_{t - 1 \to t}$ is used to propagate temporal information across frames. We conduct feature-level template mixing proposed in Section~\ref{subsubsec:bfllf} during forward training across quad-mixed domain samples.

For clarity, we illustrate with pure source and target domain, showing that the inter-mixed source and target domain also adhere to the same paradigm. During model training, the deep feature $F_t=\mathbb{S}\{ f_{t - 1},f_t\}$ of stacked adjacent frames are fused to generate the holistic feature of frame $x_t$, leveraging the video temporal continuity. Specifically, we feed feature $f_t$ and optical flow-warped feature $f_{t - 1 \to t}$ into a convolutional fusion layer $\phi_f$ to generate fused feature vector $\bar{f}_{t - 1 \to t}$. The process is described as follows:
\begin{equation}
	\bar{f}_{t - 1 \to t} = \phi_f(f_{t - 1 \to t},{\kern 1pt} {\kern 1pt} {\kern 1pt} f_t),
	\label{eq:prediction}
\end{equation}
where the optical flow-warped feature $f_{t - 1 \to t}$ is obtained by:
\begin{equation}
	f_{t - 1 \to t} = \phi_\mathcal{W}(f_{t - 1},o_{t - 1 \to t}),
	\label{eq:warpfeature}
\end{equation}
where $\phi_\mathcal{W}$ is a warp operation using bilinear interpolation.

For target domain self-supervised training, we fully exploit inherent temporal continuity to generate cross-frame coherent pseudo-labels. As shown in the left part of Fig.~\ref{fig. 2}(a), consecutive sets $Z_{t- \tau}^{\mathcal{T}}$ are fed into the segmentation network to generate the feature $\bar{f}_{t - \tau -1 \to t - \tau}^\mathcal{T}$. Then, we yield pseudo-label $\hat y_t^\mathcal{T}$ with optical flow $o_{t - \tau \to t}^{ \mathcal{T}}$ warpping, as described below:
\begin{equation}
	\hat y_t^\mathcal{T} = {\arg\max _{k}}(\phi_\mathcal{W}(\phi_\mathcal{D}(\bar{f}_{t - \tau -1 \to t - \tau}^{\mathcal{T},(k)})),o_{t - \tau \to t} ^{ \mathcal{T},(k)}),
	\label{eq:pseul}
\end{equation}
where $\phi_\mathcal{D}$ denotes the segmentation decoder layer.
For target domain testing, as shown in Fig.~\ref{fig. 2}(c), only stacked sets $Z_t^{\mathcal{T}}$ are input to the model to obtain segmentation results.

\subsubsection{Spatio-Temporal Feature Aggregation for Domain Alignment}
\label{subsubsec:dacaf}
To further mitigate the video domain gap, we propose adaptive aggregation along spatial and temporal dimensions to reduce the distribution discrepancy of scattered video features. Unlike previous video UDA-SS methods~\cite{cmom,necess} only emphasizing image feature discrepancy across domains that may overlook holistic temporal learning, we comprehensively consider spatial features of consecutive frames and fully leverage temporal information. In the spatial dimension, fused features $p_{t' \to t}^{\mathcal{T}}$ that integrate flow-guided features from previous timestamps $t'$ to frame $t$ are adaptively aggregated. For the target domain, we set $t' \in \mathbb{T}=\{ t - 1,t - \tau,t - \tau  - 1\}$, while for the source domain, $t' \in \mathbb{T}=\{ t - 1\}$, benefiting from source label supervision to reduce computational cost.

\begin{figure}[!t]
	\centering
	\includegraphics[width=\linewidth]{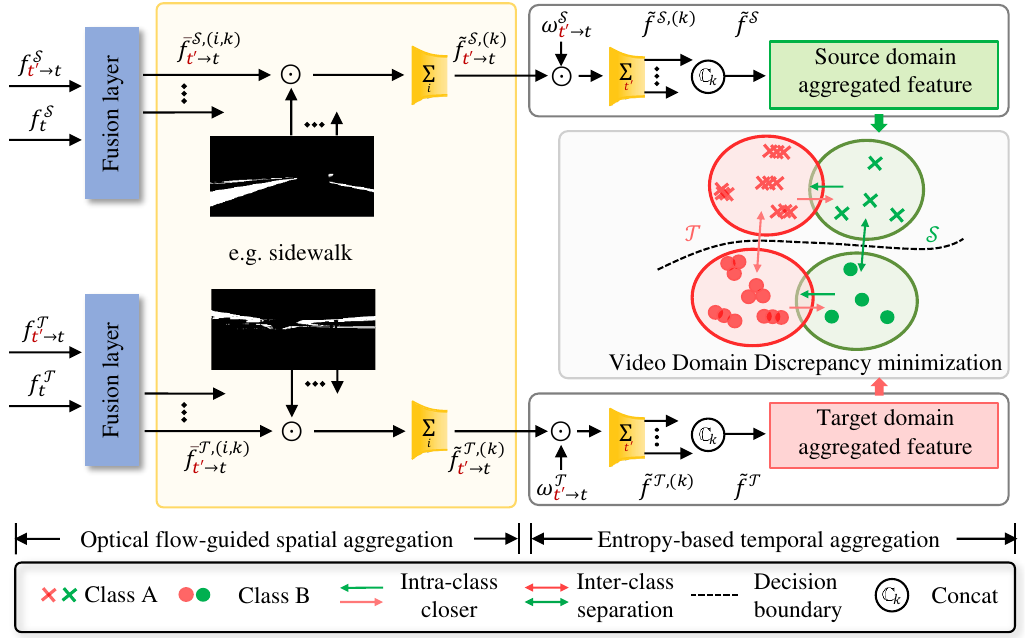}
	\caption{The illustration of the optical flow-guided spatio-temporal feature aggregation for domain alignment, where $t'$ signifies previous time step, and $\omega _{t' \to t}^\mathcal{T}$ denotes temporal aggregation weights of target frames. The symbol $f_{t' \to t}$ denotes the optical flow-guided frame feature warped from the previous time step $t'$, where ``$\to$" denotes the warp direction along the time dimension.}
	\label{fig. 6}
    \vspace{-10pt}
\end{figure}

Specifically, as shown in Fig.~\ref{fig. 6}, we warp frame features for two domains from the previous time step $t'$ to $t$ following Eq.~\eqref{eq:warpfeature} to get the warped feature $f_{t' \to t}$, respectively. Then, upon merging $f_{t' \to t}$ with feature $f_t$ through the fusion layer $\phi_f$, we generate temporal consistent spatial feature $\bar{f}_{t' \to t}^{\mathcal{T}}$ for frame $t$. Based on this, we adaptively distribute the spatial feature to category-aware subspaces by pixel mapping operation in a fine-grained manner. For the target domain, we obtain  spatial-aggregated feature $\tilde{f}_{t' \to t}^{\mathcal{T},(k)}$ defined as follows:
\begin{equation}
	\tilde{f}_{t' \to t}^{\mathcal{T},(k)} = \frac{\sum\limits_{i = 1}^{H \times W}(\bar{f}_{t' \to t}^{\mathcal{T},(i,k)} \odot \mathds{1}(\hat y_{t' \to t}^{\mathcal{T},(i,k)}==k))}{\sum\limits_{i = 1}^{H \times W} \mathds{1}(\hat y_{t' \to t}^{\mathcal{T},(i,k)}==k)},
	\label{eq:caf_t}
\end{equation}
where $k$ belongs to the whole category space. And source spatial-aggregated feature $\tilde{f}_{t' \to t}^{\mathcal{S},(k)}$ follows a similar manner. In this way, adaptive aggregation along the spatial dimension across two domains is conducted in a fine-grained way.

Based on this, we performed temporal aggregation by combining spatially aggregated features from consecutive frames to generate spatial-temporal features $\tilde{f}^{\mathcal{T},(k)}$ using entropy confidence~\cite{entroy} $\omega _{t' \to t}$, for the target domain:
\begin{equation}
	{\tilde{f}}^{\mathcal{T},(k)} \leftarrow \sum\limits_{t' \in \mathbb{T}} {\tilde{f}_{t' \to t}^{\mathcal{T},(k)} \cdot \omega _{t' \to t}^\mathcal{T}},
	\label{eq:entropyfusion}
\end{equation}
where $\omega _{t' \to t}^\mathcal{T}$ is obtained by softmax normalization of logits entropy among the fused spatial features, formulated as:
\begin{equation}
	\omega _{t' \to t}^\mathcal{T} = {\rm soft}{\max_{t'}}( -{\frac{1}{{H \times W}}}\!\!\!\!\sum\limits_{i}^{H \times W} \!\!\!{(-\bar{f}_{t' \to t}^{{\mathcal{T},(i)}}\! \cdot \!\log (\bar{f}_{t' \to t}^{{\mathcal{T},(i)}})))},
	\label{eq:omega}
\end{equation}
where the segmentation confidence and temporal weight $\omega _{t' \to t}^\mathcal{T}$ are promoted to be positively correlated. 

Subsequently, we concatenate the aggregated features of each category to generate fine-grained features $\tilde{f}^{(k)}$ for each domain, where the concatenation is denoted as $\mathbb{C}_k$. We can mitigate the fine-grained video feature gap using MMD (Maximum Mean Discrepancy)~\cite{intro2-13} minimization, thanks to the optical flow-guided temporal information propagation and distribution alignment. 

As for \textbf{image UDA-SS}, spatial feature aggregation can be used for fine-grained alignment without relying on temporal cues, i.e., optical flow guidance.

 \begin{table*}
	\caption{\textbf{video UDA-SS} results on Synthia-Seq $\rightarrow$ Cityscapes-Seq. The best results are in \textbf{bold}, and the suboptimal results are in \underline{underlined}. ``Adv.": adversarial learning, ``SSL": self-supervised learning, ``Warm": warm-up, ``ViT": vision transformer-based network. \textsuperscript{\dag} denotes retrained results of image domain methods using a video segmentation backbone~\cite{accel} for fair comparison.
		\label{tab:table1}}
  \setlength{\tabcolsep}{2.4mm}
	\centering
	\renewcommand{\arraystretch}{1.25}
	\footnotesize
	\scalebox{0.95}{
		\begin{tabular}{cccccccccccccccc}
			\bottomrule[1.0pt]
			\multicolumn{1}{c|}{Methods}          & Adv. & SSL & \multicolumn{1}{c|}{Warm} & Road & Side. & Buil. & Pole & Light & Sign & Vege. & Sky  & Pers. & Rider & \multicolumn{1}{c|}{Car}  & mIOU \\ \hline
			\multicolumn{1}{c|}{Source-Only}      &     &    & \multicolumn{1}{c|}{}     & 56.3 & 26.6  & 75.6  & 25.5 & 5.7   & 15.6 & 71.0  & 58.5 & 41.7  & 17.1  & \multicolumn{1}{c|}{27.9} & 38.3 \\ 
			\multicolumn{1}{c|}{AdvEnt\textsuperscript{\dag}{\cite{advent}}}   & \checkmark    &    & \multicolumn{1}{c|}{}     & 85.7 & 21.3  & 70.9  & 21.8 & 4.8   & 15.3 & 59.5  & 62.4 & 46.8  & 16.3  & \multicolumn{1}{c|}{64.6} & 42.7 \\
			\multicolumn{1}{c|}{CBST\textsuperscript{\dag}{\cite{cbst}}}     &      & \checkmark  & \multicolumn{1}{c|}{}     & 64.1 & 30.5  & 78.2  & 28.9 & 14.3  & 21.3 & 75.8  & 62.6 & 46.9  & 20.2  & \multicolumn{1}{c|}{33.9} & 43.3 \\
			\multicolumn{1}{c|}{IDA\textsuperscript{\dag}{\cite{ida}}}      & \checkmark    &    & \multicolumn{1}{c|}{}     & 87.0 & 23.2  & 71.3  & 22.1 & 4.1   & 14.9 & 58.8  & 67.5 & 45.2  & 17.0  & \multicolumn{1}{c|}{73.4} & 44.0 \\
			\multicolumn{1}{c|}{CRST\textsuperscript{\dag}{\cite{crst}}}     &      & \checkmark  & \multicolumn{1}{c|}{}     & 70.4 & 31.4  & 79.1  & 27.6 & 11.5  & 20.7 & 78.0  & 67.2 & 49.5  & 17.1  & \multicolumn{1}{c|}{39.6} & 44.7 \\
			\multicolumn{1}{c|}{SVMin\textsuperscript{\dag}{\cite{svmin}}}   &      &    & \multicolumn{1}{c|}{}     & 84.9 & 0.5   & 77.9  & 29.6 & 7.4   & 15.0 & 78.6  & 73.2 & 46.9  & 6.2   & \multicolumn{1}{c|}{73.8} & 44.9 \\
			\multicolumn{1}{c|}{CrCDA\textsuperscript{\dag}{\cite{crcda}}}    & \checkmark    &    & \multicolumn{1}{c|}{}     & 86.5 & 26.3  & 74.8  & 24.5 & 5.0   & 15.5 & 63.5  & 64.4 & 46.0  & 15.8  & \multicolumn{1}{c|}{72.8} & 45.0 \\
			\multicolumn{1}{c|}{RDA\textsuperscript{\dag}{\cite{rda}}}      &      & \checkmark  & \multicolumn{1}{c|}{}     & 84.7 & 26.4  & 73.9  & 23.8 & 7.1   & 18.6 & 66.7  & 68.0 & 48.6  & 9.3   & \multicolumn{1}{c|}{68.8} & 45.1 \\
			\multicolumn{1}{c|}{FDA\textsuperscript{\dag}{\cite{fda}}}      &      & \checkmark  & \multicolumn{1}{c|}{}     & 84.1 & 32.8  & 67.6  & 28.1 & 5.5   & 20.3 & 61.1  & 64.8 & 43.1  & 19.0  & \multicolumn{1}{c|}{70.6} & 45.2 \\
			\multicolumn{1}{c|}{SAC\textsuperscript{\dag}{\cite{sac}}}      &      & \checkmark  & \multicolumn{1}{c|}{}     & 87.0 & 41.1  & 64.0  & 20.4 & 12.1  & 32.8 & 38.2  & 47.6 & 53.1  & 19.3  & \multicolumn{1}{c|}{81.1} & 45.2 \\
			\multicolumn{1}{c|}{DACS\textsuperscript{\dag}{\cite{dacs}}}     &      & \checkmark  & \multicolumn{1}{c|}{}     & 86.4 & 40.0  & 74.0  & 27.8 & 9.5   & 28.2 & 71.6  & 72.0 & 55.6  & 20.0  & \multicolumn{1}{c|}{76.4} & 51.0 \\
			\multicolumn{1}{c|}{PixMatch\textsuperscript{\dag}{\cite{pixmatch}}} &      & \checkmark  & \multicolumn{1}{c|}{}     & 90.2 & \underline{49.9}  & 75.1  & 23.1 & 17.4  & 34.2 & 67.1  & 49.9 & 55.8  & 14.0  & \multicolumn{1}{c|}{84.3} & 51.0 \\ \hline
			\multicolumn{1}{c|}{DA-VSN{\cite{davsn}}}   & \checkmark    &    & \multicolumn{1}{c|}{}     & 89.4 & 31.0  & 77.4  & 26.1 & 9.1   & 20.4 & 75.4  & 74.6 & 42.9  & 16.1  & \multicolumn{1}{c|}{82.4} & 49.5 \\
			
			\multicolumn{1}{c|}{I2VDA{\cite{necess}}}   &     &\checkmark    & \multicolumn{1}{c|}{}     & 89.9 & 40.5  & 77.6  & 27.3 & 18.7   & 23.6 & 76.1  & 76.3 & 48.5  & 22.4  & \multicolumn{1}{c|}{82.1} & 53.0 \\
			\multicolumn{1}{c|}{TPS{\cite{tps}}}      &      & \checkmark  & \multicolumn{1}{c|}{}     & 91.2  & 53.7   & 74.9  & 24.6 & 17.9  & 39.3 & 68.1  & 59.7 & 57.2  & 20.3  & \multicolumn{1}{c|}{84.5} & 53.8 \\
			\multicolumn{1}{c|}{SFC{\cite{aaai}}}   &    & \checkmark    & \multicolumn{1}{c|}{}     & 90.9 & 32.5  & 76.8  & 28.6 &6.0   & 36.7 & 76.0  & 78.9 & 51.7  & 13.8 & \multicolumn{1}{c|}{85.6} & 52.5 \\
			\multicolumn{1}{c|}{TPL-SFC{\cite{aaai}}}   &    & \checkmark    & \multicolumn{1}{c|}{}     & 90.0 & 32.8  & 80.4  & 28.9 &14.9   & 35.3 & 80.8  & 81.1 & 57.5  & 19.6 & \multicolumn{1}{c|}{86.7} & 55.3 \\\multicolumn{1}{c|}{PAT{\cite{pat}}}   &    & \checkmark    & \multicolumn{1}{c|}{}     & \underline{91.5} & 41.3  & 76.1  & 29.6 &20.9   & 33.8 & 72.4  & 75.9 & 51.3  & 24.7 & \multicolumn{1}{c|}{86.2} & 54.9 \\
			
            \multicolumn{1}{c|}{CMOM{\cite{cmom}}}     &      & \checkmark  & \multicolumn{1}{c|}{\checkmark}    & 90.4 & 39.2  & 82.3  & 30.2 & 16.3  & 29.6 & 83.2  & \underline{84.9} & 59.3  & 19.7  & \multicolumn{1}{c|}{84.3} & 56.3 \\ \hline
			\multicolumn{1}{c|}{QuadMix}             &      & \checkmark  & \multicolumn{1}{c|}{}     & 90.5 & 41.2  &  81.1  &  29.3 &  23.1  &  47.5  &   82.7  &  83.8  & 61.1  &  28.4   & \multicolumn{1}{c|}{87.0} &  59.6  \\ 
			\multicolumn{1}{c|}{QuadMix}              &      & \checkmark  & \multicolumn{1}{c|}{\checkmark}    & 90.8 & 39.9  & \textbf{83.2}  & \underline{33.2} & \underline{30.1}  & \underline{50.7} & \underline{84.8}  & 82.3 & \underline{61.2}  & \underline{32.7}  & \multicolumn{1}{c|}{\underline{87.4}} & \underline{61.5}  \\
            \multicolumn{1}{c|}{QuadMix (ViT)}              &      & \checkmark  & \multicolumn{1}{c|}{}    & \textbf{94.1} & \textbf{61.9}  & \underline{82.9}  & \textbf{36.9} & \textbf{41.0}  & \textbf{59.1} & \textbf{85.2}  & \textbf{85.6} & \textbf{64.3}  & \textbf{37.8}  & \multicolumn{1}{c|}{\textbf{90.3}} & \textbf{67.2}  \\        \toprule[1.0pt]
	\end{tabular}}
    \vspace{-10pt}
\end{table*}
\subsection{Model Training}
\label{subsec:loss}
The model is trained jointly using cross-domain samples, with the objective function defined as follows:
\begin{equation}
	{\mathcal{L}_{all}} = {\mathcal{L}_{QuadMix}} + {\mathcal{L}_{Agg}} + {\mathcal{L}_{SSL}}.
	\label{eq:overalloss}
\end{equation}
Each loss is discussed first for \textbf{video UDA-SS} as below.

\textbf{QuadMix loss.} As shown in Fig.~\ref{fig. 3}, the loss for quad-directional mixing used for gap bridging is as follows:
\begin{equation}
	\begin{aligned}
		{\mathcal{L}_{QuadMix}} &= \mathcal{L}_{CE}^{\mathcal{S}'}(\mathcal{G}(X_t^{\mathcal{S}'},o_{t - 1 \to t}^{ \mathcal{S}'}),y_t^{\mathcal{S}'}) \\
		&\quad+ {\lambda _\mathcal{T}}\mathcal{L}_{CE}^{\mathcal{T}'}(\mathcal{G}(\mathcal{A}(X_t^{\mathcal{T}'}),o_{t - 1 \to t}^{ \mathcal{T}'}),\hat y_t^{\mathcal{T}'}){\kern 1pt}
	\end{aligned},
	\label{eq:loss_QuadMix}
\end{equation}
where $\mathcal{S}'$ and $\mathcal{T}'$ indicate two quad-mixed domains for simplicity, $\mathcal{G}$ denotes the semantic segmentation model, $\mathcal{A}$ means image enhancement, and $\mathcal{L}_{CE}$ denotes cross-entropy loss.

\textbf{Aggregated feature alignment loss.}
We define the following loss to minimize the discrepancy of aggregated spatio-temporal feature distribution across domains:
\begin{equation}
	{\mathcal{L}_{Agg}} = {\lambda _f}\left\|\mathbb{E}\left(\mathbb{C}_k\left(f^{\mathcal{S},\left(k\right)}\right)\right)-\mathbb{E}\left(\mathbb{C}_k\left(f^{\mathcal{T},\left(k\right)}\right)\right)\right\|_{\mathcal{H}}^2
	\label{eq:loss_Agg}
\end{equation}
where $\mathbb{E}$ denotes the mathematical expectation, and $\mathcal{H}$ represents the reproducing Hilbert space~\cite{rhks}.

\textbf{Self-supervised learning loss.} We also employ a cross-domain self-supervised learning loss for two video domains:
\begin{equation}
	\begin{aligned}
		{\mathcal{L}_{SSL}} &= \mathcal{L}_{CE}^{\mathcal{S}}(\mathcal{G}(X_t^\mathcal{S},o_{t - 1 \to t}^{ \mathcal{S}}),y_t^\mathcal{S}){\kern 1pt}{\kern 1pt}{\kern 1pt}{\kern 1pt}{\kern 1pt}{\kern 1pt}{\kern 1pt}{\kern 1pt}{\kern 1pt}{\kern 1pt}{\kern 1pt}   \\			&\quad+  {\lambda _\mathcal{T}}\mathcal{L}_{CE}^{\mathcal{T}}(\mathcal{G}(\mathcal{A}(X_t^\mathcal{T}),o_{t - 1 \to t}^{ \mathcal{T}}),\hat y_t^\mathcal{T}){\kern 1pt} {\kern 1pt}
	\end{aligned}.
	\label{eq:loss_ssl}
\end{equation}

For \textbf{image UDA-SS}, the input of the model $\mathcal{G}$ in Eq.~\eqref{eq:loss_QuadMix} and Eq.~\eqref{eq:loss_ssl} will be replaced by a single image $x_t$. The sensitivity analysis for hyperparameters ${\lambda _\mathcal{T}}$ and ${\lambda _f}$ will be discussed in the experiments.
The training algorithm is summarized in supplementary materials.

\section{Experiments}
\subsection{Experimental Setups}
\subsubsection{Video UDA-SS Datasets}
We conduct experiments on two challenging benchmarks: Synthia-Seq~\cite{synthia} $\rightarrow$ Cityscapes-Seq~\cite{citys} and VIPER~\cite{viper} $\rightarrow$ Cityscapes-Seq. \textbf{Cityscapes-Seq} (target domain) is a standard dataset for semantic urban scene understanding, featuring real-world videos from 50 cities in Germany and neighboring countries. It comprises 2,975 training video clips and 500 validation video clips, and each clip contains continuous 30 frames. \textbf{Synthia-Seq} (source domain) contains 8,000 photo-realistic frames with dense segmentation labels. We employ video frames in the \textit{RGB} folder and choose 11 categories common with Cityscapes-Seq for adaptation. \textbf{VIPER} (source domain) is another standard synthetic video dataset, consisting of 133,670 frames of virtual videos rendered from urban landscape scenes in the virtual computer game ``Grand Theft Auto V." We use 13,367 frames with segmentation labels as another source domain, choosing 15 common categories with Cityscapes-Seq for adaptation. Notably, our dataset setting is consistent with previous works~\cite{davsn,tps,cmom,aaai,necess,moda,pat}, etc.
 
\subsubsection{Image UDA-SS Datasets}
We conduct experiments on two popular benchmarks GTAV~\cite{gtav} $\rightarrow$ Cityscapes and SYNTHIA~\cite{synthia} $\rightarrow$ Cityscapes. \textbf{Cityscapes} (target domain) is a subset of Cityscapes-Seq, with each image being the $20$-th frame of each corresponding video clip. \textbf{GTAV} (source domain) is a synthetic image dataset with labels for semantic segmentation, where images are also collected from ``Grand Theft Auto V." It comprises 19 categories common with Cityscapes and 24,966 labeled images. \textbf{SYNTHIA} (source domain) is another synthetic image dataset, we use images from the \textit{SYNTHIARAND-CITYSCAPES} folder which contains 9,400 images and 16 common categories with Cityscapes, akin to previous works~\cite{dacs,bdm,I2f,adpl,sepico,fredom,dsp}, etc.

\begin{table*}
	\caption{\textbf{Video UDA-SS} results on VIPER $\rightarrow$ Cityscapes-Seq. The best results are in \textbf{bold}, and the suboptimal results are in \underline{underlined}. ``Adv.": adversarial learning, ``SSL": self-supervised learning, ``Warm": warm-up, ``ViT": vision transformer-based network. \textsuperscript{\dag} denotes retrained results of image domain methods using a video segmentation backbone~\cite{accel} for fair comparison.
		\label{tab:table2}}
	\centering
	\renewcommand{\arraystretch}{1.37}
	
	\small
	\scalebox{0.78}{
		\begin{tabular}{cccccccccccccccccccc}
			\bottomrule[1.0pt]
			\multicolumn{1}{c|}{Methods}          & Adv. & SSL & \multicolumn{1}{c|}{Warm} & Road & Side. & Buil. & Fenc. & Light & Sign & Vege. & Terr. & Sky  & Pers. & Car & Truc. & Bus  & Mot. & \multicolumn{1}{c|}{Bike}  & mIOU \\ \hline
			\multicolumn{1}{c|}{Source-Only}      & \checkmark    &    & \multicolumn{1}{c|}{}     & 56.7 & 18.7  & 78.7  & 6.0   & 22.0  & 15.6 & 81.6  & 18.3  & 80.4 & 55.9  & 66.3 & 4.5   & 16.8 & 20.4 & \multicolumn{1}{c|}{10.3} & 37.1 \\
			\multicolumn{1}{c|}{AdvEnt\textsuperscript{\dag}{\cite{advent}}}   & \checkmark    &    & \multicolumn{1}{c|}{}     & 78.5 & 31.0  & 81.5  & 22.1  & 29.2  & 26.6 & 81.8  & 13.7  & 80.5 & 58.3  & 64.0  & 6.9   & 38.4 & 4.6  & \multicolumn{1}{c|}{1.3} & 41.2 \\
			\multicolumn{1}{c|}{CBST\textsuperscript{\dag}{\cite{cbst}}}     &      & \checkmark  & \multicolumn{1}{c|}{}     & 48.1 & 20.2  &  84.8 & 12.0  & 20.6  & 19.2 & 83.8  & 18.4  & 84.9 & 59.2  & 71.5 & 3.2   & 38.0 & 23.8 & \multicolumn{1}{c|}{37.7} & 41.7 \\
			\multicolumn{1}{c|}{IDA\textsuperscript{\dag}{\cite{ida}}}      & \checkmark    &    & \multicolumn{1}{c|}{}     & 78.7 & 33.9  & 82.3  & 22.7  & 28.5  & 26.7 & 82.5  & 15.6  & 79.7 & 58.1  & 64.2  & 6.4   & 41.2 & 6.2  & \multicolumn{1}{c|}{3.1} & 42.0 \\
			\multicolumn{1}{c|}{CRST\textsuperscript{\dag}{\cite{crst}}}     &      & \checkmark  & \multicolumn{1}{c|}{}     & 56.0 & 23.1  & 82.1  & 11.6  & 18.7  & 17.2 & 85.5  & 17.5  & 82.3 & 60.8  & 73.6 & 3.6   & 38.9 & 30.5 & \multicolumn{1}{c|}{35.0} & 42.4 \\
			\multicolumn{1}{c|}{SVMin\textsuperscript{\dag}{\cite{svmin}}}   &      &    & \multicolumn{1}{c|}{}     & 51.1 & 14.3  & 80.8  & 11.9  & 30.9  & 23.1 & 83.5  & \textbf{37.7}  & 74.5 & 59.5  & 79.7 & 36.4  & \underline{53.2} & 20.0 & \multicolumn{1}{c|}{4.2}  & 44.1 \\
			\multicolumn{1}{c|}{CrCDA\textsuperscript{\dag}{\cite{crcda}}}    & \checkmark    &    & \multicolumn{1}{c|}{}     & 78.1 & 33.3  & 82.2  & 21.3  & 29.1  & 26.8 & 82.9  & 28.5  & 80.7 & 59.0  & 73.8  & 16.5  & 41.4 & 7.8  & \multicolumn{1}{c|}{2.5} & 44.3 \\
			\multicolumn{1}{c|}{RDA\textsuperscript{\dag}{\cite{rda}}}      &      & \checkmark  & \multicolumn{1}{c|}{}     & 72.0 & 25.9  & 80.8  & 15.1  & 27.2  & 20.3 & 82.6  & 31.4  & 82.2 & 56.3  & 75.5  & 22.8  & 48.3 & 19.1 & \multicolumn{1}{c|}{6.7} & 44.4 \\
			\multicolumn{1}{c|}{FDA\textsuperscript{\dag}{\cite{fda}}}      &      & \checkmark  & \multicolumn{1}{c|}{}     & 70.3 & 27.7  & 81.3  & 17.6  & 25.8  & 20.0 & 83.7  & 31.3  & 82.9 & 57.1  & 72.2  & 22.4  & 49.0 & 17.2 & \multicolumn{1}{c|}{7.5} & 44.4 \\
			\multicolumn{1}{c|}{DACS\textsuperscript{\dag}{\cite{dacs}}}     &      & \checkmark  & \multicolumn{1}{c|}{}     & 69.6 & 24.1  & 76.9  & 9.1   & 16.1  & 15.3 & 74.1  & 20.3  & 76.5 & 59.4  & 74.8 & 38.6  & 43.1 & 7.7  & \multicolumn{1}{c|}{1.9}  & 40.5 \\
			\multicolumn{1}{c|}{SAC\textsuperscript{\dag}{\cite{sac}}}      &      & \checkmark  & \multicolumn{1}{c|}{}     & 52.2 & 19.6  & 73.4  & 3.7   & 23.1  & 25.2 & 73.9  & 17.3  & 78.1 & 56.9  & 80.3 & 38.3  & 48.2 & 17.8 & \multicolumn{1}{c|}{14.1} & 41.5 \\
			\multicolumn{1}{c|}{PixMatch\textsuperscript{\dag}{\cite{pixmatch}}} &      & \checkmark  & \multicolumn{1}{c|}{}     & 79.4 & 26.1  & 84.6  & 16.6  & 28.7  & 23.0 & 85.0  & 30.1  & 83.7 & 58.6  & 75.8 & 34.2  & 45.7 & 16.6 & \multicolumn{1}{c|}{12.4} & 46.7 \\ \hline
			\multicolumn{1}{c|}{DA-VSN{\cite{davsn}}}   & \checkmark    &    & \multicolumn{1}{c|}{}     & 86.8 & 36.7  & 83.5  & 22.9  & 30.2  & 27.7 & 83.6  & 26.7  & 80.3 & 60.0  & 79.1 & 20.3  & 47.2 & 21.2 & \multicolumn{1}{c|}{11.4} & 47.8 \\
			\multicolumn{1}{c|}{TPS{\cite{tps}}}      &      & \checkmark  & \multicolumn{1}{c|}{}     & 82.4 & 36.9  & 79.5  & 9.0   & 26.3  & 29.4 & 78.5  & 28.2  & 81.8 & 61.2  & 80.2 & 39.8  & 40.3 & 28.5 & \multicolumn{1}{c|}{31.7} & 48.9 \\
			\multicolumn{1}{c|}{MoDA{\cite{moda}}}      &      & \checkmark  & \multicolumn{1}{c|}{}     & 72.2 & 25.9  & 80.9  & 18.3   & 24.6  & 21.1 & 79.1  & 23.2  & 78.3 & \textbf{68.7}  & 84.1 & 43.2  & 49.5 & 28.8 & \multicolumn{1}{c|}{38.6} & 49.1 \\
			\multicolumn{1}{c|}{I2VDA{\cite{necess}}}      &      & \checkmark  & \multicolumn{1}{c|}{}     & 84.8 & 36.1 & 84.0  &  \underline{28.0}   &  \underline{36.5}   & 36.0 &  \underline{85.9}   & 32.5  & 74.0 &  63.2  & 81.9 & 33.0  & 51.8 & \textbf{39.9} & \multicolumn{1}{c|}{0.1} & 51.2 \\
			\multicolumn{1}{c|}{SFC{\cite{aaai}}}      &      & \checkmark  & \multicolumn{1}{c|}{}     & 89.9 & 40.8  & 83.8  & 6.8   & 34.4  & 25.0 & 85.1 & \underline{34.3} &  84.1  & 62.6  & 82.1  &35.3  & 47.1 & 23.2 & \multicolumn{1}{c|}{31.3} & 51.1 \\
			\multicolumn{1}{c|}{TPL-SFC{\cite{aaai}}}      &      & \checkmark  & \multicolumn{1}{c|}{}     & 89.9 & 41.5  & 84.0  & 7.0   &  \underline{36.5}   & 27.1 & 85.6 & 33.7 &  \underline{86.6}  & 62.4  & 82.6  &36.3  & 47.6 & 23.2 & \multicolumn{1}{c|}{31.9} & 51.7 \\
			\multicolumn{1}{c|}{PAT{\cite{pat}}}      &      & \checkmark  & \multicolumn{1}{c|}{}     & 85.3 & 42.3  & 82.5  & 25.5   & 33.7  & 36.1 & 86.6 & 32.8 &  84.9  & 61.5  & 83.3  &34.9  & 46.9 & 29.3 & \multicolumn{1}{c|}{29.9} & 53.0 \\
			\multicolumn{1}{c|}{CMOM{\cite{cmom}}}     &      & \checkmark  & \multicolumn{1}{c|}{\checkmark}    & 89.0 & \textbf{53.8}  & 86.8  & \textbf{31.0}  & 32.5  & \textbf{47.3} &  85.6  & 25.1  & 80.4 &  65.1   & 79.3 & 21.6  & 43.4 & 25.7 & \multicolumn{1}{c|}{40.6} & 53.8 \\ \hline
            \multicolumn{1}{c|}{QuadMix}              &      & \checkmark  & \multicolumn{1}{c|}{}     &  \underline{91.4}  &  43.5   & 85.5   & 21.3  & 26.6  & \underline{40.2}  & 84.7  & 34.2  & 84.6 & 62.1  &  84.3  & 39.9  & 46.6 & 33.5 & \multicolumn{1}{c|}{45.6} & 54.9 \\
			\multicolumn{1}{c|}{QuadMix}             &      & \checkmark  & \multicolumn{1}{c|}{\checkmark}    & \textbf{91.6} & \underline{51.4}  & \underline{87.0}  & 24.1  & 32.3  & 37.2 & 84.1  & 28.4  & 84.8 & 64.4  & \underline{85.7} & \underline{41.4}  & 46.5 & 34.0 & \multicolumn{1}{c|}{\underline{49.6}} & \underline{56.2} \\ 
            \multicolumn{1}{c|}{QuadMix (ViT)}             &      & \checkmark  & \multicolumn{1}{c|}{}    & 87.3 & 43.8  & \textbf{87.3}  & 25.2  & \textbf{40.0}  & 36.9 & \textbf{86.7}  & 20.8  & \textbf{90.3} & \underline{65.8}  & \textbf{86.8} & \textbf{48.6}  & \textbf{65.6} & \underline{37.6} & \multicolumn{1}{c|}{\textbf{49.7}} & \textbf{58.2} \\ \toprule[1.0pt]
			
	\end{tabular}}
    \vspace{-10pt}
\end{table*}

\subsubsection{Implementation Details}
Our model is implemented in PyTorch and trained on a single NVIDIA Tesla V100 GPU. The details are as follows: (i) As for the network, we employ a variant of ACCEL~\cite{accel} for video UDA-SS, using DeepLab-V2~\cite{deeplabv2} or MiT-B5 in DAFormer~\cite{daformer} as the segmentation branch, and FlowNet~\cite{flownet} as the optical flow branch. For image UDA-SS, we only employ the segmentation branch. Notably, DeepLab-V2 uses ResNet-101~\cite{resnet} as its backbone, and both DeepLab-V2 and MiT-B5 are pre-trained on ImageNet~\cite{imagenet}; (ii) As for input resolution, for video tasks, we resize the resolution of Cityscapes-Seq, Synthia-Seq, and VIPER to $1024\times512$, $760\times1280$, and $720\times1280$, respectively. While for image tasks, images undergo resizing to dimensions of $1024\times2048$, $1440\times2560$, and $1520\times2560$ for Cityscapes, GTAV, and SYNTHIA; (iii) We set batch size $B$ to 1 for video and 2 for image scenarios; (iv) For DeepLab-V2, we use the SGD optimizer with a momentum of 0.9 and a weight decay of $5\times10^{-4}$. The initial learning rate is $2.5\times10^{-4}$ for Synthia-Seq to Cityscapes-Seq and two image benchmarks, and $1\times10^{-4}$ for VIPER to Cityscapes-Seq. For transformer networks, we use an AdamW optimizer with a weight decay of 0.01 and betas set to (0.9, 0.999). The learning rate is initially set to $6\times10^{-5}$ for the encoder and $6\times10^{-4}$ for the decoder, with a linear warmup followed by linear decay; (v) Our models are trained for a maximum of 40k iterations for four benchmarks; (vi) We apply random Gaussian blur and color jitter to the target domain for image/video frame enhancement; (vii) Aligned with~\cite{davsn,tps,cmom}, we set video temporal deviation $\tau $ to 1; (viii) Additionally, the confidence threshold for target filtering operation $\mathcal{F}$ is 0.9; (ix) During quad-directional mixing, source videos undergo random cropping to match the resolution of target videos; (x) For image tasks, we set ${\lambda _\mathcal{T}}$ and ${\lambda _f}$ to 1 for simplicity; (xi) In this paper, we use per-category IOU (Intersection over Union) and mIOU (mean IOU) as evaluation metrics. Experiment settings are consistent with prior works for fair comparison.

\begin{table*}
	\caption{\textbf{Image UDA-SS} results on GTAV $\rightarrow$ Cityscapes using DeepLab-V2 (DL-V2) and vision transformer (ViT). The best results using ViT are in \textbf{bold}, and the best results using DL-V2 are in \underline{underlined}. 
		\label{tab:tableiss1}}
	\renewcommand{\arraystretch}{1.6}
	\centering
	\small
	\scalebox{0.7}{
		\begin{tabular}{c|c|ccccccccccccccccccc|c}
			\bottomrule[1.0pt]
			\multicolumn{1}{c|}{Methods} & Network & Road & Side. & Buil. & Wall & Fence & Pole & Light & Sign & Vege. & Terr. & Sky  & Pers. & Rider & Car  & Truck & Bus  & Train & Moto. & \multicolumn{1}{c|}{Bike} & mIOU \\ \hline
			
			\multicolumn{1}{c|}{Source-Only}  & DL-V2 & 63.3 & 15.7  & 59.4  & 8.6 & 15.2  & 18.3 & 26.9  & 15.0 & 80.5  & 15.3  & 73.0 & 51.0  & 17.7  & 59.7 & 28.2  & 33.1 & 3.5   & 23.2  & \multicolumn{1}{c|}{16.7} & 32.9 \\
			
			\multicolumn{1}{c|}{PixMatch{\cite{pixmatch}}} & DL-V2& 91.6 & 51.2  & 84.7  & 37.3 & 29.1  & 24.6 & 31.3  & 37.2 & 86.5  &  44.3 & 85.3 & 62.8  & 22.6  & 87.6 & 38.9  & 52.3 & 0.7  & 37.2  & \multicolumn{1}{c|}{50.0} & 50.3 \\
			\multicolumn{1}{c|}{RDA{\cite{rda}}}  & DL-V2    & 93.1  & 55.0  & 84.7  & 33.1 & 29.5  & 38.7 & 49.3   & 44.9 & 84.8  & 41.6  & 80.2 & 62.3  & 33.2  & 85.6 & 37.3  & 51.3 &  18.5   & 34.6  & \multicolumn{1}{c|}{45.3} & 52.8 \\

            \multicolumn{1}{c|}{DACS{\cite{dacs}}}  & DL-V2   & 89.9 & 39.7  & 87.9 & 39.7 & 39.5  & 38.5 & 46.4  & 52.8 &  88.0 & 44.0  & 88.8 & 67.2  &  35.8  & 84.5 & 45.7  & 50.2 & 0.0   & 27.3  & \multicolumn{1}{c|}{34.0} & 52.1 \\
            \multicolumn{1}{c|}{SAC{\cite{sac}}}  & DL-V2    & 90.4 & 53.9  & 86.6  & 42.4 & 27.3  & 45.1  & 48.5  & 42.7 & 87.4  & 40.1  & 86.1 &  67.5  & 29.7  & 88.5 & 49.1  & 54.6 & 9.8   & 26.6  & \multicolumn{1}{c|}{45.3} & 53.8 \\
			\multicolumn{1}{c|}{DSP{\cite{dsp}}}  & DL-V2    & 92.4 & 48.0  & 87.4  & 33.4 & 35.1  & 36.4 & 41.6  & 46.0 & 87.7  & 43.2  &  89.8 & 66.6  & 32.1  & 89.9 & 57.0  & 56.1 & 0.0   & 44.1  & \multicolumn{1}{c|}{57.8} & 55.0 \\

             \multicolumn{1}{c|}{BDM{\cite{bdm}}} & DL-V2     & 91.3 & 51.8  & 86.7  & 49.9 & 49.2  & 53.3 & 43.1  & 43.3 & 85.5  & 47.9  &  85.7 & 62.3  & 45.9  & 87.8 & 55.5  & 54.4 & 4.4   & 46.3  & \multicolumn{1}{c|}{50.4} & 57.6 \\
			\multicolumn{1}{c|}{I2F{\cite{I2f}}}   & DL-V2   & 90.8 & 48.7  & 85.2  & 30.6 & 28.0  & 33.3 & 46.4  & 40.0 & 85.6  & 39.1  & 88.1 & 61.8  & 35.0  & 86.7 & 46.3  & 55.6 & 11.6  & 44.7  & \multicolumn{1}{c|}{54.3} & 53.3 \\ 
           
           \multicolumn{1}{c|}{ADPL{\cite{adpl}}}   & DL-V2   & 93.4 & 60.6  & 87.5  &  45.3  & 32.6  & 37.3 & 43.3  & 55.5 & 87.2  & 44.8  & 88.0 & 64.5  & 34.2  & 88.3 & 52.6  &  61.8 &  49.8 & 41.8  & \multicolumn{1}{c|}{59.4} & 59.4 \\ 
 
             \multicolumn{1}{c|}{SePiCo{\cite{sepico}}} & DL-V2    & 95.2 & 67.8  &  88.7  & 41.4 & 38.4  & 43.4 & 55.5  & 63.2 & 88.6  & 46.4  & 88.3 & 73.1  &  49.0 & 91.4 & 63.2  &  60.4 & 0.0   & 45.2  & \multicolumn{1}{c|}{60.0} & 61.0 \\
              \multicolumn{1}{c|}{FREDOM{\cite{fredom}}} & DL-V2    & 90.9 & 54.1  & 87.8  & 44.1 & 32.6  & 45.2 & 51.4  & 57.1 & 88.6  & 42.6  & 89.5 & 68.8  & 40.0  & 89.7 & 58.4  &  62.6 & \underline{55.3}   & 47.7  & \multicolumn{1}{c|}{40.0} & 61.3 \\
              \multicolumn{1}{c|}{\textbf{QuadMix}} & DL-V2     &  \underline{97.1} &  \underline{78.0} &  \underline{90.4}  & \underline{49.7} &\underline{40.3}    & \underline{53.4}  & \underline{61.7} & \underline{70.9}  & \underline{90.7} & \underline{49.7}  & \underline{92.9} & \underline{77.9}  &  \underline{53.9}   &  \underline{93.5}   & \underline{72.4}  &\underline{65.9}& 0.7   &\underline{60.5}  & \multicolumn{1}{c|}{\underline{68.5}} & \underline{66.8} \\  

            \hline
             \multicolumn{1}{c|}{DAFormer{\cite{daformer}}} & ViT    & 95.7  & 70.2   & 89.4  & 53.5 & 48.1  & 49.6 & 55.8  & 59.4 & 89.9  & 47.9  &  92.5  & 72.2  & 44.7  & 92.3 & 74.5  & 78.2 & 65.1   &  55.9  & \multicolumn{1}{c|}{61.8} & 68.3 \\  
            \multicolumn{1}{c|}{SePiCo{\cite{sepico}}}    & ViT  & 96.7   &  76.7    &  89.7   & 55.5 &  49.5   &  53.2  &  60.0   & 64.5 &  90.2   &  50.3  & 90.8 & 74.5  & 44.2  &  93.3 &  77.0   &  79.5 & 63.6   &    61.0   & \multicolumn{1}{c|}{65.3} &  70.3  \\ 
            \multicolumn{1}{c|}{FREDOM{\cite{fredom}}}  & ViT    & 96.7 & 74.8  & 90.9  & 58.1 & 49.0  & 57.5  & 63.4  &\textbf{71.4}& 91.6  & 52.1   & \textbf{94.4} & 78.4  & 53.1  & 94.1  & 83.9  &  85.2 & 72.5   & 62.8  & \multicolumn{1}{c|}{\textbf{68.9}}& 73.6 \\  
                        
           \multicolumn{1}{c|}{\textbf{QuadMix}}   & ViT    & \textbf{97.5} & \textbf{80.9}   &  \textbf{91.6}  & \textbf{62.3} & \textbf{57.6}  & \textbf{58.2}  & \textbf{64.5}  & 71.2 &  \textbf{91.7}  & \textbf{52.3}  &   94.3   &  \textbf{80.0}   &  \textbf{55.9}   &   \textbf{94.6}   &  \textbf{86.3}   & \textbf{90.5} & \textbf{82.3}   & \textbf{65.1}  & \multicolumn{1}{c|}{68.1} & \textbf{76.1}

   \\ \toprule[1.0pt]
	\end{tabular}}
    \vspace{-10pt}
\end{table*}

\begin{table*}
	\caption{\textbf{image UDA-SS} results on SYNTHIA $\rightarrow$ Cityscapes using DeepLab-V2 (DL-V2) and vision transformer (ViT). The best results using ViT are in \textbf{bold}, and the best results using DL-V2 are in \underline{underlined}. mIOU* denotes 13-category adaptation without three classes marked by *, while mIOU means 16-category adaptation.
		\label{tab:tableiss2}}
	\renewcommand{\arraystretch}{1.6}
	\centering
	\small
	\scalebox{0.75}{
		\begin{tabular}{c|c|cccccccccccccccc|cc}
			\bottomrule[1.0pt]
			\multicolumn{1}{c|}{Methods}  & Network & Road & Side. & Buil. & Wall* & Fence* & Pole* & Light & Sign & Vege. & Sky  & Pers. & Rider & Car   & Bus  & Moto.   & \multicolumn{1}{c|}{Bike} & mIOU  & mIOU* \\ \hline
			
			\multicolumn{1}{c|}{Source-Only}  & DL-V2  & 36.3 & 14.6  & 68.8 & 9.2  & 0.2 & 24.4    & 5.6 & 9.1  & 69.0  & 79.4 & 52.5  & 11.3 & 49.8   & 9.5 & 11.0  & \multicolumn{1}{c|}{20.7} & 33.7  & 29.5 \\
			 
			\multicolumn{1}{c|}{PixMatch{\cite{pixmatch}}} & DL-V2   & 92.5  & 54.6 & 79.8  & 4.8 & 0.1  &  24.1 & 22.8 & 17.8  & 79.4 & 76.5 & 60.8 & 24.7 & 85.7  & 33.5   & 26.4 & \multicolumn{1}{c|}{54.4} & 46.1  & 54.5  \\
			\multicolumn{1}{c|}{RDA{\cite{rda}}}   & DL-V2      & 89.4 & 39.0  & 79.8 & 8.1   & 1.2 & 33.0  & 22.6  & 28.1 & 81.8 & 82.0  & 60.9 & 30.1  & 82.7 &  41.4   & 37.5  & \multicolumn{1}{c|}{48.9} & 47.9  & 55.7 \\
			
            \multicolumn{1}{c|}{DACS{\cite{dacs}}} & DL-V2   & 80.6 & 25.1  & 81.9 & 21.5  & 2.9 &  37.2 & 22.7  & 24.0 & 83.7  &  90.8  & 67.6 & 38.3  & 82.9 & 38.9   & 28.5  & \multicolumn{1}{c|}{47.6} & 48.3  & 54.8 \\
            \multicolumn{1}{c|}{SAC{\cite{sac}}}  & DL-V2     & 89.3 & 47.2  & 85.5 & 26.5  & 1.3 & 43.0  & 45.5  & 32.0 &  87.1  & 89.3  & 63.6 & 25.4  & 86.9 & 35.6   & 30.4  & \multicolumn{1}{c|}{53.0} & 52.6  & 59.3 \\
			
			\multicolumn{1}{c|}{DSP{\cite{dsp}}}  & DL-V2     & 86.4 & 42.0  & 82.0 & 2.1  & 1.8 & 34.0  & 31.6  &  33.2 & 87.2  & 88.5  & 64.1 & 31.9  & 83.8 & 64.4   & 28.8  & \multicolumn{1}{c|}{54.0} & 51.0 & 59.9 \\

             \multicolumn{1}{c|}{BDM{\cite{bdm}}} & DL-V2      & 91.0 & 55.8  & 86.9 & -  & - & -  & 58.3  &  44.7 & 85.8  & 85.7  & 84.1 & 40.3  & 86.0 & 55.2   & 45.0  & \multicolumn{1}{c|}{50.6} & -  & 66.8 \\
			\multicolumn{1}{c|}{I2F{\cite{I2f}}}   & DL-V2   & 84.9 & 44.7  & 82.2 & 9.1  & 1.9 & 36.2  & 42.1  & 40.2 & 83.8  & 84.2  & 68.9 & 35.3  & 83.0 & 49.8  & 30.1  & \multicolumn{1}{c|}{52.4} & 51.8 & 60.1  \\ 
           
           \multicolumn{1}{c|}{ADPL{\cite{adpl}}}  & DL-V2     & 86.1  & 38.6  & 85.9 & 29.7  & 1.3 & 36.6  & 41.3  & 47.2 & 85.0  & 90.4  & 67.5 & 44.3  & 87.4 &  57.1 & 43.9  & \multicolumn{1}{c|}{51.4} & 55.9  & 63.6 \\ 
 
             \multicolumn{1}{c|}{SePiCo{\cite{sepico}}}  & DL-V2     & 77.0 & 35.3  & 85.1 & 23.9  & 3.4 & 38.0  & 51.0  & 55.1 & 85.6  & 80.5   & 73.5 & 46.3  &  87.6 & \underline{69.7}   & 50.9  & \multicolumn{1}{c|}{66.5} & 58.1  & 66.5 \\
              \multicolumn{1}{c|}{FREDOM{\cite{fredom}}}  & DL-V2    & 86.0 & 46.3  & 87.0 & \underline{33.3}  & \underline{5.3} & 48.7  & 38.1  & 46.8 & 87.1  & \underline{89.1}  & 71.2 & 38.1  &  87.1 & 54.6   & 51.3  & \multicolumn{1}{c|}{59.9} & 59.1 & 66.0  \\
               \multicolumn{1}{c|}{\textbf{QuadMix}} & DL-V2      & \underline{88.5} & \underline{52.9}  & \underline{87.1} & 3.7  & 1.4 & \underline{56.2} & \underline{62.7} &   \underline{59.2}  &  \underline{87.2}   & 89.0  &  \underline{79.1}   & \underline{55.8} & \underline{87.9}   & 61.7 & \underline{58.1} & \multicolumn{1}{c|}{\underline{71.2}} & \underline{62.6} & \underline{72.3} \\ \hline
               
             \multicolumn{1}{c|}{DAFormer{\cite{daformer}}}  & ViT  & 84.5 & 40.7  & 88.4 & 41.5  & 6.5 & 50.0  & 55.0  & 54.6 & 86.0  & 89.8  & 73.2 & 48.2  & 87.2 & 53.2   &  53.9  & \multicolumn{1}{c|}{61.7} & 60.9  & 67.4 \\  
           
            \multicolumn{1}{c|}{SePiCo{\cite{sepico}}}   & ViT  & 87.0 & \textbf{52.6}  & 88.5 & 40.6  & \textbf{10.6} & 49.8  & 57.0  & 55.4 & 56.8  & 86.2  & 75.4 & 52.7  & \textbf{92.4}& 78.9   &  53.0  & \multicolumn{1}{c|}{62.6}  & 64.3  & 71.4 \\ 
            \multicolumn{1}{c|}{FREDOM{\cite{fredom}}}   & ViT   & \textbf{89.4} & 50.8  & \textbf{89.3} & \textbf{48.8}  & 9.3 & 57.3  & \textbf{65.1}  & 60.1 & \textbf{89.9}  & 93.7  & 79.4 & 51.6  & 90.5 & 66.0   &  62.3  & \multicolumn{1}{c|}{\textbf{68.1}} & 67.0  & 73.6\\

			\multicolumn{1}{c|}{\textbf{QuadMix}}  & ViT     & 88.1 & 51.2  & 88.9 & 46.7  & 7.9 & \textbf{58.6} & 64.7  & \textbf{63.7} & 88.1  &  \textbf{93.9}  &  \textbf{81.3} & \textbf{56.6}  & 90.3 & \textbf{66.9}   & \textbf{66.8} & \multicolumn{1}{c|}{66.0} & \textbf{67.5}  & \textbf{74.3}
   \\ \toprule[1.0pt]
	\end{tabular}}
    \vspace{-10pt}
\end{table*}

\subsection{Experimental Results}

\subsubsection{Comparison With State-of-the-Arts of Video UDA-SS}
In Table~\ref{tab:table1} and Table~\ref{tab:table2}, we compare our method with the state-of-the-art methods on Synthia-Seq $\rightarrow$ Cityscapes-Seq and VIPER $\rightarrow$ Cityscapes-Seq benchmark. To the best of our knowledge, recent methods~\cite{davsn,necess,tps,moda,aaai,pat,cmom} are the most superior works in video UDA-SS, all employing CNN-based architectures. Notably, for the first time, we explore \textbf{transformer-based architectures} for video UDA-SS. Moreover, we also compare several variants to video UDA-SS that are initially designed for image UDA-SS, including works~\cite{advent,cbst,ida,crst,svmin,crcda,rda,fda,sac,dacs,pixmatch}. Technically, following the previous video UDA-SS works~\cite{davsn,necess,tps,cmom}, we replace their image segmentation backbones with a video segmentation network (denoted by \textsuperscript{\dag}). Additionally, ``Adv." signifies methods that employ adversarial learning, and ``SSL" denotes self-supervised learning. ``Warm" denotes initializing parameters of the DeepLab-V2 model with a warm-up model, whereas others use ResNet-101 pre-trained on ImageNet. ``ViT" refers to the transformer backbone.

\textbf{Synthia-Seq $\rightarrow$ Cityscapes-Seq.} In Table~\ref{tab:table1}, we present experimental results comparing our method with state-of-the-art techniques on the Synthia-Seq $\rightarrow$ Cityscapes-Seq. For clarity, the best per-category IOU and mIOU results are in bold, while suboptimal results are underlined. We observed that: (i) Our method is
highly compatible with both CNN and transformer-based architectures. The proposed CNN-based Quadmix significantly outperforms the previous best CMOM~\cite{cmom} by 5.2$\%$ mIOU. When equipped with a transformer, we further boost mIOU by 5.7$\%$, achieving the highest result of 67.2$\%$; (ii) For CNN-based QuadMix, without a warm-up model, we achieved 59.6$\%$ mIOU, which is 4.3$\%$ higher than TPL-SFC~\cite{aaai} by a large margin. When using a warm-up model, we outperform the CMOM, which relies heavily on the warm-up model in~\cite{davsn} for both model initialization and pseudo-label generation; (iii) Adversarial learning methods can facilitate domain adaptation but are less improving compared to self-supervised methods; (iv) Our method boosts the IOU for most categories. It can be attributed to our comprehensive quad-directional mixing, which bridges the feature gap effectively and addresses the intra-domain discontinuity and feature inconsistencies, as well as flow-guided video feature aggregation for cross-domain alignment.

\textbf{Viper $\rightarrow$ Cityscapes-Seq.} The domain gap between Viper and Cityscapes-Seq is much larger than that of Synthia-Seq and Cityscapes-Seq. As shown in Table~\ref{tab:table2}, our proposed QuadMix also achieves state-of-the-art results. We can conclude that: (i) For CNN-based QuadMix, we achieve 56.2$\%$ mIOU with a warm-up model and 54.9$\%$ mIoU without it; furthermore, we obtain 58.2$\%$ mIOU when we employ transformer-based network. All our results significantly outperform previous methods; (ii) Recent work I2VDA~\cite{necess} focuses exclusively on transferring source spatial information to target domain videos, resulting in a target domain mIOU of 51.2$\%$. This is inferior to our model, which comprehensively investigates both spatial and temporal level domain shifts. (iii) Variants of image domain adaptation methods generally underperform on target videos compared to video UDA-SS methods, likely due to inadequate utilization of crucial temporal information.

\begin{figure*}[!t]
	\centering
	\includegraphics[width=0.99\linewidth]{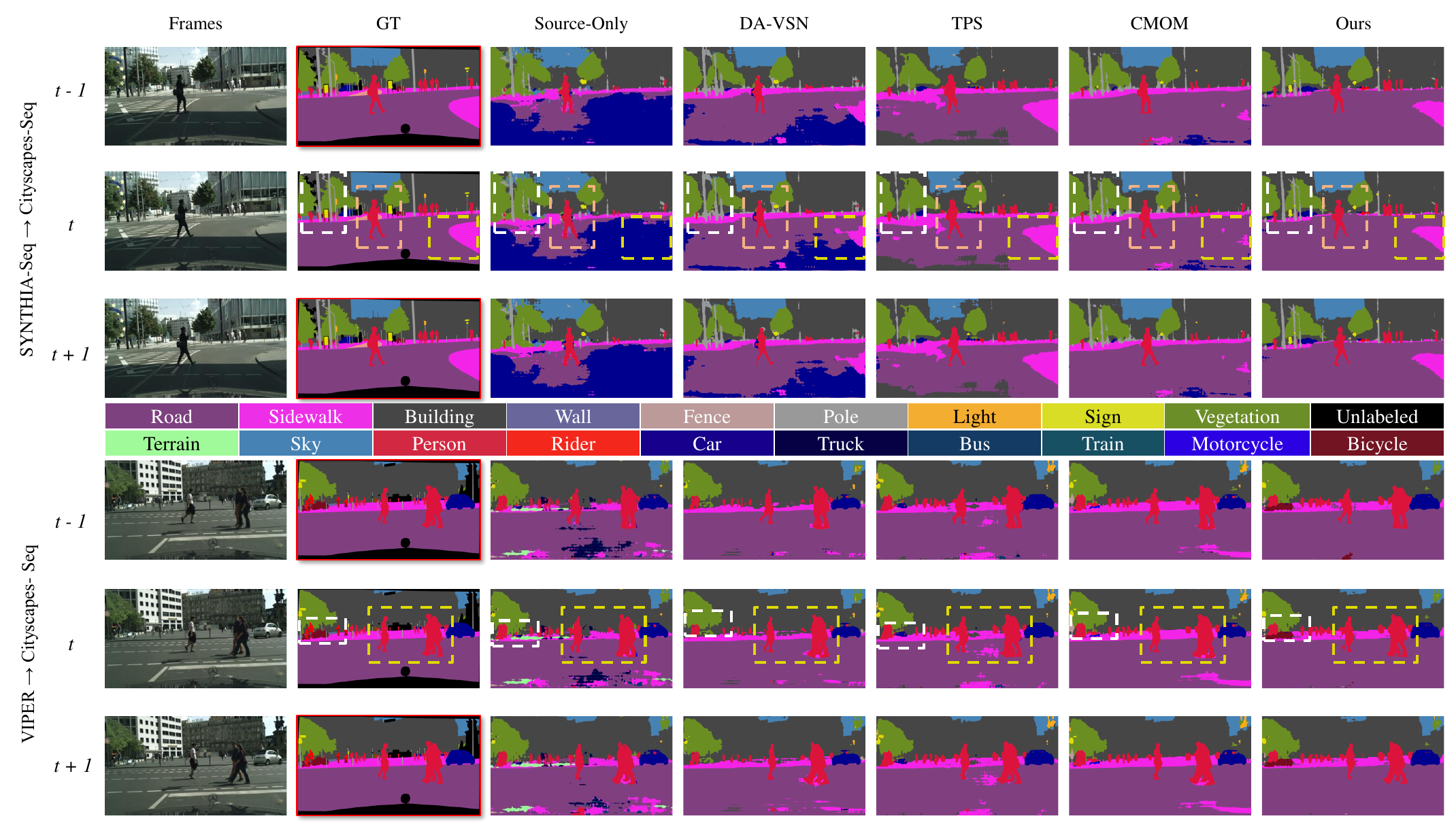}
        \vspace{-5pt}
	\caption{Qualitative results for three consecutive frames from the Cityscapes-Seq validation set. The results are presented in the following order: target image, ground truth, semantic segmentation results from the source-only model, DA-VSN~\cite{davsn}, TPS~\cite{tps}, CMOM~\cite{cmom}, and the proposed method. We achieve the highest segmentation accuracy, exhibiting smoother edges and refined details. Please zoom in for details.
	}
	\label{fig. 7}
    \vspace{-10pt}
\end{figure*}

\textbf{Qualitative results comparison on two benchmarks.}
In Fig.~\ref{fig. 7}, we present adjacent three frames and corresponding qualitative segmentation results of our method along with the source-only model, DA-VSN~\cite{davsn}, TPS~\cite{tps}, and CMOM~\cite{cmom} on Cityscape-Seq validation dataset. Notably, it only offers a ground-truth label for the $19$-th frame of each 30-frame video clip, so we replicate the label from frame $t$ for frames $t-1$ and $t+1$, marked with red borders for clarity. Our results exhibit smoother edges and refined details. For example, on Synthia-Seq $\rightarrow$ Cityscapes-Seq benchmark, we obtain more accurate segmentation for categories like \textit{person}, \textit{sign}, and \textit{sidewalk}. Similarly, in the case of Viper $\rightarrow$ Cityscapes-Seq, our model predicts contours that closely resemble actual \textit{bicycle} and \textit{person}, indicating a significant improvement. Please refer to the \href{https://drive.google.com/file/d/1OT5GtsbC0CcW6aydBL27ADjve95YE5oj/view?usp=sharing}{supplementary video} for more visual comparisons.

\subsubsection{Comparison With State-of-the-Arts of Image UDA-SS}
For image UDA-SS, we replace spatio-temporal QuadMix with spatial QuadMix, and substitute flow-guided spatio-temporal feature aggregation with spatial feature aggregation. Technically, we shift elements of $Z_{t,n}^{\mathcal{S^*},k}$ and $Z_{t,n}^{\mathcal{T^*},k}$ for images defined in Section~\ref{subsec:322}, and remove the reliance on temporal cues that are not present in image scenarios. Additionally, target pseudo-labels are generated using a momentum network~\cite{mean}.

Table~\ref{tab:tableiss1} and Table~\ref{tab:tableiss2} compare our method with existing image UDA-SS approaches. Impressively, we maintain superiority on two challenging benchmarks: GTAV $\rightarrow$ Cityscapes and SYNTHIA $\rightarrow$ Cityscapes. We compare with recent advanced CNN-based methods~\cite{pixmatch,rda,dacs,sac,dsp,bdm,I2f,adpl,sepico,fredom} and transformer-based methods~\cite{daformer,sepico,fredom}. Notably, DACS~\cite{dacs} proposes one-way mixing, which is followed by subsequent works~\cite{daformer,sepico,fredom} combing contrastive learning or long-tail learning. Other works~\cite{bdm,adpl} employ two-way mixing, with the latter integrating style transfer. However, as analyzed above, they overlooked the intra-domain discontinuity within each domain, leading to fragmented and insufficient domain gap bridging. Additionally, their focus on pixel-level mixing alone can result in feature inconsistencies, hindering effective knowledge transfer. In contrast, we address these challenges comprehensively and obtain remarkable results.

\textbf{GTAV $\rightarrow$ Cityscapes.} 
As shown in Table~\ref{tab:tableiss1}, our method achieves mIOUs of 66.8$\%$ with DeepLab-V2 and 76.1$\%$ with transformer networks. Notably, QuadMix (ViT) more effectively extracts spatial features from long-tail categories. Specifically, for the \textit{train} class, our method achieves an IOU of 82.3$\%$, surpassing that of FREDOM~\cite{fredom} (72.5$\%$), which is tailored for long-tail learning. Notably, QuadMix (DL-V2) underperforms for the \textit{train} class due to the limited feature extraction capability of CNNs. We believe that more frequent resamples for models to learn such categories could improve the results. Generally, our method boosts the performance of most categories, exhibiting robust improvement.

\textbf{SYNTHIA $\rightarrow$ Cityscapes.} In Table~\ref{tab:tableiss2}, we present results for 13-category (mIOU*) and 16-category (mIOU) adaptation, following~\cite{sepico, adpl}, etc. BDM~\cite{bdm} only studies 13-category adaption. Note that QuadMix (DL-V2) has already outperformed DAFormer (ViT) by 1.7$\%$ in 13-category adaptation and 4.9$\%$ in 16-category adaptation, with further improvements of 4.9$\%$ and 2.0$\%$, respectively, when equipped with ViT. Overall, our method consistently demonstrates superior performance and is well-adapted for transformers and CNNs.

For a qualitative comparison with previous works, please refer to the supplementary material.

\subsection{Ablation Study}
\label{subsec:ablation}
Given that an image is essentially a special case of a video without the temporal dimension, we opted for the more general video UDA-SS for ablation studies, using the DeepLab-V2 backbone by default.

\begin{table}
	\caption{Ablation studies on Synthia-Seq $\rightarrow$ Cityscapes-Seq. All models are trained end-to-end for a total of 40k iterations. ``V-template" represents our video patch template as defined in Eq.~\eqref{eq:z_tmpl}. ``F-template" denotes feature-level template mixing across the generated quad-mixed domains, ``Agg." refers to the optical-guided video feature aggregation, ``Warm" means initializing model parameters using a warm-up model.
		\label{tab:table3}}
	\centering
	\renewcommand\arraystretch{1.45}
	\setlength{\tabcolsep}{2.3mm}
	\scalebox{1.00}{
		\begin{tabular}{l|cccc|c}
			\bottomrule[1.0pt]
			
			ID &  V-template & F-template & Agg. & Warm & mIOU (Gain) \\ \hline
			$0^*$                                      &                    &                                          &                   &         & 52.01     \\
			1                                          & \checkmark                  &                                           &                   &         & 55.71 (+3.70) \\
			2                                       & \checkmark                 & \checkmark                                    &                   &         & 57.25 (+5.24) \\
			
			3                                        &                    &                                           & \checkmark                 &         & 54.01 (+2.00) \\
			4                                        & \checkmark                  & \checkmark                                     & \checkmark                 &         &  59.62 (+7.61) \\ \hline

			5                                         & \checkmark                  & \checkmark                                   &                   & \checkmark       & 57.95 (+5.94) \\
			6                                         &                    &                                          & \checkmark                 &   \checkmark      & 55.97 (+3.96)\\
			
			7                                       & \checkmark                  &                                      &  \checkmark                 &     \checkmark    & 60.26 (+8.25) \\
			8                                        & \checkmark                  & \checkmark                                 & \checkmark                 & \checkmark       & \textbf{61.48} (\textbf{+9.47}) \\ \toprule[1.0pt]
	\end{tabular}}
    \vspace{-10pt}
\end{table}

\subsubsection{Ablation Study on Each Component}
In Table~\ref{tab:table3}, we conduct ablation studies to verify the effectiveness of three key components: video patch template (V-template) mixing, feature-level template (F-template) mixing, and optical-guided video feature aggregation (Agg.). Additionally, we also study the impact of the warm-up mentioned in Table~\ref{tab:table1} and Table~\ref{tab:table2}. For a fair comparison, we take the result of TPS~\cite{tps} without random scale augmentation as the baseline (Exp. ID~$0^*$).

It can be concluded that: (i) Video patch template (V-template) $Z_{t,n}^{*,k}$ result in a 3.70$\%$ mIOU gain in Exp. ID~1; (ii) As for feature-level template mixing (F-template) across two enhanced quad-mixed domains, Exp. ID~2 further incorporates the F-template and demonstrates a 1.54$\%$ mIOU gain. In addition, Exp.~8 outperforms Exp.~7 by 1.22$\%$, also validating the efficacy of ``F-template"; (iii) In another line, Exp.~3 introduces the optical-guided video feature aggregation (Agg.), resulting in growth to 54.01$\%$; (iv) Furthermore, Exp. ID~4 combines ``V-template", ``F-template", and ``Agg.", achieving an mIOU of 59.62$\%$, surpassing the state-of-the-art TPL-SFC~\cite{aaai} mIOU of 55.3$\%$ by 4.32$\%$; (v) Moreover, we also employ the warm-up setup in Exp. ID~5 to Exp. ID~8 for model parameter initialization, exceeding the state-of-the-art CMOM of 56.31$\%$ with a warm-up stage by 5.17$\%$, resulting in the highest mIOU of 61.48$\%$, a remarkable 9.47$\%$ increase compared to the baseline Exp. ID~$0^*$.

\textbf{Further comparison.} To valid the video patch templates (V-template) $Z_{t,n}^{*,k}$ in Exp. ID~1, we perform a variant using image patch templates (I-template) akin to~\cite{bdm,dacs}. Thus, the model input is also substituted with $x_t^{\mathcal{S'}}$ and $x_t^{\mathcal{T'}}$ in Eq.~\eqref{eq:loss_QuadMix}. It is observed that with ``I-template", the mIOU dropped by 1.37$\%$ compared with Exp. ID~1, which suggests that ``V-template" enhances the video adaptation more effectively.

\begin{figure*}[!t]
	\centering
	\includegraphics[width=\linewidth]{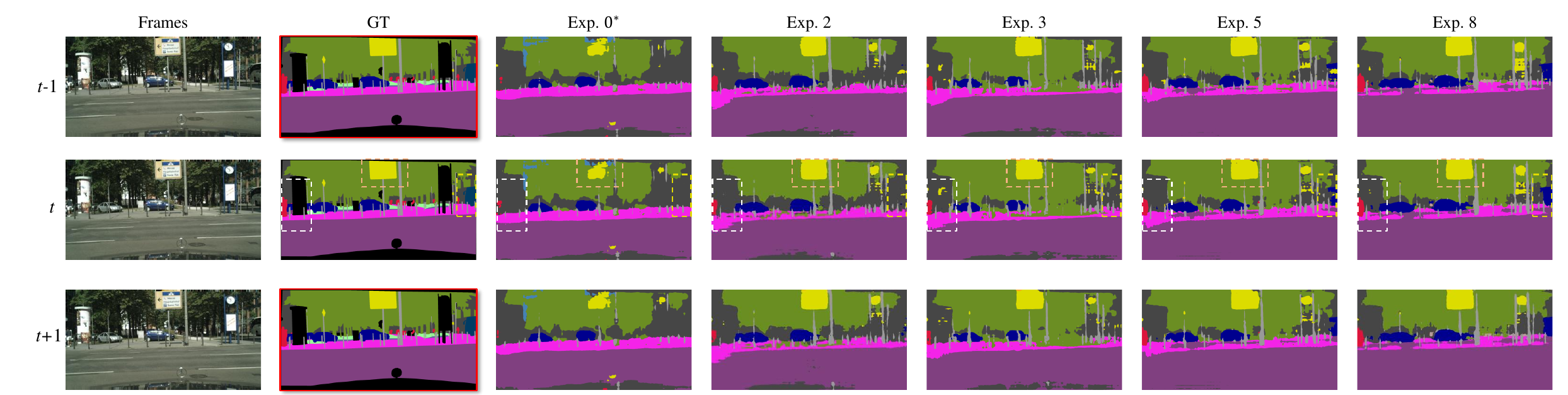}
        \vspace{-12pt}
	\caption{Qualitative results of ablation studies in Table~\ref{tab:table3} for three consecutive frames on Synthia-Seq $\rightarrow$ Cityscapes-Seq. The best results are obtained by integrating all of the proposed components (Exp. ID~8), especially for categories of \textit{person}, \textit{rider}, and \textit{car}. Please zoom in for details.}
	\label{fig. 8}
    \vspace{-10pt}
\end{figure*}

\textbf{Qualitative results of each component.} In Fig.~\ref{fig. 8}, we present segmentation results of three adjacent frames on Synthia-Seq $\rightarrow$ Cityscapes-Seq, following experiment settings corresponding to Exp. ID~0*, ID~2, ID~3, ID~5, ID~8 in Table~\ref{tab:table3}. These results demonstrate the incremental integration of each component. The segmentation results progressively improve from left to right, particularly for categories like \textit{sign} and \textit{car}. Notably, the challenging issue with a \textit{person} on the far left within the frame is effectively addressed by integrating the proposed key components (Exp. ID~8).
 
\subsubsection{Ablation Study on Quad-Directional Mixing}
We evaluate the effects of QuadMix paths through quantitative experiments and detailed feature visualization.

\textbf{The effectiveness of quad-directional mixing paths.}
In this paper, we propose four comprehensive mixing paths designed for sufficient domain gap bridging. It naturally prompts us to explore the outcomes when only the intra/inter-mixed source or target domains are constructed or when templates are exclusively from the source or target domain. Table~\ref{tab:table6} provides thorough ablation results for these scenarios. For fairness, we employ the flow-guided feature aggregation module (i.e., Exp. ID~6 in Table~\ref{tab:table3}) as the baseline (Exp. ID~$0^*$).

\begin{table}[]
	\caption{Ablation studies of the quad-directional mixing paths on Synthia-Seq $\rightarrow$ Cityscapes-Seq.
		\label{tab:table6}}
	\renewcommand\arraystretch{1.45}
	\centering
	\setlength{\tabcolsep}{0.9mm}
	\scalebox{0.9}{
		\begin{tabular}{c|cccccc|c}
			\bottomrule[1.0pt]
			ID & $\mathcal{S}$$\rightarrow$$\mathcal{T}$ & $\mathcal{T}$$\rightarrow$$\mathcal{S}$ & $\mathcal{S}$$\rightarrow$$\mathcal{S}$ & $\mathcal{T}$$\rightarrow$$\mathcal{T}$  & $\mathcal{S}$$\rightarrow$$(\mathcal{T}$$\rightarrow$$\mathcal{T})$ & $\mathcal{T}$$\rightarrow$$(\mathcal{S}$$\rightarrow$$\mathcal{S})$ &  mIOU (Gain)  \\ \hline
						
			CMOM                            & \checkmark                                                   &                                                                             &                                                                             &                                                                             &                                                                                                                                         &                                                                                                                                         & 56.31 \\ \hline
			$0^*$                                &                                                    &                                                                             &                                                                             &                                                                             &                                                                                                                                         &                                                                                                                                         & 55.97 \\
			
			1                               & \checkmark                                                   &                                                                             &                                                                             &                                                                             &                                                                                                                                         &                                                                                                                                         & 58.11 (+2.14) \\
			2                              &                                                                             & \checkmark                                                   &                                                                             &                                                                             &                                                                                                                                         &                                                                                                                                         & 57.84 (+1.87) \\
			3                               & \checkmark                                                   & \checkmark                                                   &                                                                             &                                                                             &                                                                                                                                         &                                                                                                                                         & 58.77 (+2.80) \\ \hline
			4                              &                                                                             &                                                                             & \checkmark                                                   &                                                                             &                                                                                                                                         &                                                                                                                                         & 58.98 (+3.01) \\
			5                               &                                                                             &                                                                             &                                                                             & \checkmark                                                   &                                                                                                                                         &                                                                                                                                         & 56.67 (+0.70) \\
			6                             &                                                                             &                                                                             & \checkmark                                                   & \checkmark                                                   &                                                                                                                                         &                                                                                                                                         & 59.24 (+3.27) \\ \hline
			7                               & \checkmark                                                   &                                                                             &                                                                             & \checkmark                                                   &                                                                                                                                         &                                                                                                                                         & 58.39 (+2.42) \\
			8                              &                                                                             & \checkmark                                                   & \checkmark                                                   &                                                                             &                                                                                                                                         &                                                                                                                                         & 59.55 (+3.58) \\
			9                             & \checkmark                                                   &                                                                             & \checkmark                                                   &                                                                             &                                                                                                                                         &                                                                                                                                         & 60.20 (+4.23) \\
			10                             &                                                                             & \checkmark                                                   &                                                                             & \checkmark                                                   &                                                                                                                                         &                                                                                                                                         & 59.39 (+3.42) \\ \hline
			11                           & \checkmark                                                   & \checkmark                                                   & \checkmark                                                   & \checkmark                                                   &                                                                                                                                         &                                                                                                                                         & 60.39 (+4.42) \\
			12                            &                                                                             &                                                                             &                                                                             &                                                                             & \checkmark                                                                                                               & \checkmark                                                                                                               & \textbf{61.48 (+5.51)} \\ \toprule[1.0pt]
	\end{tabular}}
    \vspace{-10pt}
\end{table}

It can be observed: (i) In Exp. ID~1, our $\mathcal{S}$$\rightarrow$$\mathcal{T}$ setting outperformed the CMOM~\cite{cmom} by 1.80$\%$ in mIoU, which also follows a source to target pasting paradigm. We attribute this improvement to the proposed flow-guided feature aggregation for domain alignment; (ii) The construction of intra/inter-mixed video domains always yields better results than solely one-way mixed domains. It is because the model can learn consistent spatio-temporal features with diverse domain contexts, with the feature-level mixing across quad-mixed domains alleviating feature inconsistencies; (iii) Utilizing source domain templates commonly outperforms using target domain templates with higher predictive confidence. The overall mIOU exhibits further improvement when incorporating templates from both domains; (iv) More intriguingly, compared with incrementally combining various directional mixing as in Exp. ID~11, we not only reduce computational load significantly as shown in Table~\ref{tab:table10}, but also achieve a mIoU gain of 0.92$\%$ to 61.48$\%$ by fully utilizing the enhanced quad-mixed domains for comprehensive domain bridging (Exp. ID~12).

It's noteworthy that regardless of the intra/inter-mixing strategy adopted, the results of ablation studies in Table~\ref{tab:table6} \texttt{all surpass} previous state-of-the-art methods CMOM~\cite{cmom} and TPL~\cite{aaai}, which highlights the effectiveness of our novel and unified perspective that has been largely overlooked by prior works~\cite{cmom,tps,dacs,dsp,bdm}, etc.

\begin{figure*}[!t]
	\centering
	\centering
	\includegraphics[width=0.99\linewidth]{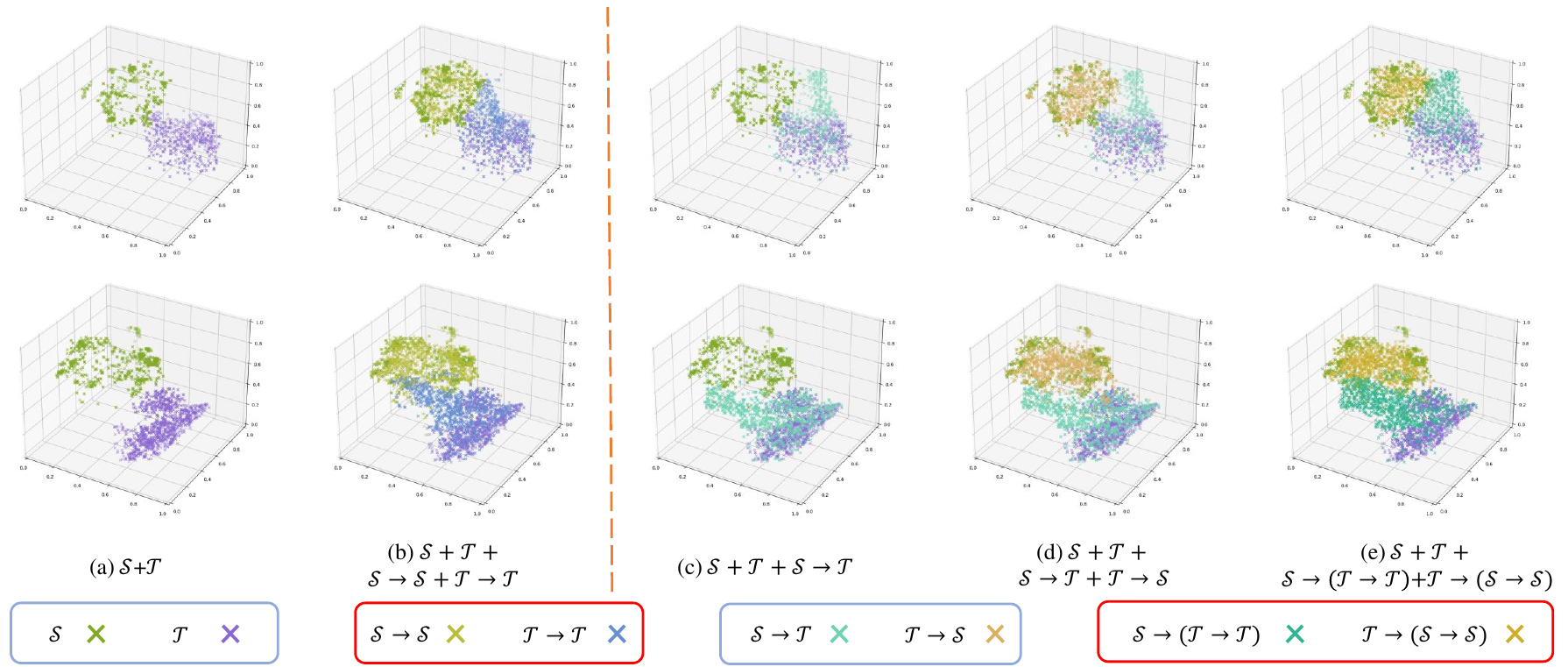}
    \vspace{-5pt}
	\caption{T-SNE visualization of different mixing paradigms using \textit{sign} (line one) and \textit{pole} (line two) as examples. The experimental settings for subfigures (a) to (e) correspond to ablation studies Exp. ID~0*, ID~6, ID~1, ID~3, and ID~12  in Table~\ref{tab:table6}. Note that the number of points in (b), (d), and (e) is equal. Subfigure (b) shows intra-mixed domains with continuous feature embeddings, while (e) depicts quad-mixed domains with a more generalized feature distribution, rather than a clustered one, which better addresses intra-domain discontinuity and fragmented gap bridging for knowledge transfer. Please zoom in for details.}
	\label{fig. 12}
    \vspace{-12pt}
\end{figure*}

\textbf{Feature visualization for quad-directional mixing paths.}
In Fig.~\ref{fig. 12}, following experimental setings in Table~\ref{tab:table6}, we present t-SNE~\cite{tsne} feature visualization for pure two domains (Exp. ID~0*), our \texttt{intra-mixed} domains (Exp. ID~6), one-way mixing (Exp. ID~1), two-way mixing (Exp. ID~3), and the proposed \texttt{QuadMix} (Exp. ID~12) from left to right. Notably, the number of points in Fig.~\ref{fig. 12}(b)(d)(e) are equal. It can be found that quad-directional mixing effectively generalizes intra-domain features and enhances inter-domain features, addressing the insufficient feature gap bridging across distinct domains. By bridging domain gaps sufficiently, it promotes dense knowledge transfer from the source to the target domain.

\begin{table}[t]
	\caption{Ablation studies on category spaces for video patch template generation on Synthia-Seq $\rightarrow$ Cityscapes-Seq.\label{tab:table5}}
	\centering
	\renewcommand\arraystretch{1.5}
	\setlength{\tabcolsep}{1.2mm}
	\scalebox{0.98}{
		\begin{tabular}{c|ccccc}
			\bottomrule[1.0pt]
			Mixed categories & Things & Stuffs & Movable & Stationary & All   \\ \hline
			mIOU(\%)                             & 58.53  & 57.50  & 58.29   & 55.55      & \textbf{61.48} \\ \toprule[1.0pt]
	\end{tabular}}
    \vspace{-15pt}
\end{table}

\textbf{The effect of category selection on video patch template generation.} The video patch templates used for quad-mixing involve patch regions corresponding to the randomly selected two categories from the source or target domain. We investigated the impact of category selection on generating video patch templates. Drawing on the concept from panoramic segmentation~\cite{quanjing}, objects are often categorized into ``things" (e.g., \textit{sign}, \textit{car}) and ``stuff" (e.g., \textit{sky}, \textit{road}). Therefore, in Table~\ref{tab:table5}, we examined the results when the category selection comprised only ``things" or ``stuff". Moreover, we also explore the movable (e.g., \textit{person}, \textit{car}) or stationary (e.g., \textit{sign}, \textit{light}) objects. The experimental findings indicate that ``things" and movable categories yield better results than ``stuff" and stationary categories, and the model performs best when the category selection encompasses all categories, which highlights the versatility of our method across different categories.

\begin{table}[t]
	\caption{Ablation studies on flow-guided feature aggregation for alignment, and image enhancement $\mathcal{A}$ on the target domain.
		\label{tab:table7}}
	\centering
	\renewcommand\arraystretch{1.45}
	\setlength{\tabcolsep}{1.82mm}
	\scalebox{0.95}{
		\begin{tabular}{c|c|c}
			\bottomrule[1.0pt]
			Method                                        & SYN-Seq $\rightarrow$ City-Seq & VIPER $\rightarrow$ Citys-Seq \\ \hline
			w/o align    & 57.95                        & 52.80                  \\ 
			global align    & 58.24                        & 53.62                  \\
			flow-guided aggregation & \textbf{61.48}                       & \textbf{56.16}                 \\ \hline w/o $\mathcal{A}$                                        & 60.23                        & 55.25                  \\ 
			w/ $\mathcal{A}$                                         & \textbf{61.48}             &  \textbf{56.16}                 \\ \toprule[1.0pt]
		\end{tabular}
	}
    \vspace{-10pt}
\end{table}

\subsubsection{Ablation Study on Flow-Guided Spatio-Temporal Feature Aggregation Module}
 In this paper, to further address spatio-temporal gaps across video domains, we propose the optical flow-guided video feature aggregation module for fine-grained alignment. When exclusively substituting our fine-grained aggregation with treating video features as a whole~\cite{dsp}, the mIOUs on two benchmarks decreased by 3.24$\%$ and 2.54$\%$, respectively, as shown in Table~\ref{tab:table7}. It demonstrates the effectiveness of our method in compressing video features across domains into a more aligned distribution, thereby reducing the domain discrepancy with precise alignment. In addition, we also investigate the effect of image enhancement $\mathcal{A}$ in Eq.~\eqref{eq:loss_QuadMix} and Eq.~\eqref{eq:loss_ssl} for target domain samples. 
 
 Note that our contribution focuses on enhancing spatio-temporal coherence of aggregated features for domain alignment, which alignment loss is optimal to minimize discrepancies of aggregated features across domains is another research branch and goes beyond the main consideration of this work.

\begin{table}[t]
	\caption{Loss weights sensitivity analysis on Viper $\rightarrow$ Cityscapes-Seq.\label{tab:table8}}
	\centering
	\renewcommand\arraystretch{1.4}
	\setlength{\tabcolsep}{7mm}
	\scalebox{1.00}{
		\begin{tabular}{cc}
			\bottomrule[1.0pt]
			Weights          & mIoU(\%) \\ \hline
			${\lambda _\mathcal{T}}$,  ${\lambda _f}$ = 0.10, 0.01 & 54.56   \\
			${\lambda _\mathcal{T}}$,   ${\lambda _f}$ = 0.20, 0.01 & \textbf{56.16}   \\
			${\lambda _\mathcal{T}}$,   ${\lambda _f}$ = 0.50, 0.01 & 54.49   \\
			${\lambda _\mathcal{T}}$,  ${\lambda _f}$ = 0.50, 0.10  & 54.56   \\
			${\lambda _\mathcal{T}}$,   ${\lambda _f}$ = 0.70, 0.10 & 54.58   \\
			${\lambda _\mathcal{T}}$,  ${\lambda _f}$ = 1.00, 0.10  & \underline{55.21}   \\ \toprule[1.0pt]
	\end{tabular}}
    \vspace{-10pt}
\end{table}

\begin{table}[t]
	\caption{Effectiveness comparison on model initialization. The ``Original" column denotes the results of original works, while the ``QuadMix" column means the results of our models initialized with warm-up models from the ``Original" column.
		\label{tab:table9}}
	\renewcommand\arraystretch{1.4}
	\centering
	\scalebox{1.00}{
		\begin{tabular}{ccccc}
			\bottomrule[1.0pt]
			\multicolumn{1}{c|}{} & \multicolumn{2}{c|}{SYN-Seq $\rightarrow$ City-Seq}  & \multicolumn{2}{c}{VIPER $\rightarrow$ Citys-Seq}                                                                              \\ \hline
			\multicolumn{1}{c|}{Method}         & Original & \multicolumn{1}{c|}{QuadMix}  & Original & QuadMix \\ \hline
			\multicolumn{1}{c|}{ResNet-101~\cite{resnet}} & 32.01    & \multicolumn{1}{c|}{59.62}           & 24.60 & 54.94    \\
			\multicolumn{1}{c|}{DA-VSN~\cite{davsn}}         & 49.53        & \multicolumn{1}{c|}{59.07}               & 47.84   & 55.31     \\
			
			\multicolumn{1}{c|}{CMOM~\cite{cmom}}           & 56.31   & \multicolumn{1}{c|}{60.33}         & 53.81  & 56.10 \\ 
			\multicolumn{1}{c|}{TPS~\cite{tps}}            & 53.76    & \multicolumn{1}{c|}{\textbf{61.48}}           & 48.91    & \textbf{56.16} \\
			\toprule[1.0pt]
	\end{tabular}}
    \vspace{-10pt}
\end{table}

\begin{table}[t]
	\caption{Throughput is measured with a batch size of 1. Results are obtained using a V100-32G GPU. ``$\text{QuadMix*}$'' corresponds to the ablation setting ID 11 in Table~\ref{tab:table6}.
		\label{tab:table10}}
	\setlength{\tabcolsep}{0.25mm}
	\renewcommand\arraystretch{1.4}
	\centering
	\scalebox{1.00}{
		\begin{tabular}{cccc|cc|cc}
			\bottomrule[1.0pt]
			\multicolumn{4}{c|}{Modules} & \multicolumn{2}{c|}{SYN-Seq$\rightarrow$City-Seq}                         & \multicolumn{2}{c}{VIPER$\rightarrow$City-Seq}                                \\ \hline
			Src.    &   QuadMix  & $\text{QuadMix*}$   & Agg.   & \makecell{Time\\(s/iter)} & \makecell{GPU\\memory}   & \makecell{Time\\(s/iter)} & \makecell{GPU\\memory} \\ \hline
			\checkmark       &      &         &        & 1.40                                                             & 8.78 G   & 1.38                                                             & 8.78 G   \\
			
      & \checkmark  &        &        & 3.78                                                             & 21.24 G & 3.99                                                             & 21.90 G \\
			   & \checkmark   &       & \checkmark      & 4.26                                                             & 24.92 G & 4.37                                                             & 25.90 G \\ \hline
   
      &    &    \checkmark   &     \checkmark    & 5.64                                                             & 29.11 G & 5.89                                                             & 29.85 G \\\toprule[1.0pt]
	\end{tabular}}
    \vspace{-10pt}
\end{table}  
\subsection{Further Analysis}
We conduct further analysis on the general task of video UDA-SS, using the DeepLab-V2 backbone by default.

\subsubsection{Hyperparameter Sensitivity} 
For model optimization, we utilize two hyperparameters, ${\lambda _\mathcal{T}}$ in Eq.~\eqref{eq:loss_QuadMix} and Eq.~\eqref{eq:loss_ssl} and ${\lambda _f}$ in Eq.~\eqref{eq:loss_Agg}. In Table~\ref{tab:table8}, we progressively increase the values of ${\lambda _\mathcal{T}}$ and ${\lambda _f}$ using Viper $\rightarrow$ Cityscapes-Seq as an example to explore the hyperparameter sensitivity. The model achieves an mIOU of 56.16$\%$ when these parameters are set to 0.20 and 0.01 and a suboptimal mIOU of 55.21$\%$ when the two parameters are set to 1.00 and 0.10. The results display slight fluctuations with parameter setting changes, showing our method's robustness. For convenience, we set ${\lambda _\mathcal{T}}$ and ${\lambda _f}$ to 1 for another Synthea-Seq $\rightarrow$ Cityscapes-Seq benchmark.

\subsubsection{Results with Model Initialization}
As shown in Table~\ref{tab:table9}, we achieve 59.62$\%$ and 54.94$\%$ mIOU on two benchmarks by end-to-end training, surpassing previous methods trained both end-to-end and stage-wise. Initializing our models with warm-up parameters from previous works in the ``Original" column can further enhance adaptation. Notably, CMOM~\cite{cmom} initializes model parameters and generates initial pseudo-labels offline using a warm-up model trained in \cite{davsn}. In contrast, we solely undergo model
initialization. The above analysis underscores the remarkable adaptation of our method on video UDA-SS.

\begin{figure}[!t]
	\centering
	\subfloat[Source]{\includegraphics[width=.15\columnwidth]{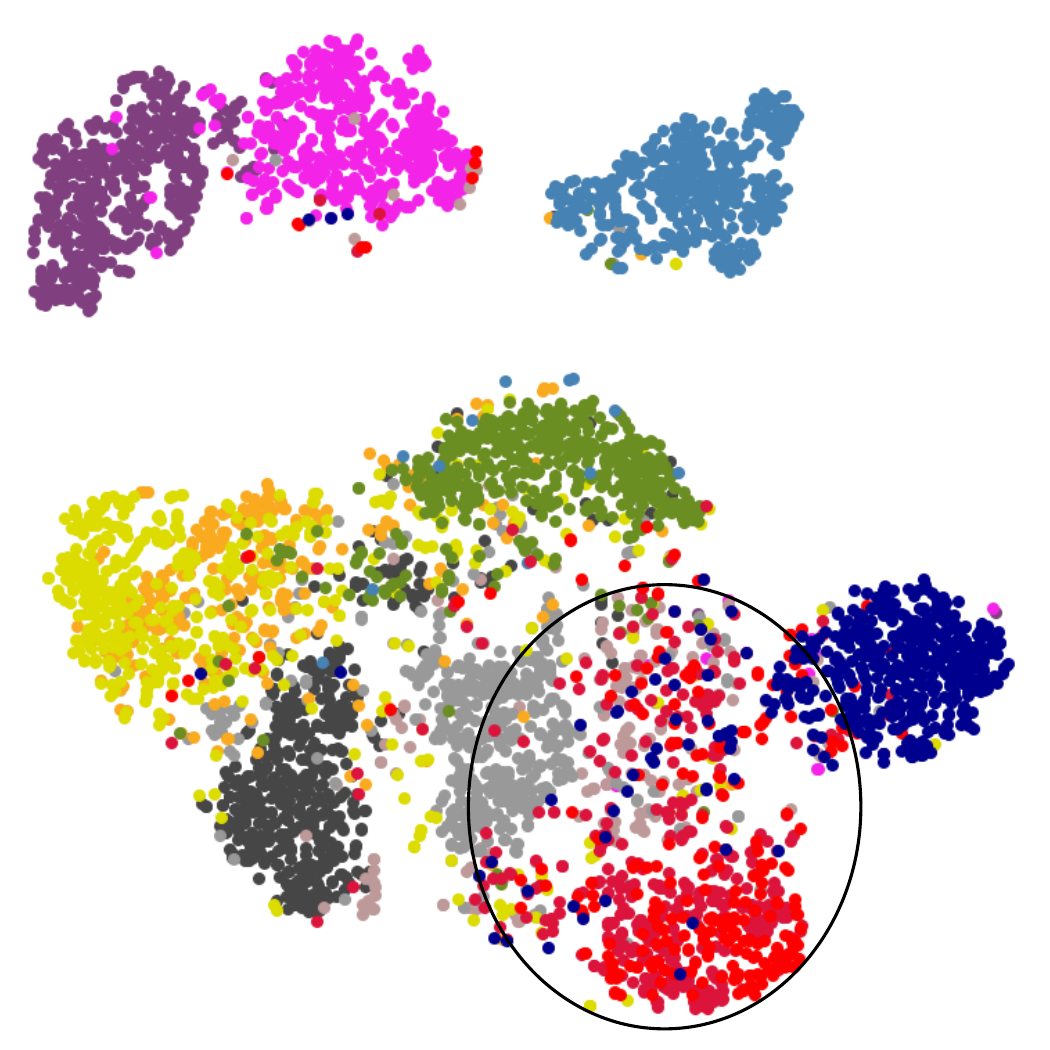}\label{a}}
	 \hfil
	\subfloat[DA-VSN]{\includegraphics[width=.15\columnwidth]{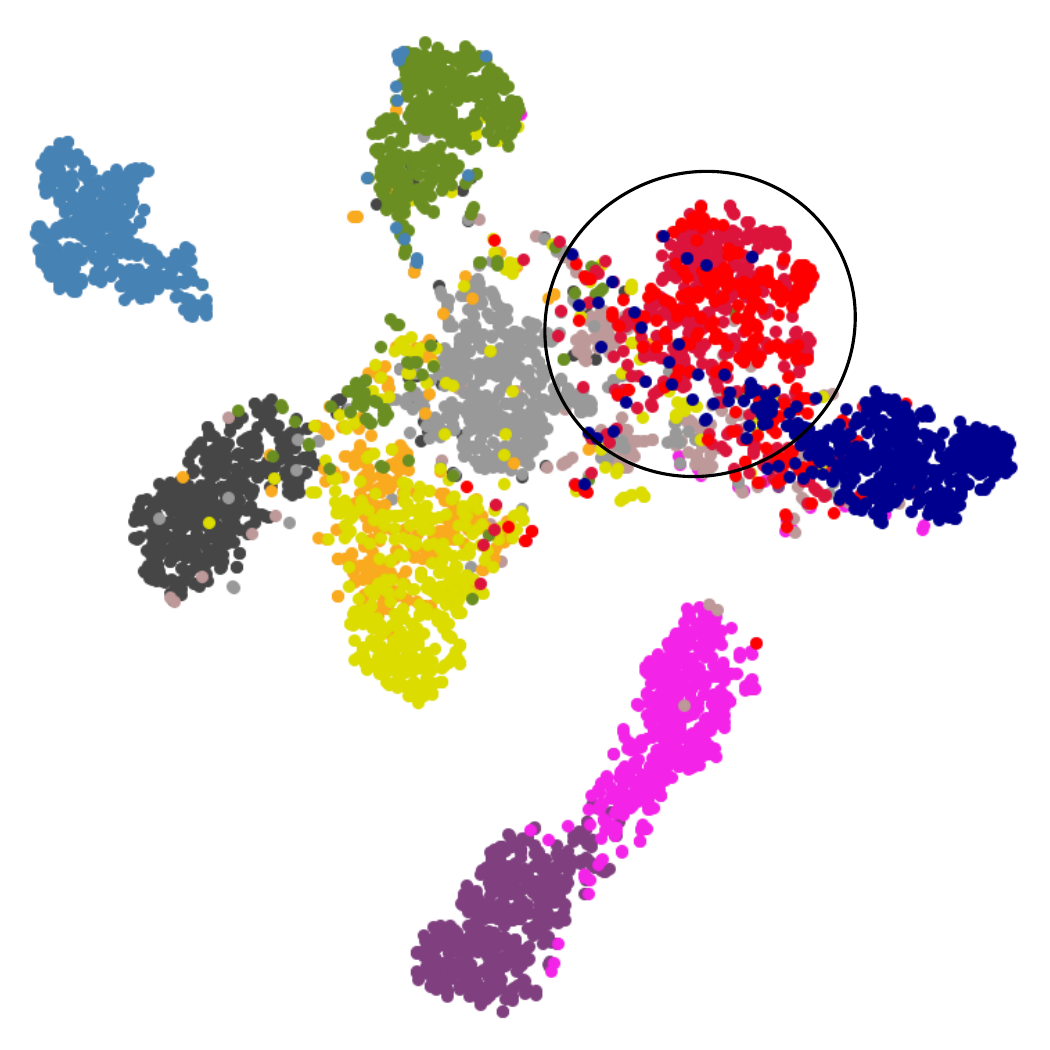}\label{b}}
	 \hfil
	\subfloat[TPS]{\includegraphics[width=.15\columnwidth]{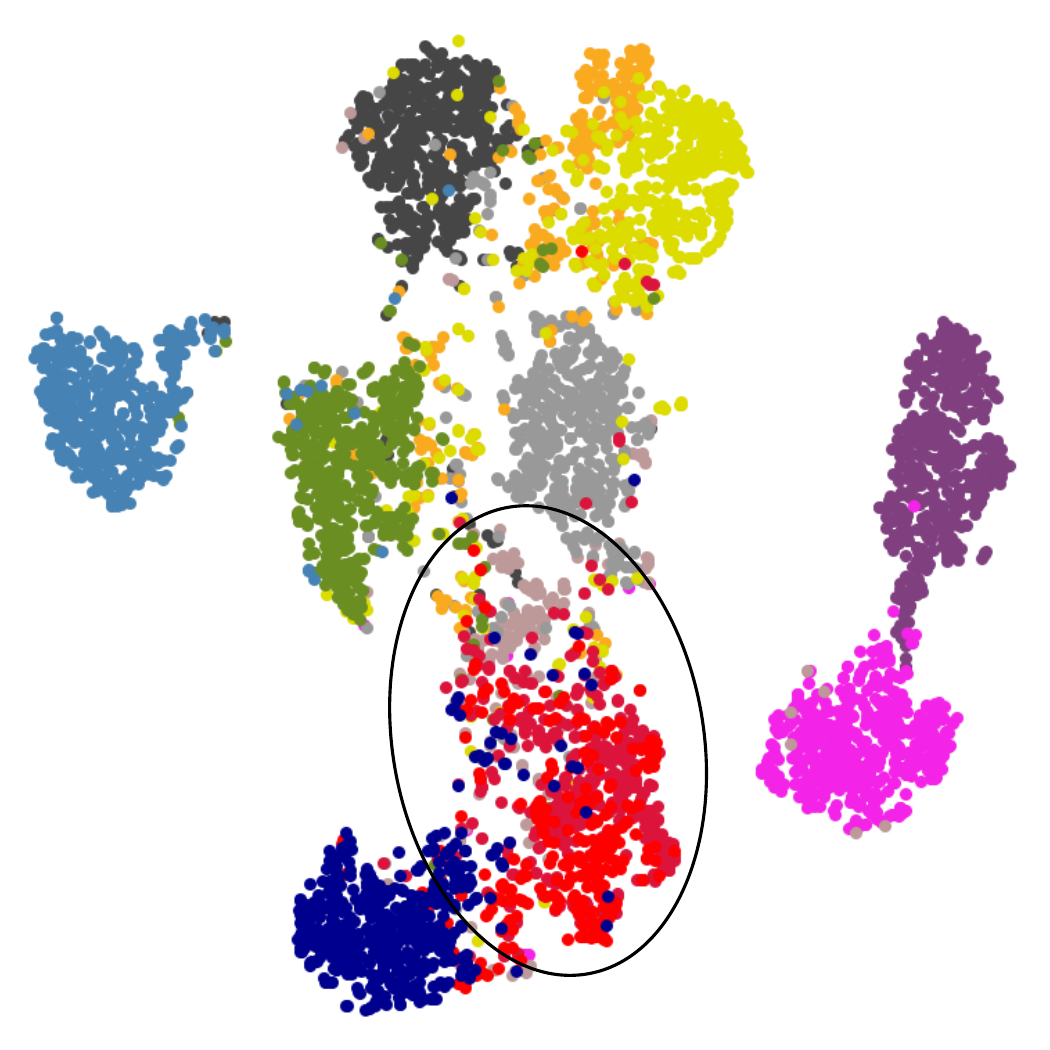}\label{c}}
	 \hfil
	\subfloat[CMOM]{\includegraphics[width=.15\columnwidth]{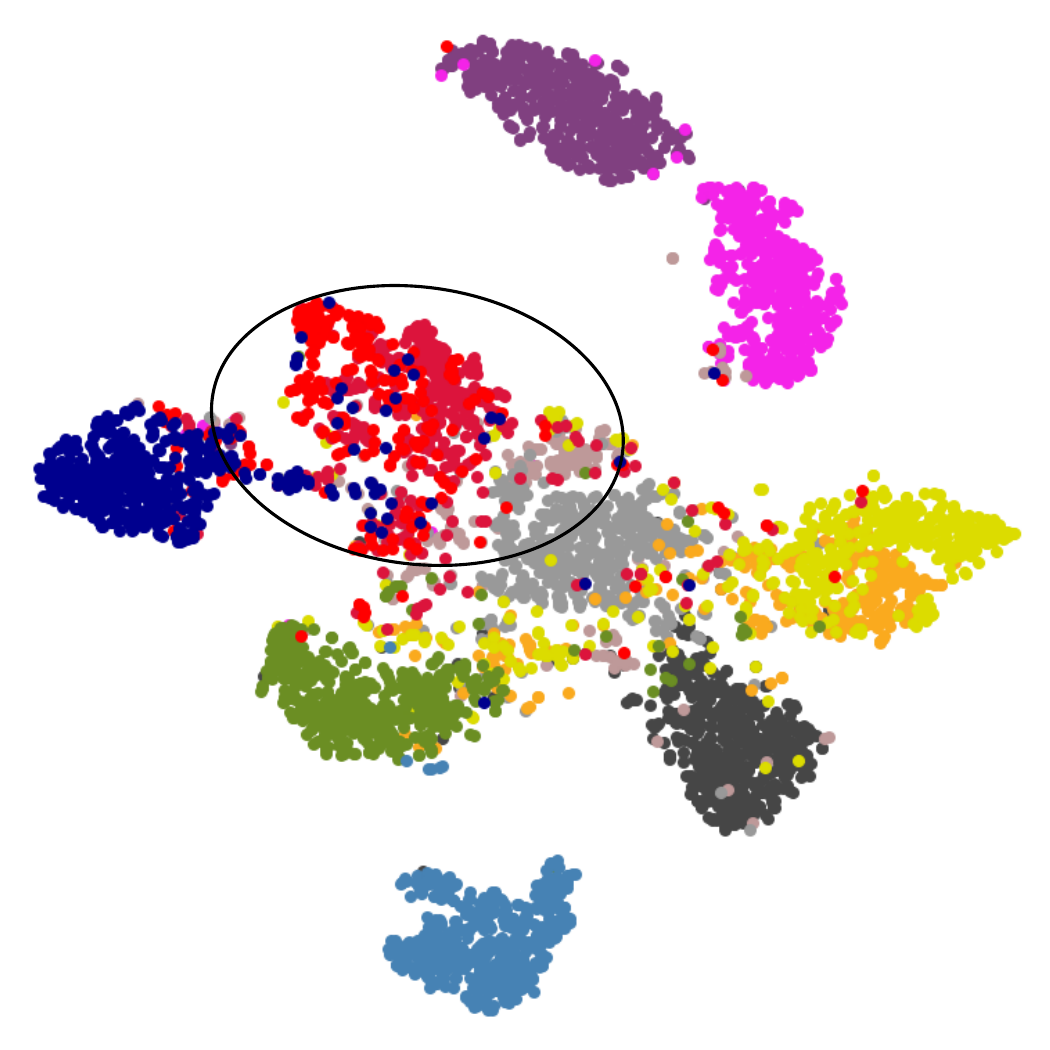}\label{d}}
	 \hfil
	\subfloat[Ours]{\includegraphics[width=.15\columnwidth]{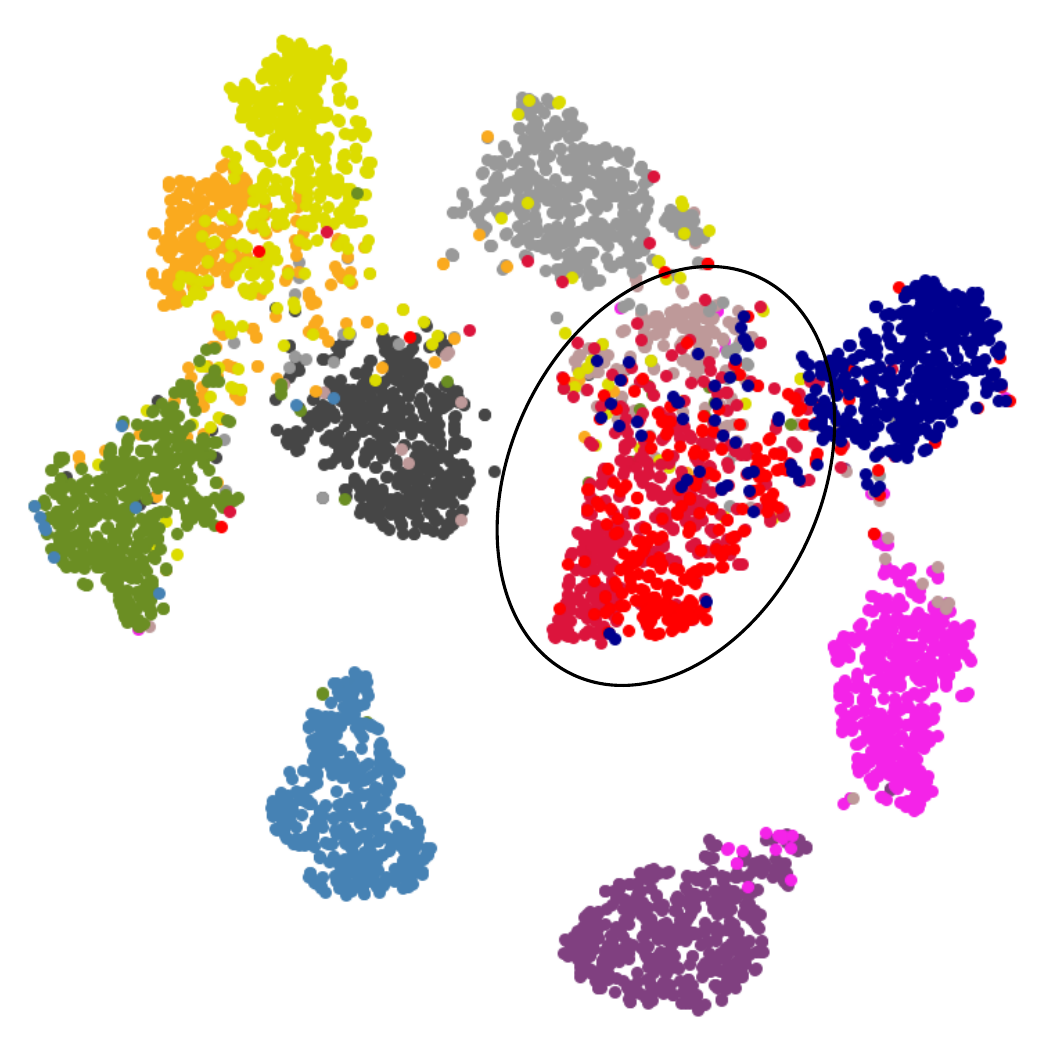}\label{e}}
	\\ 
	\subfloat[Source]{\includegraphics[width=.15\columnwidth]{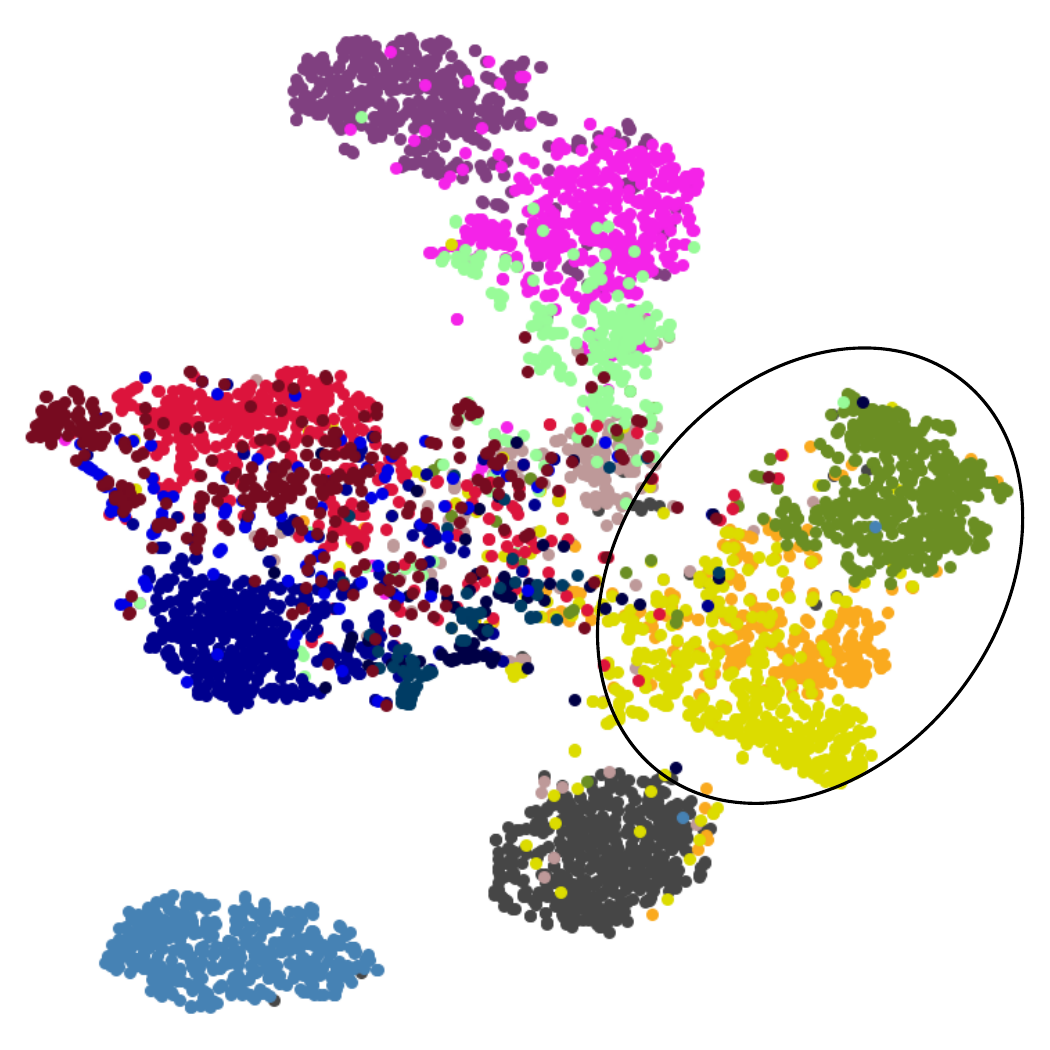}\label{f}}
	 \hfil
	\subfloat[DA-VSN]{\includegraphics[width=.15\columnwidth]{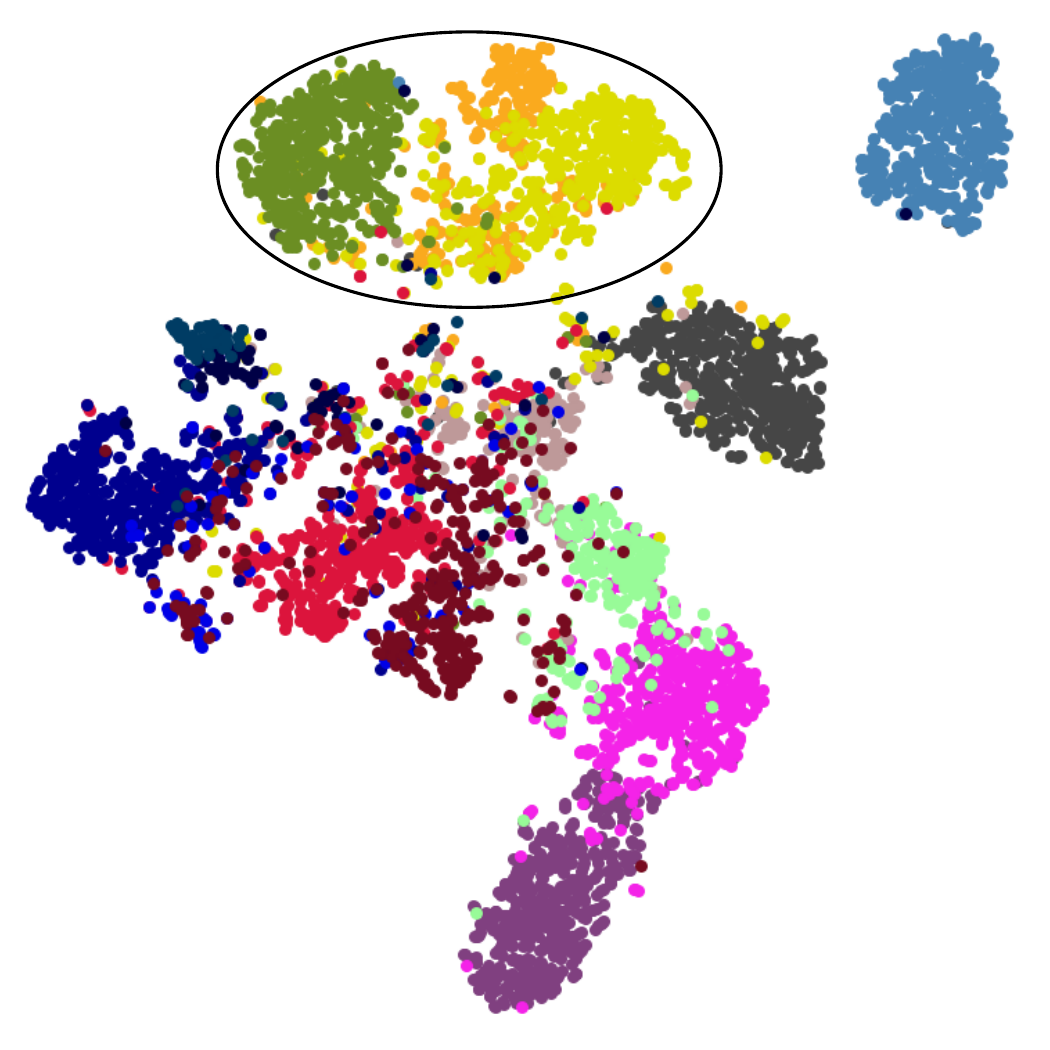}\label{g}}
 	\hfil
	\subfloat[TPS]{\includegraphics[width=.15\columnwidth]{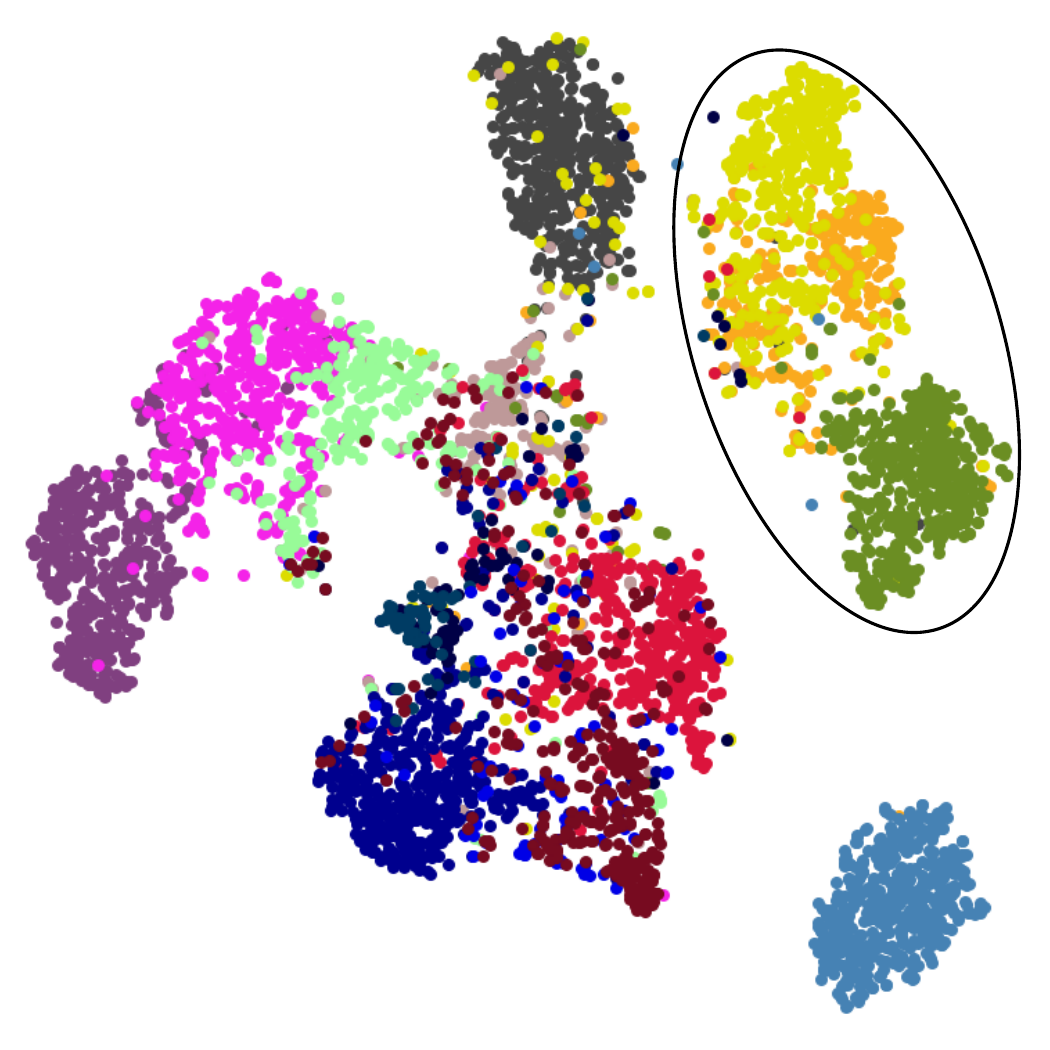}\label{h}}
	 \hfil
	\subfloat[CMOM]{\includegraphics[width=.15\columnwidth]{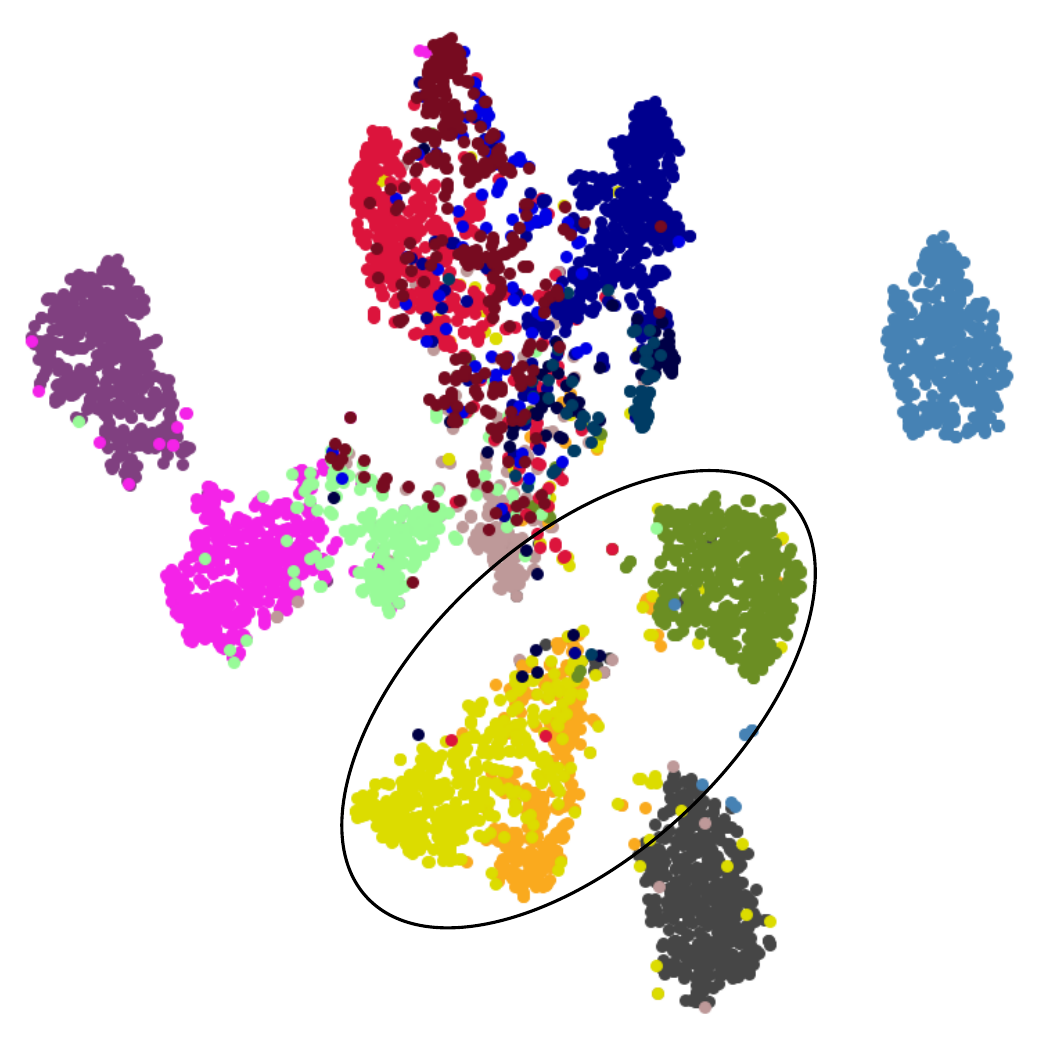}\label{i}}
	 \hfil
	\subfloat[Ours]{\includegraphics[width=.15\columnwidth]{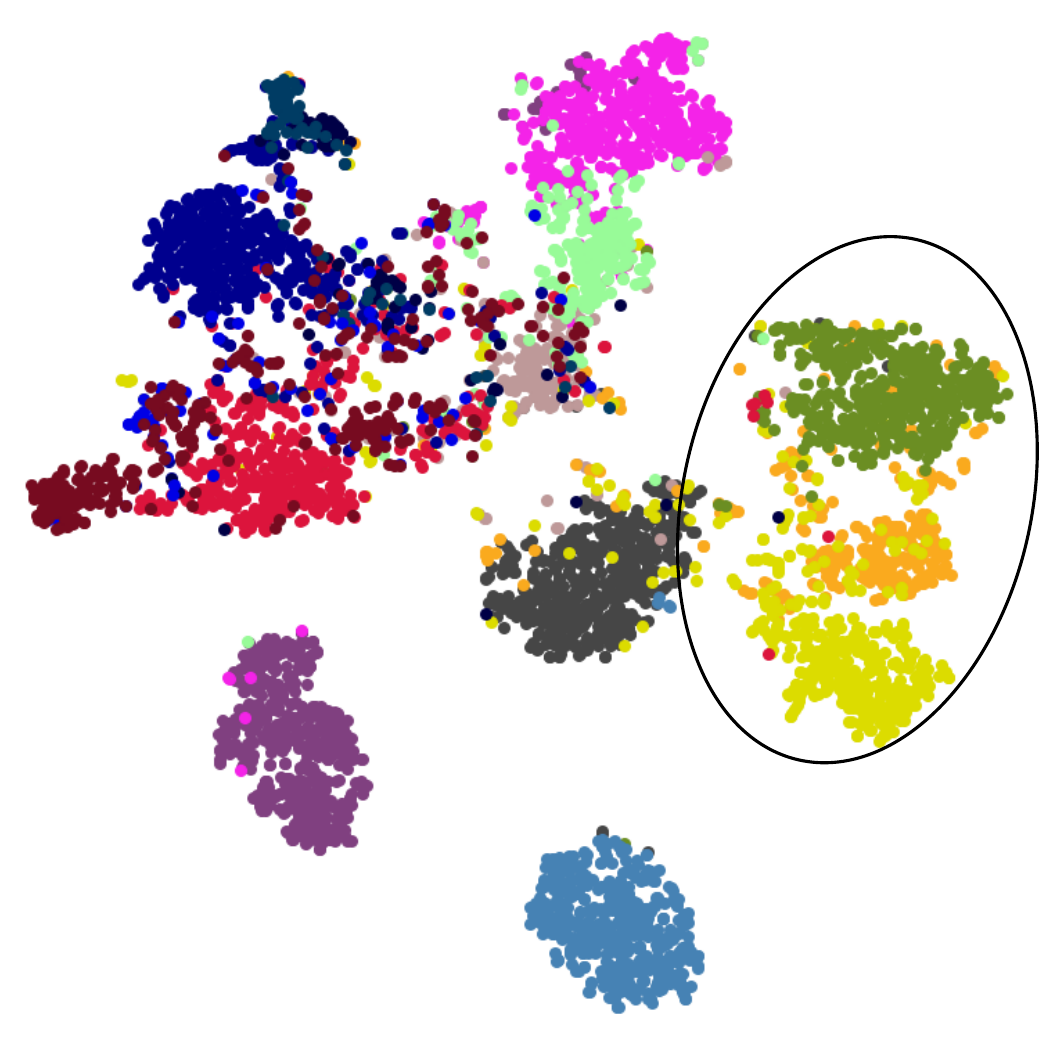}\label{j}}
	
\caption{T-SNE analysis of existing methods and our method on Synthia-Seq $\rightarrow$ Cityscapes-Seq (line one) and VIPER $\rightarrow$ Cityscapes-Seq (line two). Our model excels in learning more discriminative features of target videos within embedding space. Please zoom in for details.}
\label{fig. 9}
\vspace{-13pt}
\end{figure}

\subsubsection{The Throughput Measured for Different Modules} In Table~\ref{tab:table10}, we present the training time and GPU memory footprint associated with the integration of different components. ``Src." refers to training only with the source domain data. ``$\text{QuadMix*}$'' corresponds to the ablation setting ID 11 in Table~\ref{tab:table6}. Compared to ``Src.", the proposed QuadMix $(\mathcal{T}$$\rightarrow$$(\mathcal{S}$$\rightarrow$$\mathcal{S}))$ and $(\mathcal{S}$$\rightarrow$$(\mathcal{T}$$\rightarrow$$\mathcal{T}))$ leads to a moderate increase in training time and GPU memory, and ``Agg" exhibits relatively lighter overhead. 

We also investigate the computational cost of Exp. ID 11 in Table~\ref{tab:table6} (incrementally using 
$\mathcal{S}$$\rightarrow$$\mathcal{S}+\mathcal{T}$$\rightarrow$$\mathcal{T}+\mathcal{S}$$\rightarrow$$\mathcal{T}+\mathcal{T}$$\rightarrow$$\mathcal{S}$), which is higher in cost than our QuadMix. Notably, only adjacent frames and optical flow are required as inputs during model testing. In line with~\cite{davsn,tps,cmom,aaai}, our method maintains similar inference speed and memory footprint.

\subsubsection{T-SNE Analysis on Two Benchmarks.} 
For better visualization, we present t-SNE~\cite{tsne} visualizations of video feature embeddings from three recent works and the source-only model, compared with our QuadMix in Fig.~\ref{fig. 9}. The dot colors for categories are consistent with Fig.~\ref{fig. 7}. We observe that category-wise clusters in recent methods tend to be more scattered and overlapped, whereas our method generates more compact clusters. For instance, in our method, categories like \textit{perdon}, \textit{rider}, and \textit{fence} in Synthia-Seq $\rightarrow$ Cityscapes-Seq, and categories like \textit{light}, \textit{sign}, and \textit{vegetation} in VIPER $\rightarrow$ Cityscapes-Seq, exhibit more compact intra-category distribution and more distinct inter-category differences, revealing better discriminative capability for the target domain.

\begin{figure}[!t]
	\centering
	\includegraphics[width=\linewidth]{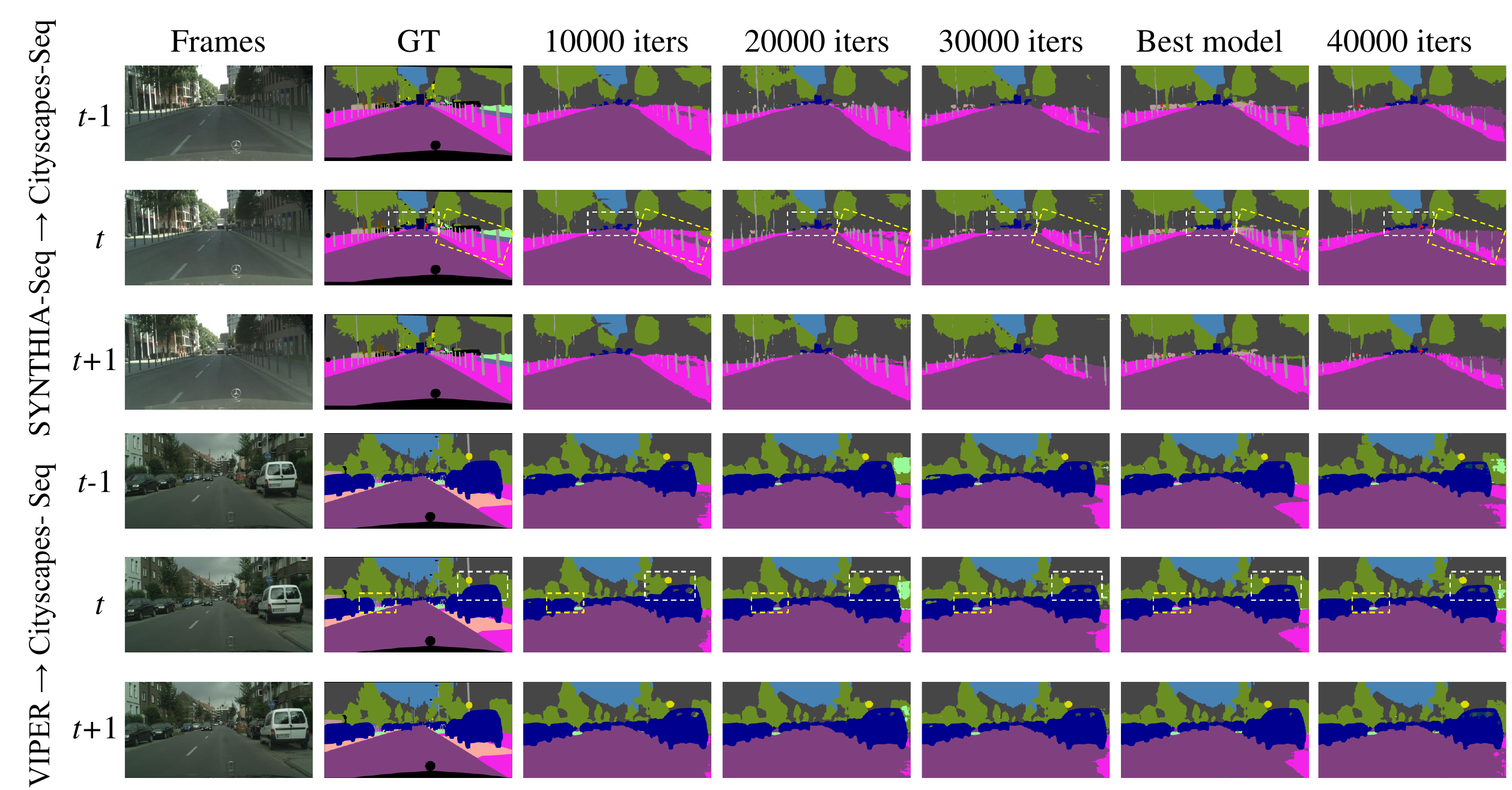}
	\caption{Qualitative semantic segmentation results of our models at different training iterations on video UDA-SS benchmarks. Please zoom in for details.}
	\label{fig. 10}
    \vspace{-12pt}
\end{figure}

\subsubsection{Results of Different Training Iterations} 
In Fig.~\ref{fig. 10}, we provide a more comprehensive presentation of the semantic segmentation results of our models trained at different iterations. Notably, the best models for both two benchmarks emerge between the 35k and 40k iterations, providing a clearer insight into the evolution and stability of the training process.

\section{Conclusion}
This paper tackles the more challenging task of unified UDA-SS across both video and image scenarios by presenting an efficient and comprehensive approach.
Specifically, we propose the QuadMix that involves quad-directional intra- and inter-domain mixing at both pixel and feature levels, aiming to generalize intra-domain features, enhance inter-domain gap bridging, and alleviate feature inconsistencies for domain invariant learning. Furthermore, we propose flow-guided video feature aggregation for fine-grained domain alignment, applicable to image scenarios as well. 
Our method is compatible with both CNN and transformer-based architectures. 
Extensive experiments on four challenging benchmarks demonstrate the superiority of our method.

\textbf{Limitation:} One limitation is that, while the quad-mixing enhances domain adaptation effectively, the theoretical interpretability of underlying mechanisms remains to be explored.

\textbf{Future work:} An interesting aspect of future work is to explore unified techniques across video and image scenarios, particularly in multi-modal domain adaptation, open-set domain adaptation, and extensions to instance or panoramic segmentation, as well as pixel-level anomaly detection.

\section*{Acknowledgment}
This work is supported in part by the Major Program of the National Natural Science Foundation of China (NSFC) under Grant No. 61991404, in part by the Research Program of the Liaoning Liaohe Laboratory under Grant No. LLL23ZZ-05-01, in part by the Key Research and Development Program of Liaoning Province under Grant No. 2023JH26/10200011, in part by the Natural Science Foundation of Liaoning Province of China under Grant No. 2024-MSBA-42, and in part by the Program of China Scholarship Council 202306080142. Dr. Tao's research is partially supported by NTU RSR and Start Up Grants. The authors would like to thank Dr. Chunhua Shen and Lan Deng for their valuable suggestions.

\clearpage
\section*{\centering \Large Supplementary Material}

\renewcommand{\thesection}{S\arabic{section}}
\setcounter{section}{0}

This document presents the supplementary materials omitted from the main paper due to space limitations.

\section{Training algorithm}
In this paper, our unified method follows an end-to-end learning paradigm. The training process for \textbf{UDA-VSS} is outlined in Algorithm~\ref{supp_alg1}. For \textbf{UDA-ISS}, the process is similar but excludes temporal information, i.e., optical flow. Please refer to our main paper for detailed equations and loss functions.

\begin{algorithm}[H]
	\caption{Unified Domain Adaptive Semantic Segmentation.}
	
	\begin{algorithmic}[1]
		\STATE \textbf{Input:} $X_t^\mathcal{S}$, $X_t^\mathcal{T}$, $X_{t - \tau }^\mathcal{T}$, $o_{t - 1 \to t}^{ \mathcal{S}}$, $o_{t - \tau  - 1  \to t - \tau }^{ \mathcal{T}}$, $o_{{t'\to t}}^{ \mathcal{T}}$, ${\lambda _\mathcal{T}}$, ${\lambda _f}$, batch size $B$, and maximum iteration N.
		\STATE Initialize our model using ImageNet pre-trained model.
		\FOR{$n \gets 0$ to $N$ }
		\IF{$n == 0$} 
		\STATE Obtain initial pseudo-label $\hat Y_t^T$ and mask $M_t^{\mathcal{T}^*}$.
		\ELSE
		\STATE Generate video patch templates from the source and target domains adaptively via Eq.~(1).
		\STATE Create intra-mixed domains via Eq.~(8) to Eq.~(10).
		\STATE Create more enhanced inter-mixed domains via Eq.~(11) to Eq.~(12).
		\STATE Perform feature-level sequence mixing across two quad-mixed domains via Eq.~(13).
        \STATE Conduct flow-guided feature aggregation for domain alignment via Eq.~(18) to Eq.~(20).
		\STATE Train the model with loss $\mathcal{L}_{all}$ via Eq.~(21).
		\ENDIF
		\ENDFOR
		\STATE \textbf{Output:} Final weights of the UDA-VSS model.
	\end{algorithmic}
	\label{supp_alg1}
\end{algorithm} 	

Note that the two inter-mixed domains derived from the two intra-mixed domains are also termed quad-mixed domains, as they are obtained through quad-directional mixing.

\section{Differentiation of image UDA-SS Between Our method and MoDA~\cite{moda}}
Recently, Pan et al. propose MoDA~\cite{moda} and report segmentation mIOU on the \textit{GTAV → Cityscapes-\textbf{Seq}}, referring to it as domain adaptive \texttt{image} semantic segmentation. However, MoDA differs fundamentally from the concepts within mainstream methods~\cite{sepico, daformer, fredom, adpl, I2f, bdm, dsp, sac, dacs, rda, pixmatch} and \texttt{image UDA-SS} in this paper. The reason is as follows: MoDA utilizes temporal information from the target domain videos as geometric constraints to refine pseudo-labels. In contrast, we generate pseudo-labels using only spatial image information, following prior works.

Essentially, MoDA represents an attempt at domain adaptation from a source \texttt{image} domain to a target \texttt{video} domain, similar to I2VDA~\cite{necess}, which is discussed in the main paper. As for performance, MoDA achieves mIOUs of 54.9$\%$ and 75.2$\%$ on \textit{GTAV → Cityscapes-\textbf{Seq}}, using CNN~\cite{deeplabv2} or transformer-based~\cite{segformer} networks, respectively. In comparison, we obtain mIOUs of 66.8$\%$ and 76.1$\%$ on \textit{GTAV → Cityscapes}, respectively, as shown in Table III of the main paper. Despite leveraging temporal information, the results of MoDA remain below ours. Notably, using the CNN-based network, we outperform MoDA by 11.9$\%$, demonstrating its effectiveness and substantial potential.

Additionally, MoDA only reports results on one video UDA-SS benchmark, Viper → Cityscapes-Seq, with an mIOU of 49.1$\%$ (CNN-based). In contrast, our method achieved 56.2$\%$ (CNN-based), significantly surpassing MoDA. Moreover, unlike previous works~\cite{davsn, tps, necess, aaai, pat, cmom} that only employ CNN-based networks, \textit{for the first time}, we explore the vision transformer~\cite{daformer} in video UDA-SS, achieving a higher mIOU of 58.2$\%$. Furthermore, we also conduct experiments on the Synthia-Seq → Cityscapes-Seq benchmark, achieving mIOUs of 61.5$\%$ and 67.2$\%$ with CNN or transformer-based networks, respectively. Based on the above analysis, we indeed pioneer unified image and video UDA-SS, setting new state-of-the-art benchmarks on four challenging benchmarks.
 
\section{Extended Experimental Analysis}
\subsection{Qualitative Results for Image UDA-SS} 
Besides the qualitative results in the main paper (Fig.~7) for video UDA-SS tasks, in Fig.~\ref{supp_fig. syn} and Fig.~\ref{supp_fig. gta}, we present the results for image UDA-SS tasks: SYNTHIA $\rightarrow$ Cityscapes and GTAV $\rightarrow$ Cityscapes, as predicted by our method. We compare these results with those predicted by ADPL~\cite{adpl}, DAFormer~\cite{daformer}, and SePiCo~\cite{sepico}. Our model's predictions are smoother and contain fewer spurious areas compared to the other models, indicating a significant performance improvement.

\subsection{More Qualitative Comparisons}
\begin{figure*}[!t]
	\centering
	\includegraphics[width=1.\linewidth]{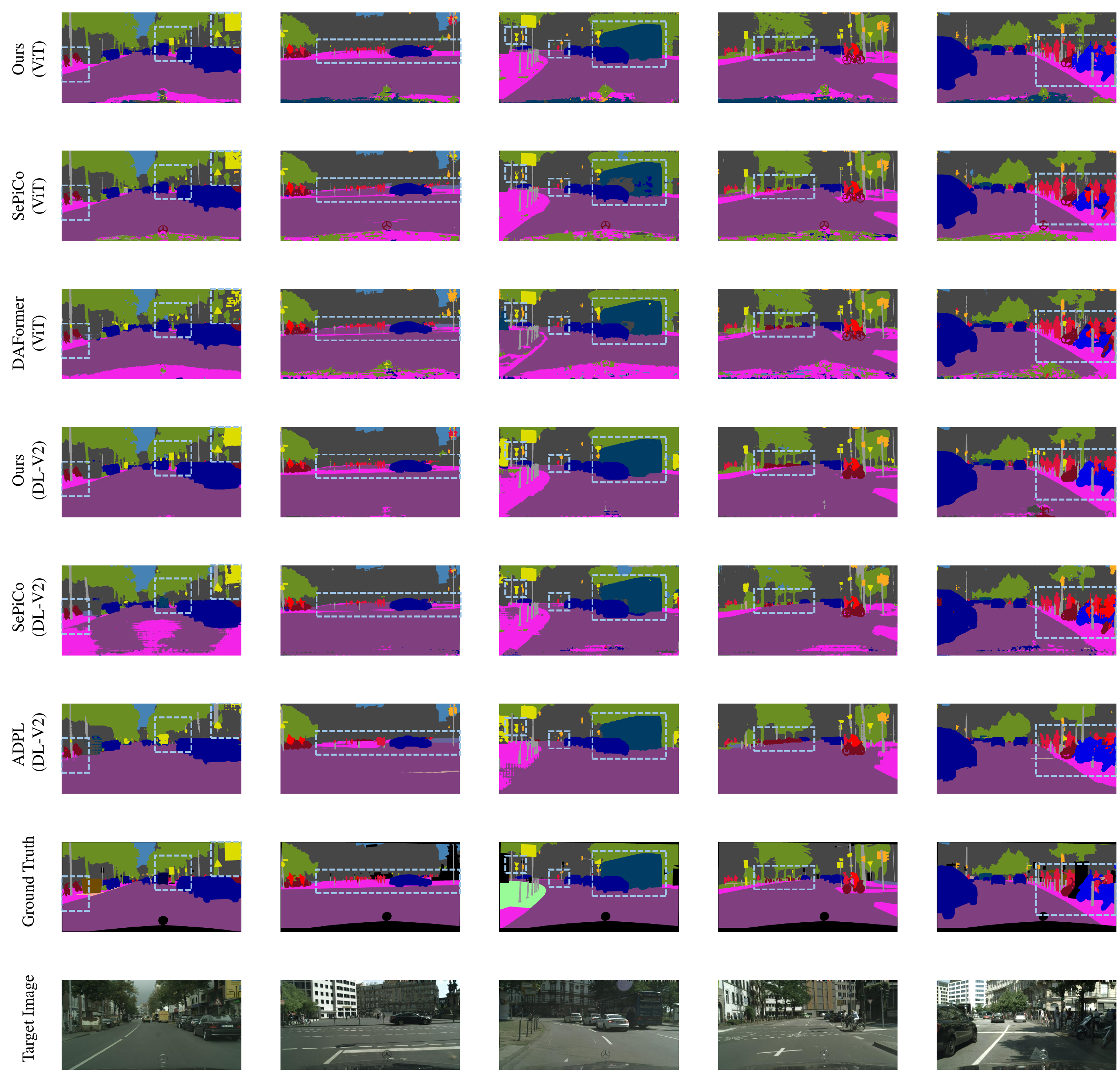}
	\caption{Qualitative comparison of results on the image UDA-SS task (SYNTHIA $\rightarrow$ Cityscapes). We compare our method with ADPL~\cite{adpl}, DAFormer~\cite{daformer}, and SePiCo~\cite{sepico}, including results using CNN~\cite{deeplabv2} and Vision Transformer~\cite{daformer} architectures. Please zoom in for details.
	}
	\label{supp_fig. syn}
\end{figure*}

\begin{figure*}[!t]
	\centering
	\includegraphics[width=1.\linewidth]{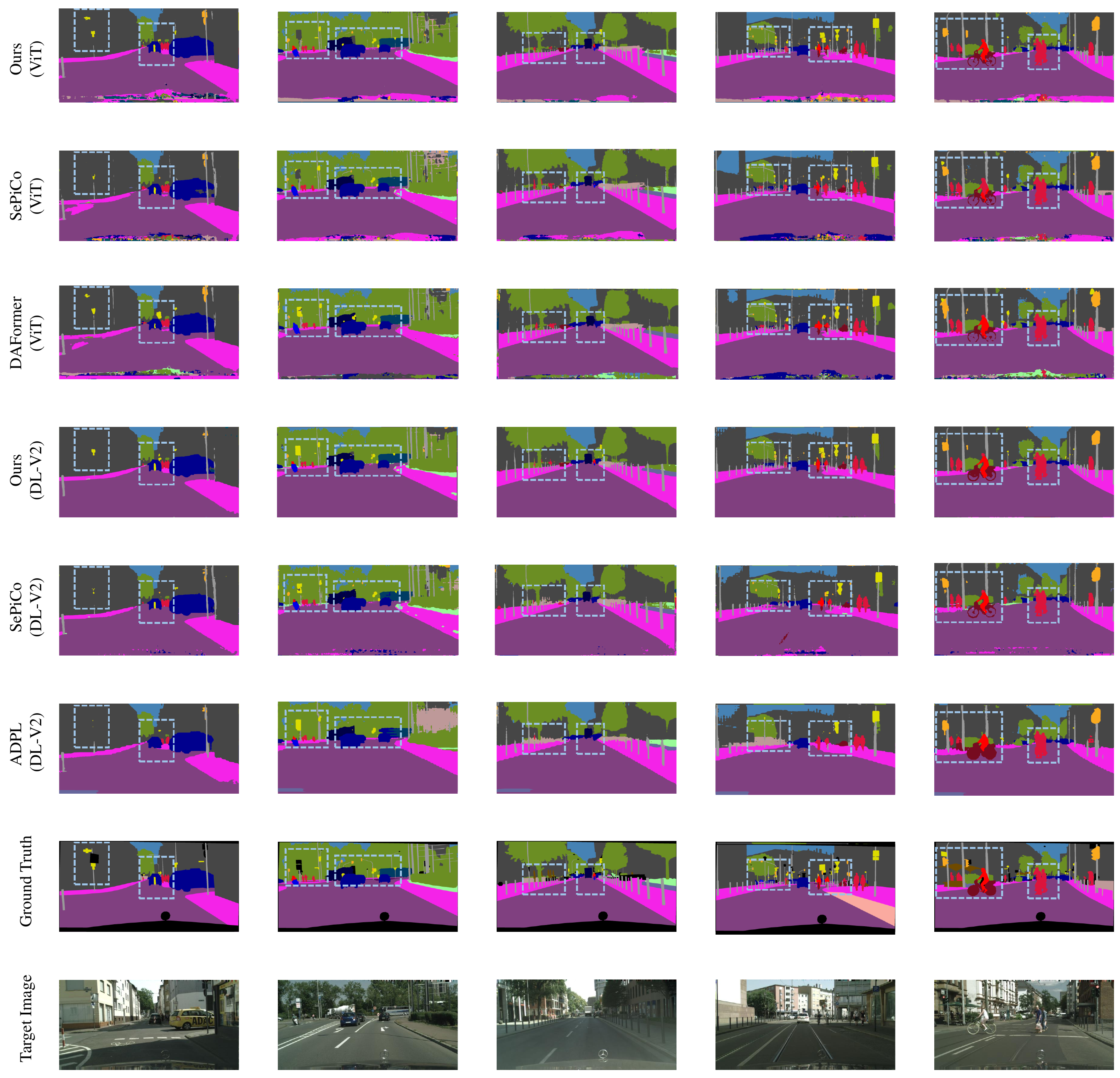}
	\caption{Qualitative comparison of results on the image UDA-SS task (GTAV $\rightarrow$ Cityscapes). We compare our method with ADPL~\cite{adpl}, DAFormer~\cite{daformer}, and SePiCo~\cite{sepico}, including results using CNN~\cite{deeplabv2} and Vision Transformer~\cite{daformer} architectures. Please zoom in for details.
	}
	\label{supp_fig. gta}
\end{figure*}

Besides Fig.~\ref{supp_fig. syn} and Fig.~\ref{supp_fig. gta}, and Fig.~7 in the main paper, please refer to the \href{https://drive.google.com/file/d/1OT5GtsbC0CcW6aydBL27ADjve95YE5oj/view?usp=drive_link}{\textit{supplementary video}} for more visual comparisons of both \texttt{image} and \texttt{video} UDA-SS segmentation results with previous methods.

\begin{table}[t]
	\caption{Classes division regarding ``things" or``stuff", and ``movable" or ``stationary" categories on Synthia-Seq $\rightarrow$ Cityscapes-Seq.
		\label{supp_tab:table3}}
	\centering
	\renewcommand{\arraystretch}{1.47}
	\begin{tabular}{cc}
 \toprule[1.0pt]
\multicolumn{2}{c}{SynthiaSeq $\rightarrow$ Cityscapes-Seq}                                                             \\ \hline
\multicolumn{1}{c|}{thing}   & building, pole, light, sign, person, rider, car \\\hline
\multicolumn{1}{c|}{stuff}   & road, side, vegetation, sky                                \\ \hline
\multicolumn{1}{c|}{movable object}    & person, rider, car \\\hline
\multicolumn{1}{c|}{stationary object} & road, side, building, pole, light, sign, vegetation, sky            \\   \toprule[1.0pt]
\end{tabular}
\end{table}

\begin{table}
	\caption{Classes division regarding ``things" or``stuff", and ``movable" or ``stationary" categories on Viper $\rightarrow$ Cityscapes-Seq.
		\label{supp_tab:table4}}
	\centering
	\renewcommand{\arraystretch}{1.47}
	\begin{tabular}{c|c}
 \toprule[1.0pt]
\multicolumn{2}{c}{Viper $\rightarrow$ Cityscapes-Seq}                                                             \\ \hline
 
\multirow{2}{*}{thing} & building, fence, light, sign,  \\ 
        & person, car, truck, bus, motorcycle, bike \\  \hline
\multicolumn{1}{c|}{stuff}   & road, side, vegetation, terr, sky                                 \\ \hline
\multicolumn{1}{c|}{movable object}    & person, car, truck, bus, motorcycle, bike                    \\ \hline
\multirow{2}{*}{stationary object} & road, side, building, fence,  \\ 
        & light, sign, vegetation, terr, sky              \\   \toprule[1.0pt]
\end{tabular}
\end{table}

\subsection{Category Selection for Video Patch Template Generation}
In Table~VII of the main paper, we investigated the impact of category selection on video patch template generation ($Z_{t}^{\mathcal{S^*},k}$ and $Z_{t}^{\mathcal{T^*},k}$) from both domains. We explored the results when the category selection space (distinct from the long-tail category space) comprised only ``things" or ``stuff"~\cite{quanjing}, as well as solely movable or stationary objects instead of the whole category space. Specifically, the classes division for Synthia-Seq $\rightarrow$ Cityscapes-Seq is shown in Table~\ref{supp_tab:table3} and for Viper $\rightarrow$ Cityscapes-Seq is shown in Table~\ref{supp_tab:table4}.

\subsection{Long-Tail Categories for Templates}
Long-tail class learning is an aspect for tackling rare object learning~\cite{fredom,adpl,dsp}. In this paper, we also involve two selected long-tail categories in \textit{Source to Source Mixing $(\mathcal{S}$$\rightarrow$$\mathcal{S})$} and \textit{Source to Intra-mixed Target Mixing }$(\mathcal{S}$$\rightarrow$$(\mathcal{T}$$\rightarrow$$\mathcal{T}))$ at each training iteration. Specifically, we randomly select two long-tail classes from categories of $\textit{pole}$, $\textit{light}$, $\textit{sign}$, and $\textit{rider}$ for Synthia-Seq $\rightarrow$ Cityscapes-Seq, and $\textit{fence}$, $\textit{light}$, $\textit{sign}$, $\textit{terrain}$, $\textit{motor}$, and $\textit{bike}$ for VIPER $\rightarrow$ Cityscapes-Seq. 

\subsection{Extended Visualization of Mixing Strategies in QuadMix}
\begin{figure*}[!h]
	\centering
	\includegraphics[width=1.0\linewidth]{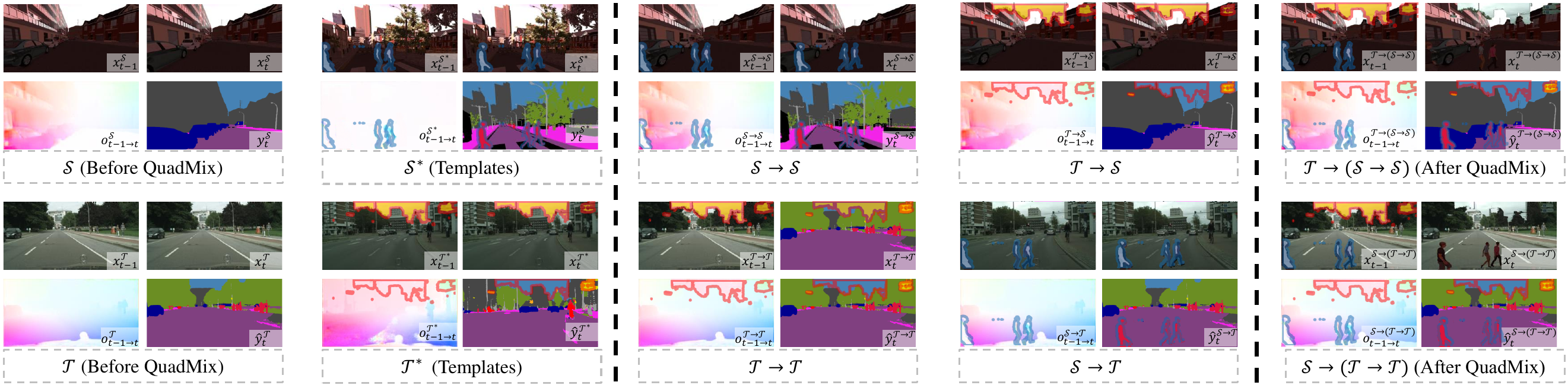}
	\caption{Examples of various mixing strategies in QuadMix on the \textbf{SYNTHIA-Seq $\rightarrow$ Cityscapes-Seq} benchmark: (i) $\mathcal{S}$ and $\mathcal{T}$ (before QuadMix), (ii) $\mathcal{S^*}$ and $\mathcal{T^*}$ (source templates: $\textit{person}$ and $\textit{rider}$, target templates: $\textit{sign}$ and $\textit{sky}$), 
    (iii) $\mathcal{S} \rightarrow \mathcal{S}$ and $\mathcal{T} \rightarrow \mathcal{T}$ (intra-domain mixing), (iv) $\mathcal{S} \rightarrow \mathcal{T}$ and $\mathcal{T} \rightarrow \mathcal{S}$ (not directly applied in this work), (v) $\mathcal{S} \rightarrow (\mathcal{T} \rightarrow \mathcal{T})$ and $\mathcal{T} \rightarrow (\mathcal{S} \rightarrow \mathcal{S})$ (further with inter-domain mixing, i.e. after QuadMix). The effects of these strategies on video frames, optical flow, and labels/pseudo-labels are illustrated. We present $x_t^{\mathcal{T} \rightarrow (\mathcal{S}\rightarrow \mathcal{S})}$ and $x_t^{\mathcal{S} \rightarrow (\mathcal{T}\rightarrow \mathcal{T})}$ without masks for better understanding. Notably, the patch templates required in training iteration $n$ are generated online adaptively from iteration $n-1$. Please zoom in for details.
    }
	\label{supp_fig. 1}
\end{figure*}

\begin{figure*}[!h]
	\centering

	\includegraphics[width=1.0\linewidth]{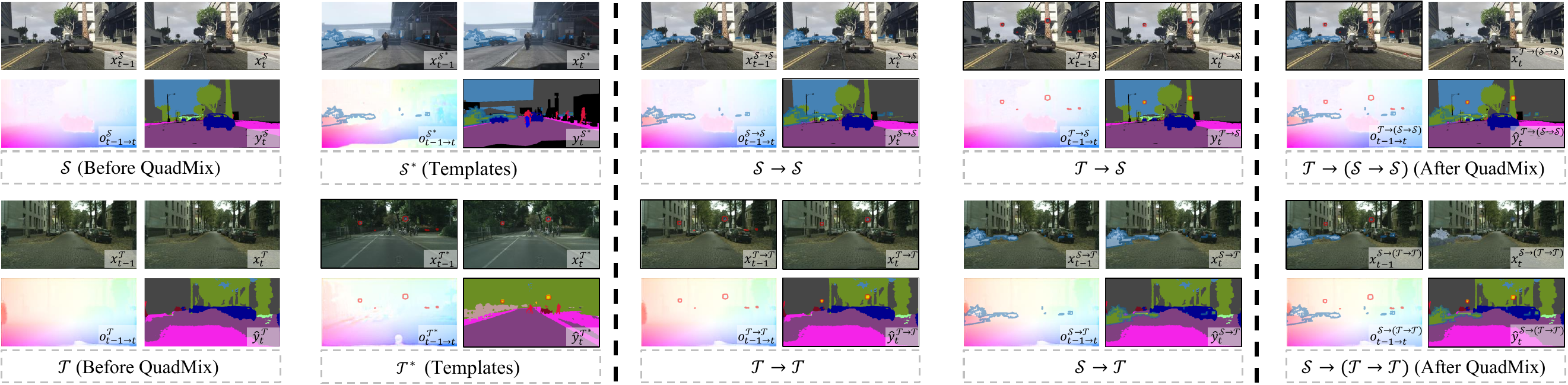}
	\caption{Examples of various mixing strategies in QuadMix on the \textbf{Viper $\rightarrow$ Cityscapes-Seq} benchmark: (i) $\mathcal{S}$ and $\mathcal{T}$ (before QuadMix), (ii) $\mathcal{S^*}$ and $\mathcal{T^*}$ (source templates: $\textit{vegetation}$ and $\textit{terrain}$, target templates: $\textit{sign}$ and $\textit{building}$), (iii) $\mathcal{S} \rightarrow \mathcal{S}$ and $\mathcal{T} \rightarrow \mathcal{T}$ (intra-domain mixing), (iv) $\mathcal{S} \rightarrow \mathcal{T}$ and $\mathcal{T} \rightarrow \mathcal{S}$ (not directly applied in this work), (v) $\mathcal{S} \rightarrow (\mathcal{T} \rightarrow \mathcal{T})$ and $\mathcal{T} \rightarrow (\mathcal{S} \rightarrow \mathcal{S})$ (further with inter-domain mixing, i.e. after QuadMix). The effects of these strategies on video frames, optical flow, and labels/pseudo-labels are illustrated. We present $x_t^{\mathcal{T} \rightarrow (\mathcal{S}\rightarrow \mathcal{S})}$ and $x_t^{\mathcal{S} \rightarrow (\mathcal{T}\rightarrow \mathcal{T})}$ without masks for better understanding. Notably, the patch templates required in training iteration $n$ are generated online adaptively from iteration $n-1$. Please zoom in for details.}
	\label{supp_fig. 2}
\end{figure*}

In Fig. \ref{supp_fig. 1}, we extend the visual analysis presented in Fig. 4 of the main paper, further illustrating the effects of $\mathcal{S}\rightarrow\mathcal{T}$ and $\mathcal{T}\rightarrow\mathcal{S}$ for the SYNTHIA-Seq \cite{synthia} $\rightarrow$ Cityscapes-Seq \cite{citys} benchmark. For additional insights, Fig. \ref{supp_fig. 2} presents corresponding visualizations for the Viper \cite{viper} $\rightarrow$ Cityscapes-Seq benchmark. Moreover, Fig. \ref{supp_fig. 5} to Fig. \ref{supp_fig. 10} present an extensive set of QuadMix visualizations across both video benchmarks, covering various strategies including $\mathcal{S}\rightarrow\mathcal{S}$, $\mathcal{T}\rightarrow\mathcal{T}$, $\mathcal{S}\rightarrow(\mathcal{T}\rightarrow\mathcal{T})$, and $\mathcal{T}\rightarrow(\mathcal{S}\rightarrow\mathcal{S})$, thereby enabling a more thorough understanding.

\subsection{Visualization of the Evolution of Target Pseudo-Labels}
\begin{figure*}[!h]
	\centering

	\includegraphics[width=1.0\linewidth]{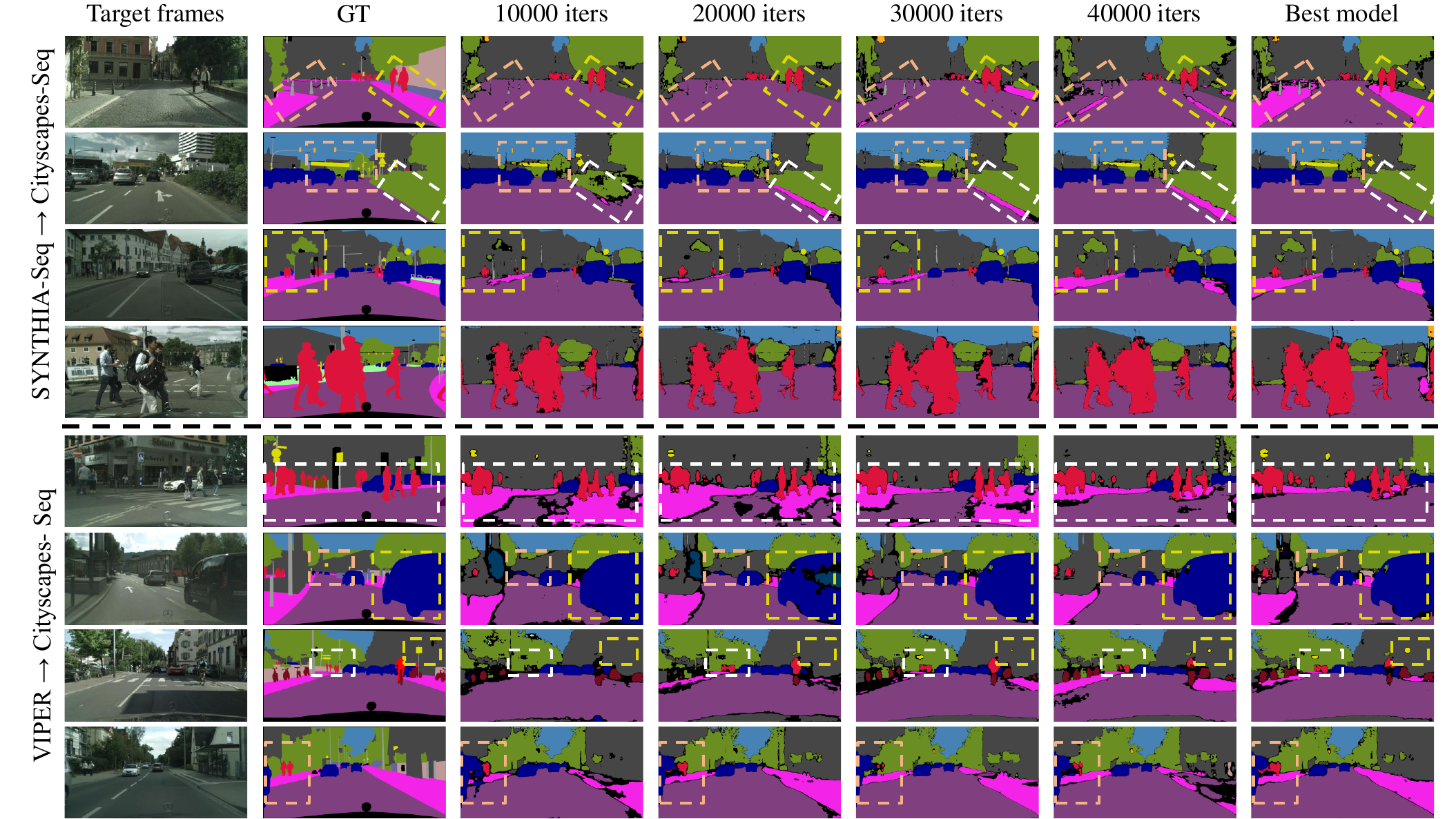}
	\caption{Visualization of the evolution of pseudo-labels generated by our model with a \textbf{CNN} backbone on the target domain (Cityscapes-Seq). The pseudo-label quality improves progressively throughout training.}
	\label{supp_fig. 3}
\end{figure*}

\begin{figure*}[!h]
	\centering

	\includegraphics[width=1.0\linewidth]{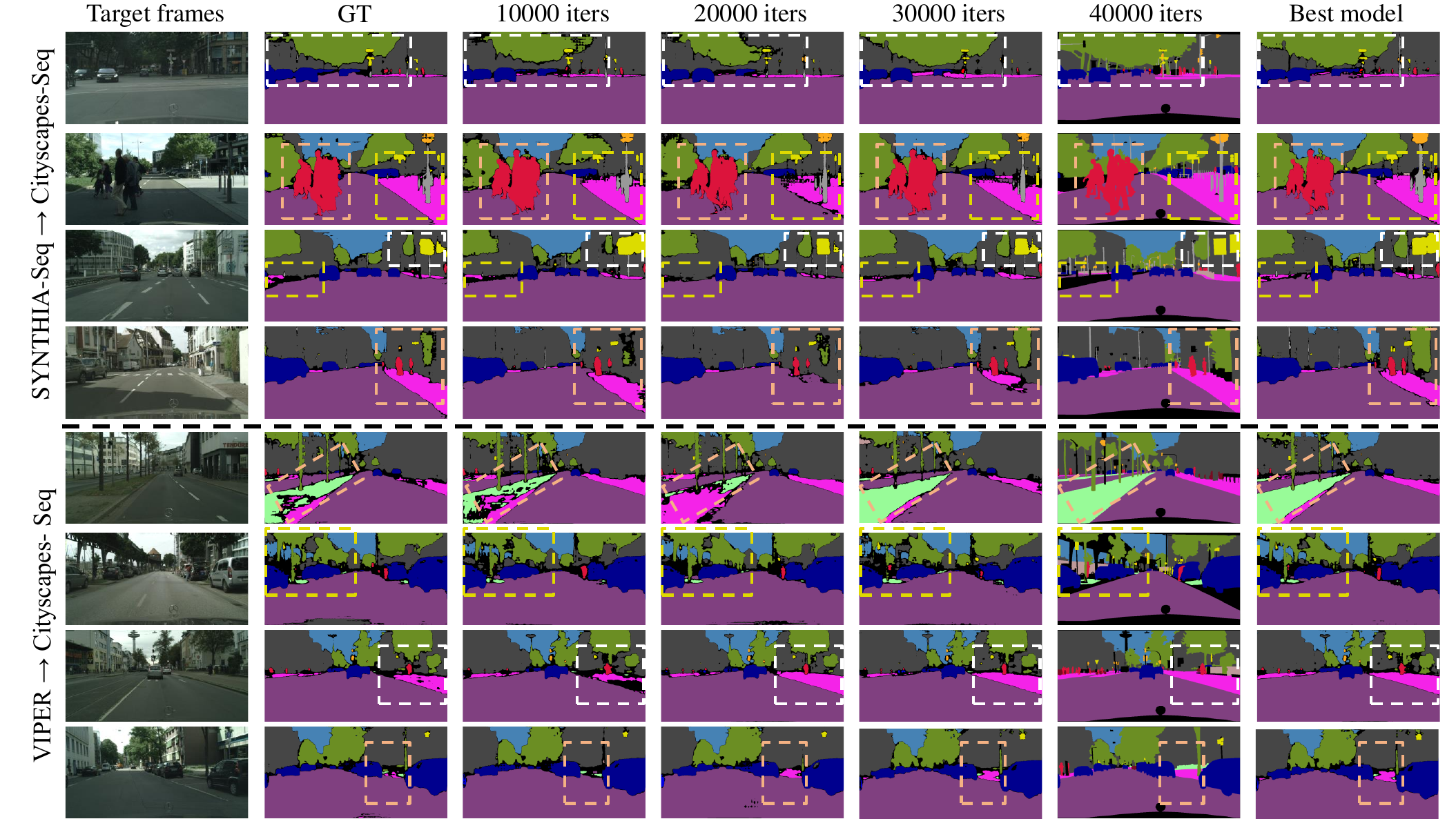}
	\caption{Visualization of the evolution of pseudo-labels generated by our model with a \textbf{ViT} backbone on the target domain (Cityscapes-Seq). The pseudo-label quality improves progressively throughout training.}
	\label{supp_fig. 4}
\end{figure*}

\begin{figure*}[!t]
	\centering

	\includegraphics[width=0.85\linewidth]{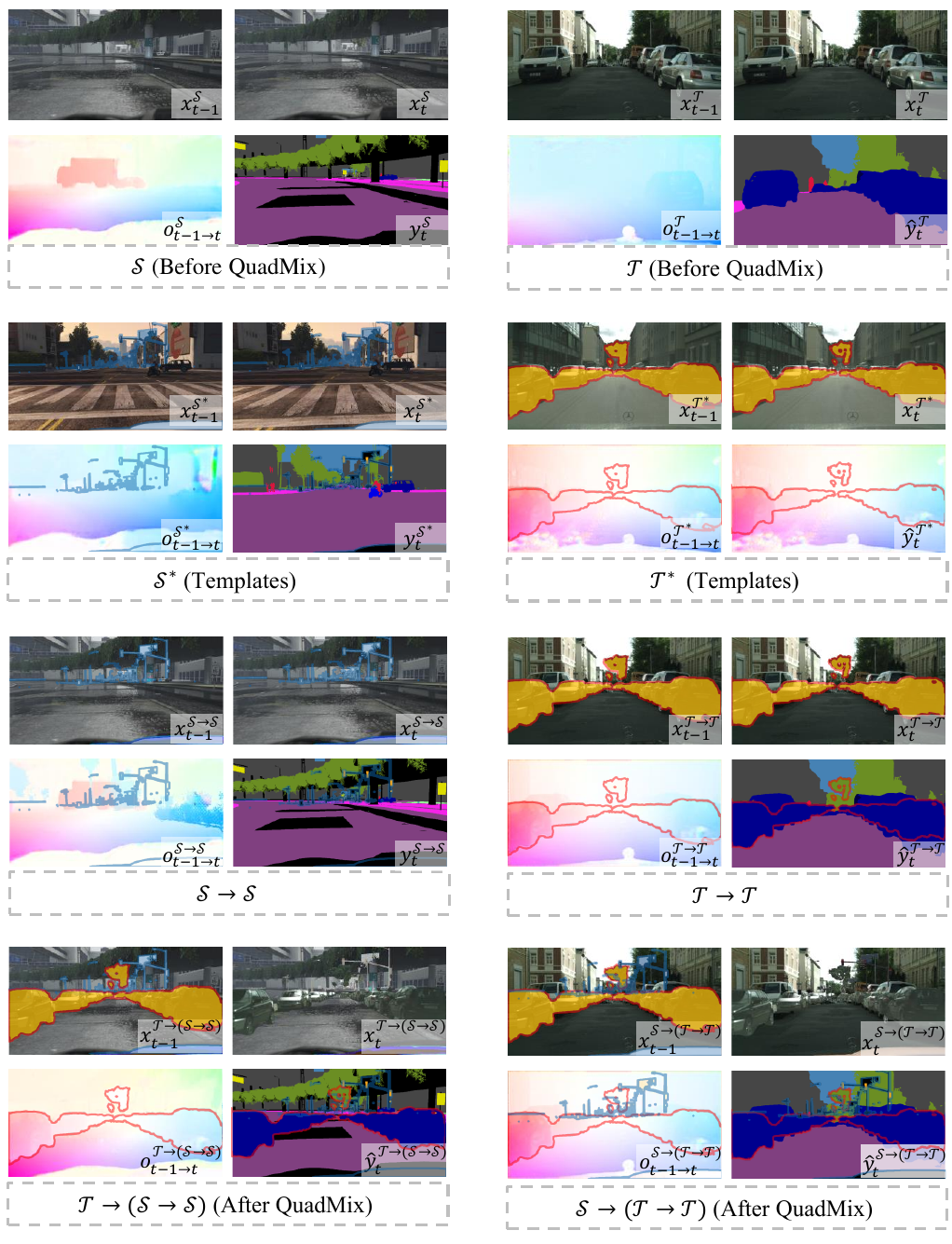}
	\caption{Visual illustration 1 of the various mixing strategies employed in QuadMix on the \textbf{Viper $\rightarrow$ Cityscapes-Seq} benchmark. The final augmentation of frame $t$ is shown without masks ($x_t^{\mathcal{T} \rightarrow (\mathcal{S}\rightarrow \mathcal{S})}$ and $x_t^{\mathcal{S} \rightarrow (\mathcal{T}\rightarrow \mathcal{T})}$) to better highlight the effects of QuadMix.}
	\label{supp_fig. 5}
\end{figure*}

\begin{figure*}[!t]
	\centering

	\includegraphics[width=0.85\linewidth]{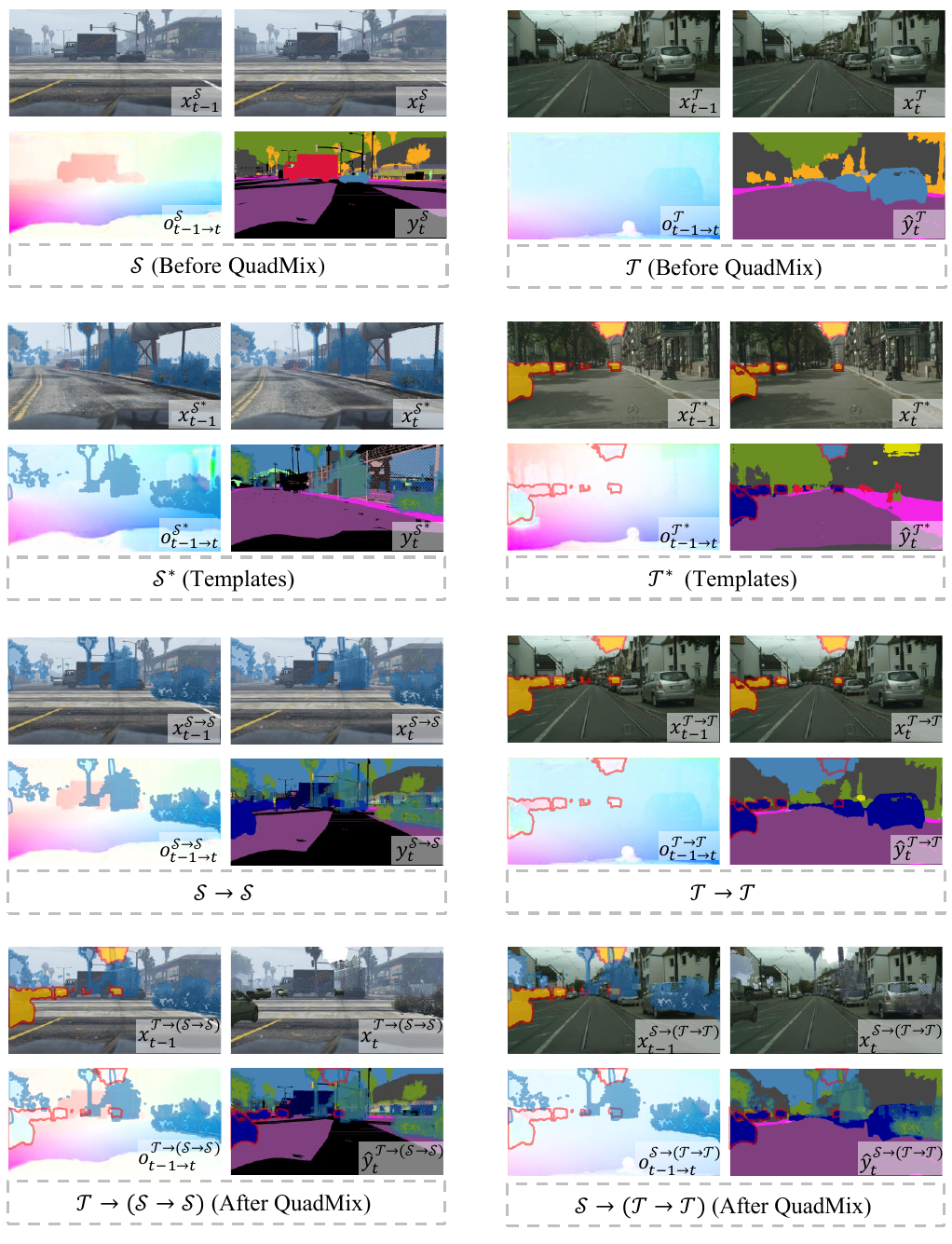}
	\caption{Visual illustration 2 of the various mixing strategies employed in QuadMix on the \textbf{Viper $\rightarrow$ Cityscapes-Seq} benchmark. The final augmentation of frame $t$ is shown without masks ($x_t^{\mathcal{T} \rightarrow (\mathcal{S}\rightarrow \mathcal{S})}$ and $x_t^{\mathcal{S} \rightarrow (\mathcal{T}\rightarrow \mathcal{T})}$) to better highlight the effects of QuadMix.}
	\label{supp_fig. 6}
\end{figure*}

\begin{figure*}[!t]
	\centering

	\includegraphics[width=0.85\linewidth]{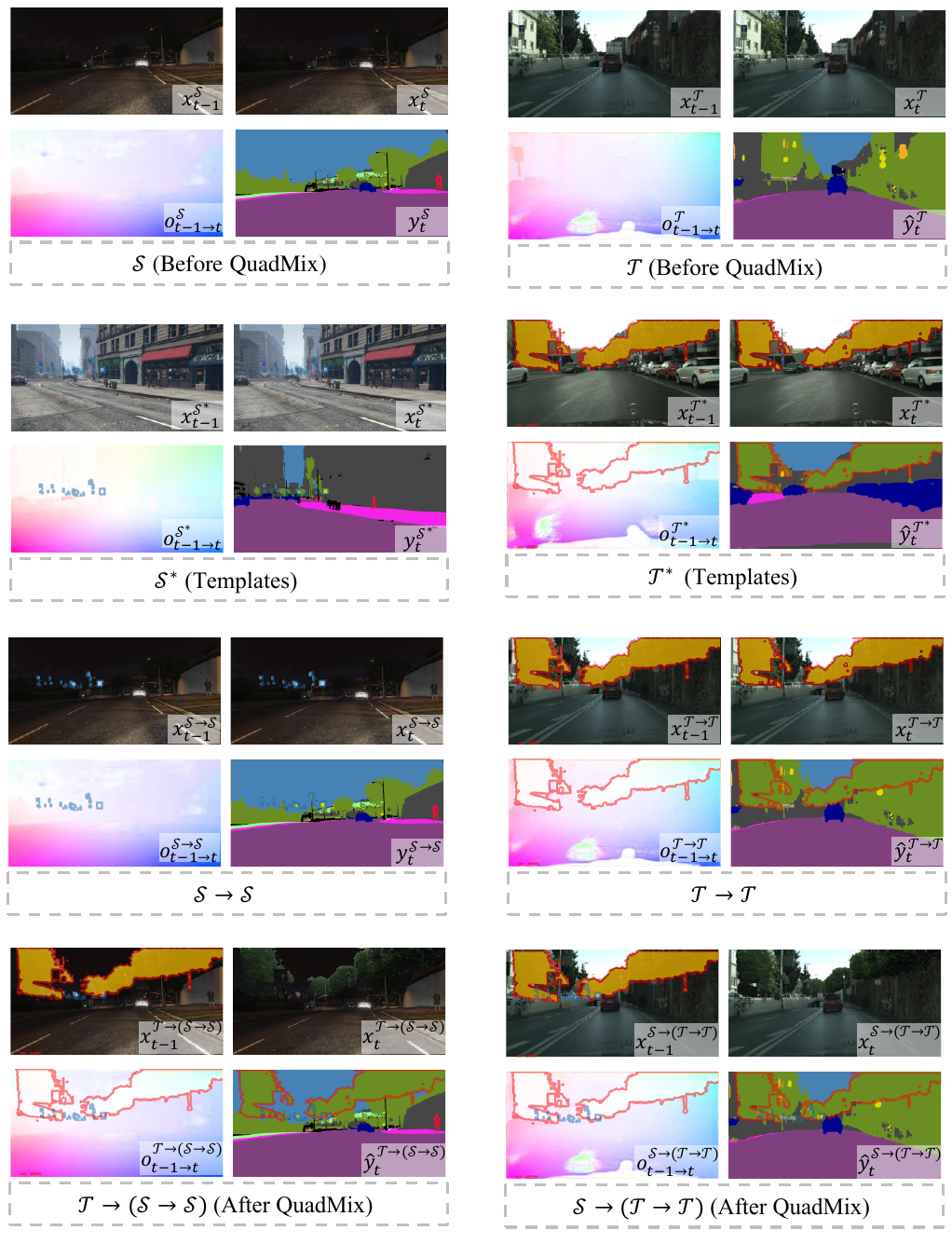}
	\caption{Visual illustration 3 of the various mixing strategies employed in QuadMix on the \textbf{Viper $\rightarrow$ Cityscapes-Seq} benchmark. The final augmentation of frame $t$ is shown without masks ($x_t^{\mathcal{T} \rightarrow (\mathcal{S}\rightarrow \mathcal{S})}$ and $x_t^{\mathcal{S} \rightarrow (\mathcal{T}\rightarrow \mathcal{T})}$) to better highlight the effects of QuadMix.}
	\label{supp_fig. 7}
\end{figure*}

\begin{figure*}[!t]
	\centering

	\includegraphics[width=0.85\linewidth]{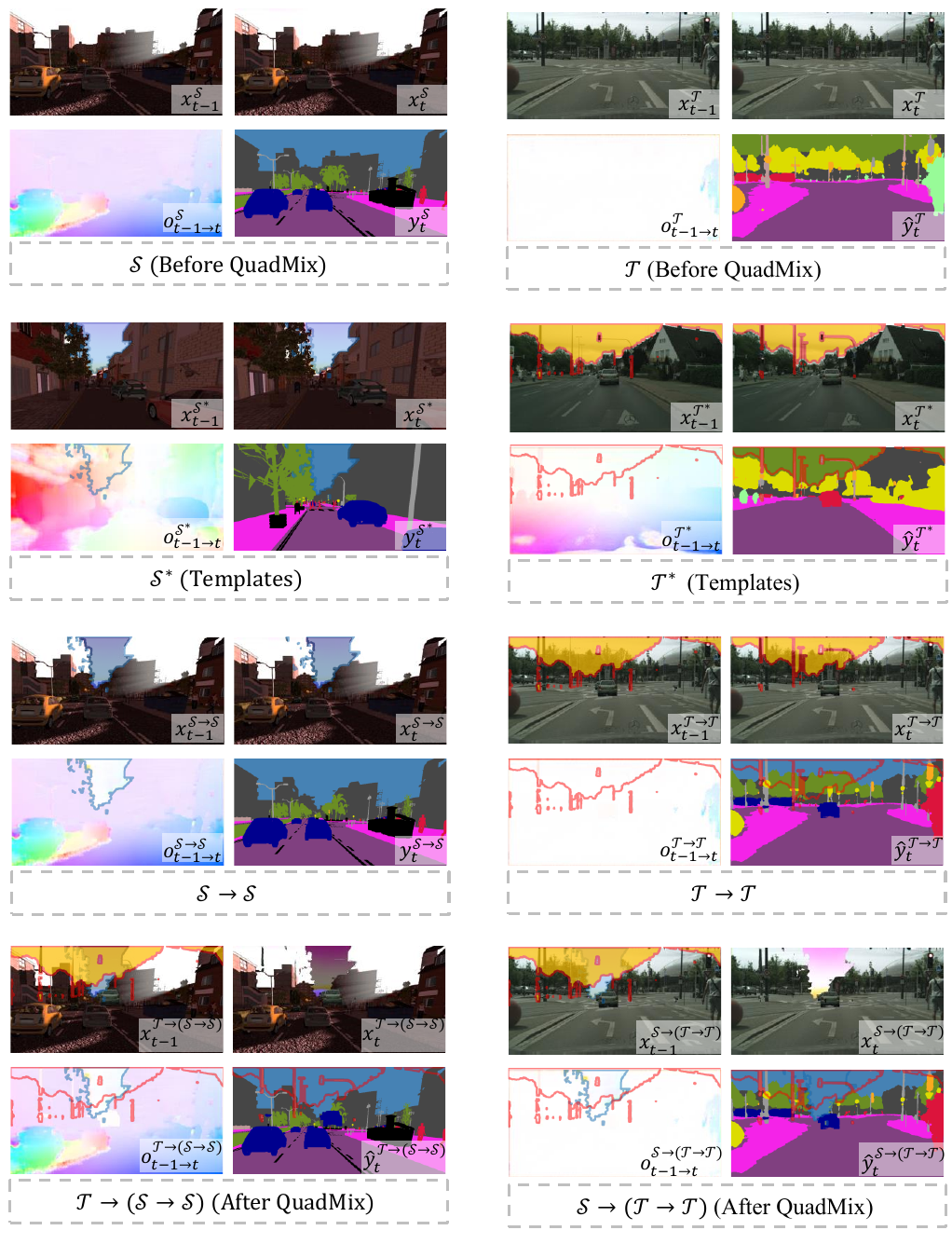}
	\caption{Visual illustration 1 of the various mixing strategies employed in QuadMix on the \textbf{SYNTHIA-Seq $\rightarrow$ Cityscapes-Seq} benchmark. The final augmentation of frame $t$ is shown without masks ($x_t^{\mathcal{T} \rightarrow (\mathcal{S}\rightarrow \mathcal{S})}$ and $x_t^{\mathcal{S} \rightarrow (\mathcal{T}\rightarrow \mathcal{T})}$) to better highlight the effects of QuadMix.}
	\label{supp_fig. 8}
\end{figure*}

\begin{figure*}[!t]
	\centering

	\includegraphics[width=0.85\linewidth]{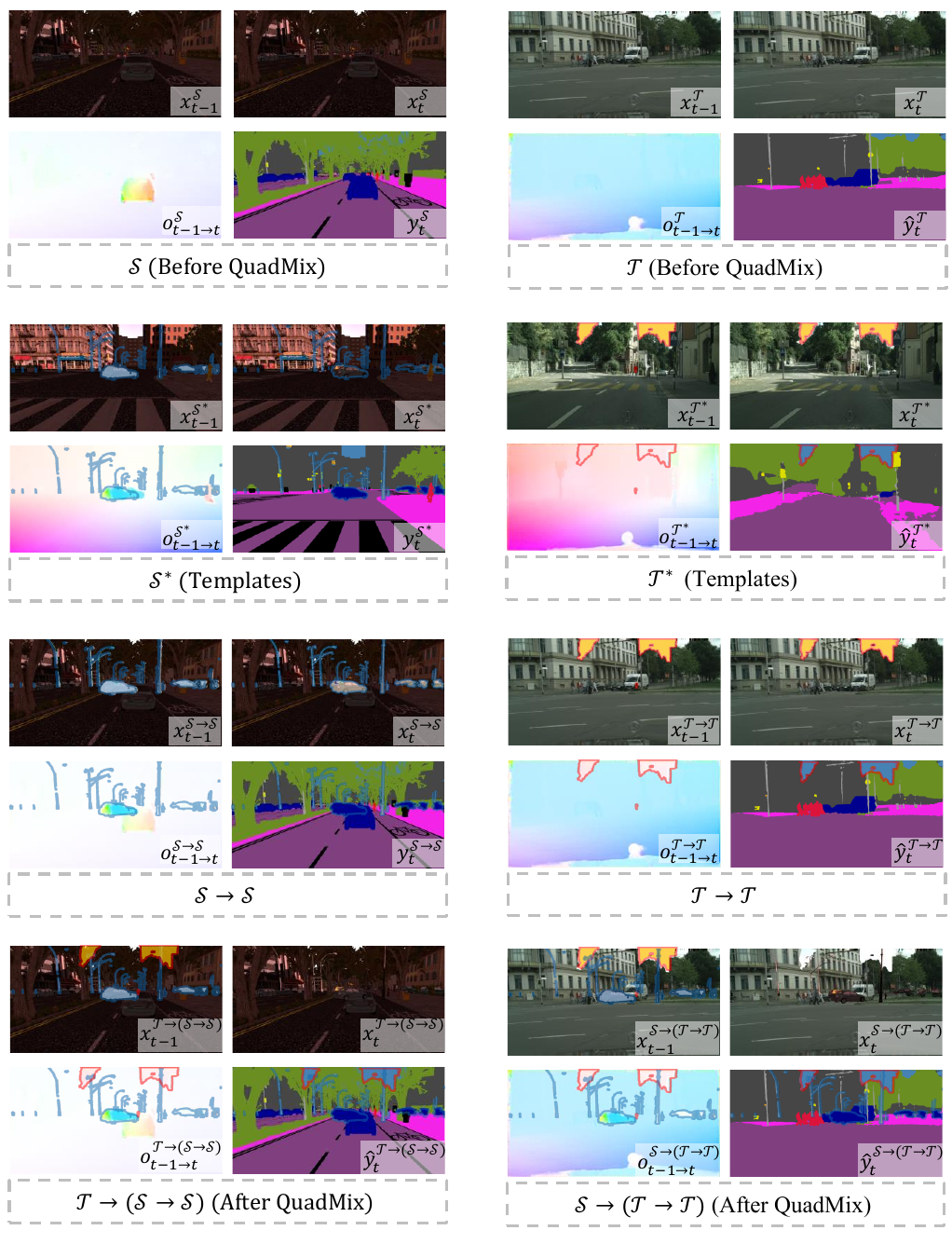}
	\caption{Visual illustration 2 of the various mixing strategies employed in QuadMix on the \textbf{SYNTHIA-Seq $\rightarrow$ Cityscapes-Seq} benchmark. The final augmentation of frame $t$ is shown without masks ($x_t^{\mathcal{T} \rightarrow (\mathcal{S}\rightarrow \mathcal{S})}$ and $x_t^{\mathcal{S} \rightarrow (\mathcal{T}\rightarrow \mathcal{T})}$) to better highlight the effects of QuadMix.}
	\label{supp_fig. 9}
\end{figure*}

\begin{figure*}[!t]
	\centering

	\includegraphics[width=0.85\linewidth]{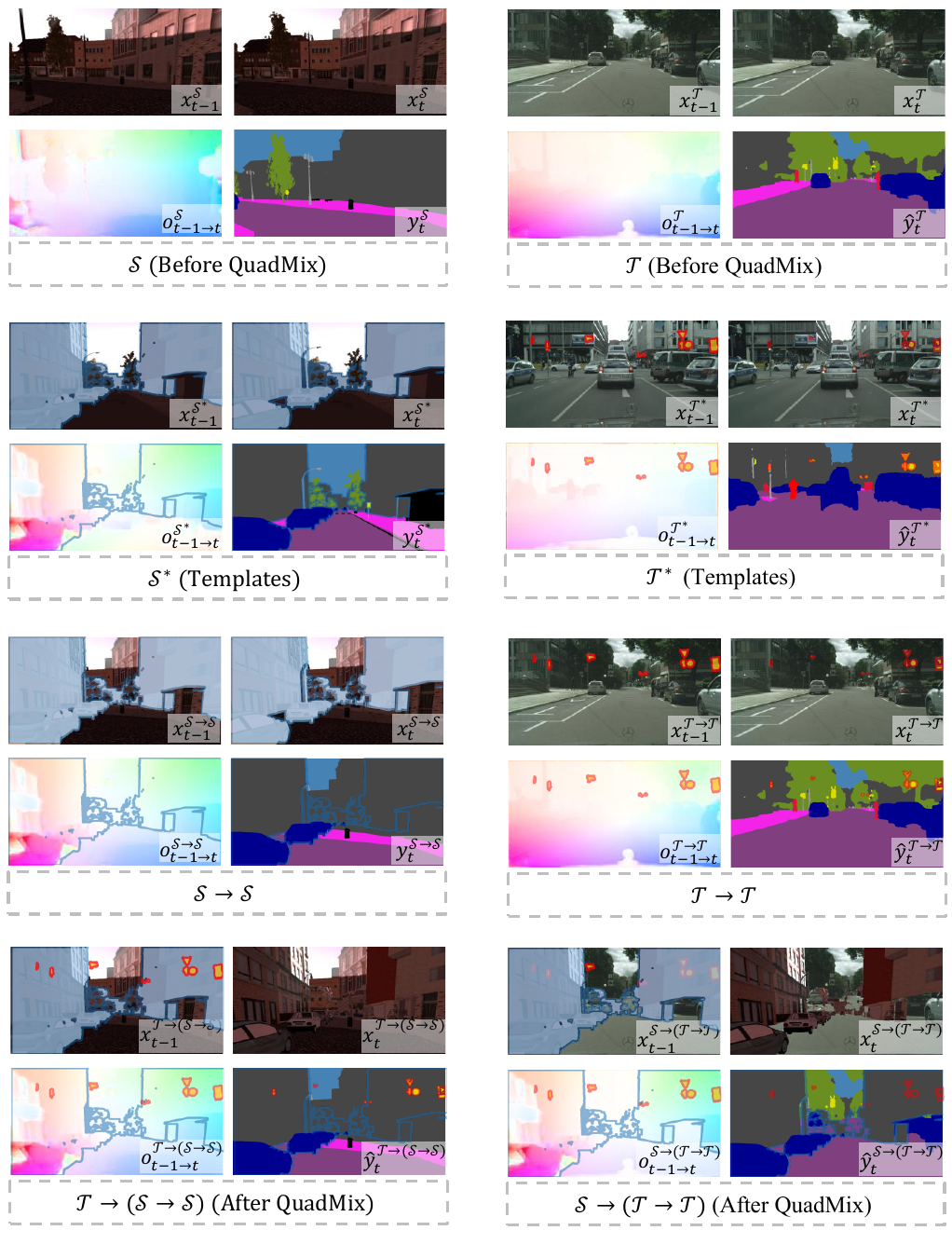}
	\caption{Visual illustration 3 of the various mixing strategies employed in QuadMix on the \textbf{SYNTHIA-Seq $\rightarrow$ Cityscapes-Seq} benchmark. The final augmentation of frame $t$ is shown without masks ($x_t^{\mathcal{T} \rightarrow (\mathcal{S}\rightarrow \mathcal{S})}$ and $x_t^{\mathcal{S} \rightarrow (\mathcal{T}\rightarrow \mathcal{T})}$) to better highlight the effects of QuadMix.}
	\label{supp_fig. 10}
\end{figure*}

We visualize the evolution of pseudo-label quality throughout training to illustrate how their accuracy improves across different stages. As shown in Fig.~\ref{supp_fig. 3} (with CNN backbone) and Fig.~\ref{supp_fig. 4} (with ViT backbone), the pseudo-labels undergo progressive refinement, increasingly aligning with the underlying structure of the target domain.

\subsection{Ablation Study on Image UDA-SS} 
For the image UDA-SS task, we also conducted ablation experiments on the GTA $\rightarrow$ Cityscapes dataset (using the CNN architecture). The comparison includes spatial feature aggregation for domain alignment, I-template, and F-template. As shown in Table~\ref{supp_tab:table5} and Table~\ref{supp_tab:table6}, all three components are effective and work better in combination, resulting in complementary benefits and enhanced domain alignment.

\subsection{Effectiveness of Temporal Information for Video UDA-SS}
For the video UDA-SS task, leveraging temporal information improves both consistency and feature learning by capturing dependencies across consecutive frames. As shown in Table~\ref{supp_tab:table0}, our ablation studies demonstrate the effectiveness of two key components: optical flow warp-based pseudo-label generation, as well as the optical flow-based QuadMix. By aligning features over time, these strategies improve temporal coherence and feature representation, resulting in more robust and stable performance and ultimately enhancing adaptation to the target video domain.

\subsection{Quantitative Impact of the Number of Mixed Categories}
We conducted quantitative experiments to evaluate the impact of varying the number of mixed categories (0, 1, 2, 3, 4) from each domain on the SYNTHIA-Seq $\rightarrow$ Cityscapes-Seq benchmark, as shown in Table \ref{supp_tab:table1}. The results indicate that the proposed QuadMix maintains strong performance as the number of mixed categories varies across domains.

\subsection{Further Illustration of Temporal Cues in Video UDA-SS}
\begin{figure*}[!h]
	\centering

	\includegraphics[width=\linewidth]{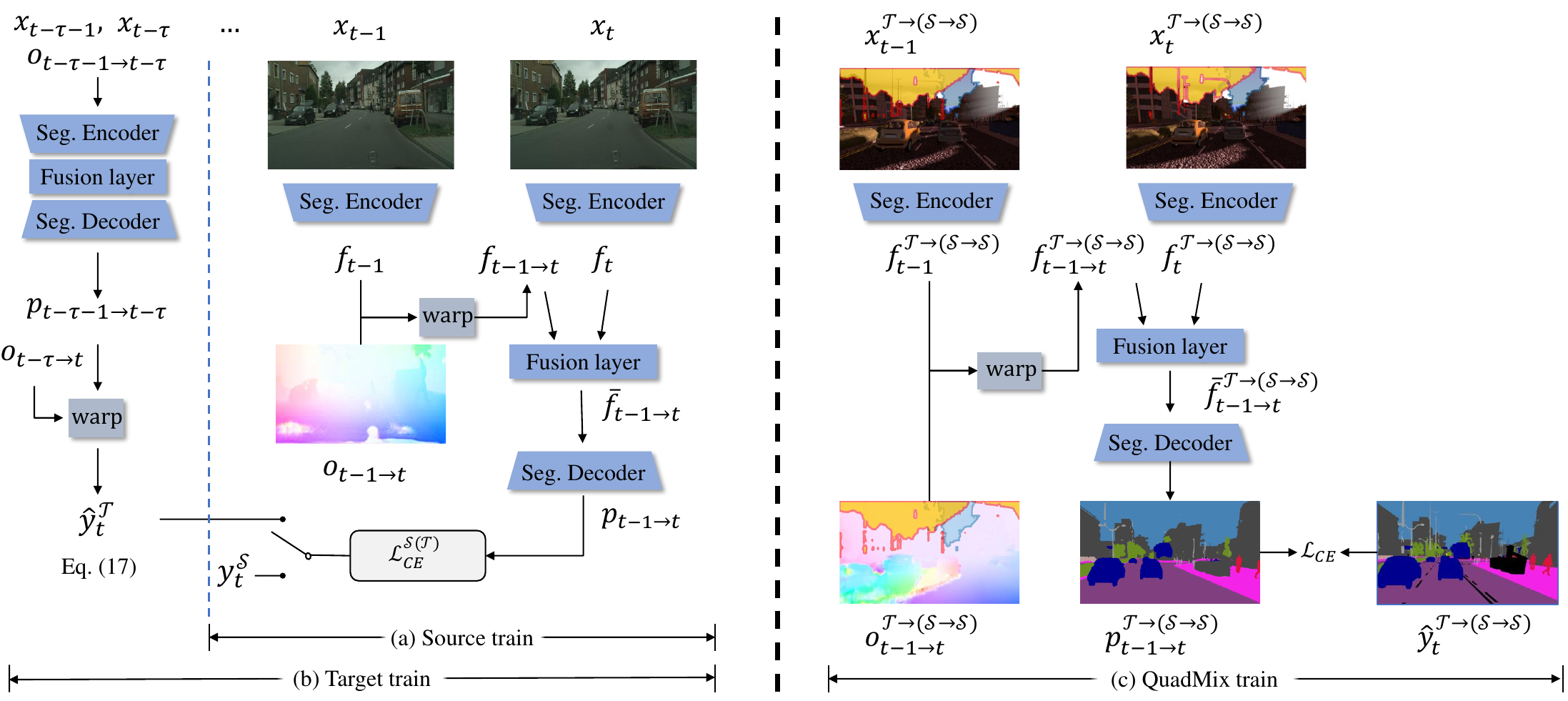}
	\caption{Illustration of video pseudo-label generation, and the role of temporal information (i.e., optical flow) in the model's learning of QuadMixed video features.}
	\label{supp_fig. 12}
\end{figure*}

Fig. \ref{supp_fig. 12} further illustrates the role of temporal information in pseudo-label generation (Eq. 15, Eq. 17, and Eq. 18) and quad-directional mixed sample learning (Eq. 12). This visualization offers a clearer insight into how temporal dependencies across consecutive frames enhance pseudo-label accuracy and facilitate more effective spatio-temporal learning in video UDA-SS.

\subsection{Visualization of Image/Video Frame Enhancement}
\begin{figure*}[!t]
	\centering
\includegraphics[width=1.0\linewidth]{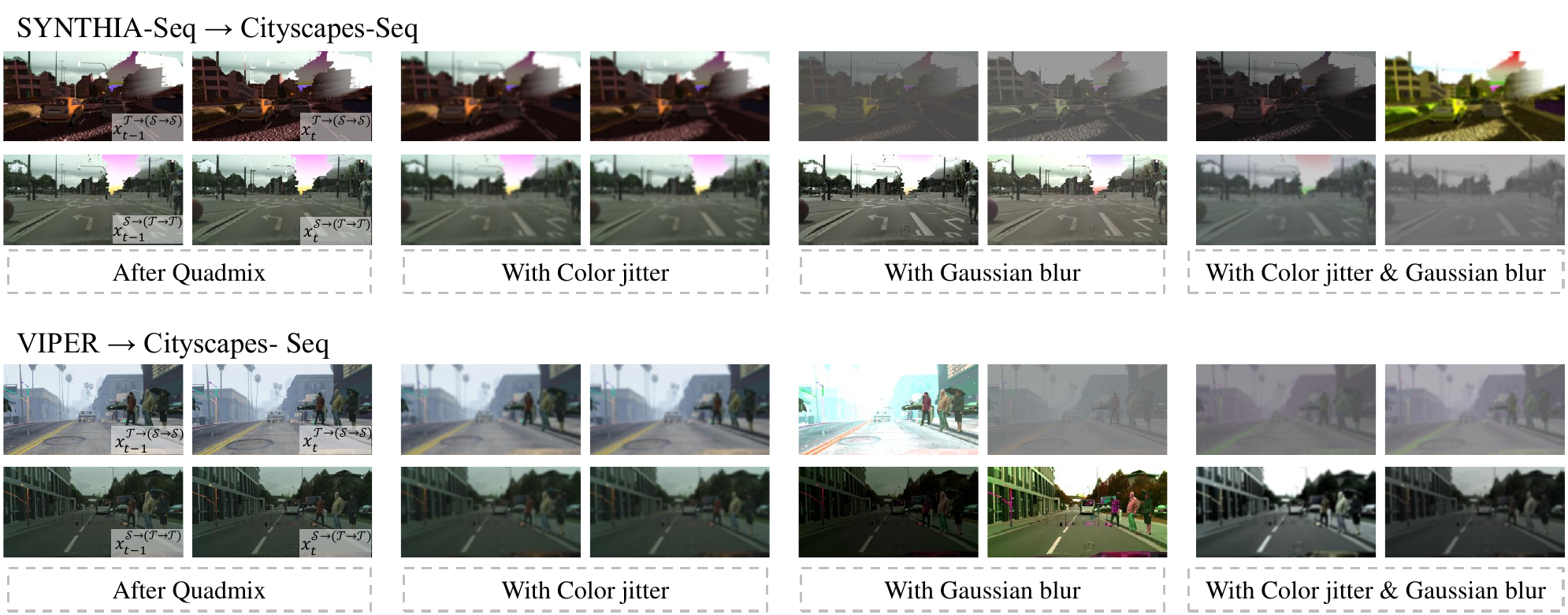}
	\caption{Visualization of video frame enhancement (random Gaussian blur or color jitter) on two video UDA-SS benchmarks.}
	\label{supp_fig. 11}
\end{figure*}

To promote consistent learning, we apply random Gaussian blur and color jitter, as defined in Eq. 22 and Eq. 24, with an 80\% probability and varying intensities for each image or video frame. Fig.~\ref{supp_fig. 11} presents visual comparisons of QuadMix results before and after enhancement for better understanding.

\subsection{Detailed Analysis of Category-Aware IoU Results}
We present the category-aware IoU results corresponding to the ablation studies in Sections IV-C and IV-D of the main paper. These results are reported in Tables~\ref{supp_tab:table111} and~\ref{supp_tab:table112}. For clarity, all values are rounded to one decimal place.


			
			

			

\begin{table*}
	\caption{Ablation studies on \textbf{image UDA-SS} using the GTAV $\rightarrow$ Cityscapes benchmark. ``I-template" represents the image patch template, ``F-template" refers to feature-level template mixing across the quad-mixed domains, and ``Agg." denotes the aggregated image feature alignment.}
		\label{supp_tab:table5}
	\centering
	\renewcommand\arraystretch{1.45}
	\setlength{\tabcolsep}{1.5mm}
	\scalebox{1.0}{
		\begin{tabular}{c|ccccc|c}
			\hline
			ID & Agg. & I-template & F-template  & mIOU (Gain) \\ \hline

			0                                                     & \checkmark                                             &                  &         & 64.41  \\
			1                                                    & \checkmark                                     & \checkmark                 &         &  65.67  \\

			2                                                                        & \checkmark     &   \checkmark               & \checkmark       & 66.76  \\
			
		\hline
	\end{tabular}}
\end{table*}

\begin{table*}
	\caption{Category-aware IoU results corresponding to the ablation study presented in Table~\ref{supp_tab:table5}.
		\label{supp_tab:table6}}
	\renewcommand{\arraystretch}{1.6}
    	\setlength{\tabcolsep}{1.5mm}

	\centering
	\small
	\scalebox{0.8}{
		\begin{tabular}{c|ccccccccccccccccccc|c}
		\hline
			\multicolumn{1}{c|}{ID} &  Road & Side. & Buil. & Wall & Fence & Pole & Light & Sign & Vege. & Terr. & Sky  & Pers. & Rider & Car  & Truck & Bus  & Train & Moto. & \multicolumn{1}{c|}{Bike} & mIOU \\ \hline

   \multicolumn{1}{c|}{0 }     & 96.5 & 74.1   &  89.1  & 39.8 & 31.3  & 50.7  & 59.9  & 69.5 &  90.6  & 48.5  &   92.8   &  78.4   &  51.4   &   92.9   &  66.1   & 63.3 & 1.8   & 58.3  & \multicolumn{1}{c|}{68.7} & 64.41 \\

   \multicolumn{1}{c|}{1 }     & 96.6 & 77.7   &  90.4  & 47.3 & 37.9  & 54.2  & 62.8  & 71.4 &  90.8  & 50.3  &   90.6   &  78.2   &  43.4  &   93.3   &  70.2   & 64.6 & 0.3   & 59.3  & \multicolumn{1}{c|}{68.3} & 65.67 \\

   \multicolumn{1}{c|}{2 }     & 97.1 & 78.0 & 90.4 & 49.7 & 40.3  & 53.4 & 61.7 & 70.9  & 90.7 &  49.7 &  92.9 &  77.9  &  53.9   &  93.5  &  72.4 & 65.9 & 0.7 & 60.5 & \multicolumn{1}{c|}{ 68.5 } & 66.76 \\
   \hline
   
	\end{tabular}}
\end{table*}

\begin{table*}
	\caption{Analysis of Temporal Information in Video UDA-SS on the SYNTHIA-Seq $\rightarrow$ Cityscapes-Seq Benchmark.
		\label{supp_tab:table0}}
  \setlength{\tabcolsep}{2.2mm}
	\centering
	\renewcommand{\arraystretch}{1.3}
	\footnotesize
	\scalebox{0.95}{
		\begin{tabular}{c|cc|ccccccccccc|c}
			\hline
            \multicolumn{1}{c|}{ID}          & Warp  & \multicolumn{1}{c|}{Flow Mix} & Road & Side. & Buil. & Pole & Light & Sign & Vege. & Sky  & Pers. & Rider & \multicolumn{1}{c|}{Car}  & mIOU \\ \hline
			0               & \checkmark         & \multicolumn{1}{c|}{}    & 88.2 & 18.6  & 79.8  & 29.0 & 26.5  & 44.7 & 82.2  & 83.8 & 58.6  & 35.3  & \multicolumn{1}{c|}{84.4} & 57.39  \\

            1               &          & \multicolumn{1}{c|}{\checkmark}    & 89.2 & 43.8  & 81.0  & 30.1 & 20.9  & 45.0 & 82.3  & 82.6 & 60.1  & 26.9  & \multicolumn{1}{c|}{87.2} & 59.01  \\

            \multicolumn{1}{c|}{2}              & \checkmark        & \multicolumn{1}{c|}{\checkmark}    & 90.8 & 39.9  & 83.2  & 33.2 & 30.1  & 50.7 & 84.8  & 82.3 & 61.2  & 32.7  & \multicolumn{1}{c|}{87.4} & 61.48  \\ \hline

           \hline
	\end{tabular}}
\end{table*}

\begin{table*}
	\caption{Ablation Studies on the Number of Mixed Categories from Source and Target Domains on the SYNTHIA-Seq $\rightarrow$ Cityscapes-Seq.
		\label{supp_tab:table1}}
  \setlength{\tabcolsep}{2.5mm}
	\centering
	\renewcommand{\arraystretch}{1.3}
	\footnotesize
	\scalebox{1.0}{
		\begin{tabular}{cccccccccccccccc}
		\\ \hline
			 \multicolumn{1}{c|}{Mixed Number} & Road & Side. & Buil. & Pole & Light & Sign & Vege. & Sky  & Pers. & Rider & \multicolumn{1}{c|}{Car}  & mIOU \\ \hline
			\multicolumn{1}{c|}{0}     & 89.4 & 40.3  & 80.7  & 28.2 & 20.8   & 50.8 & 81.5  & 83.8 & 60.8 & 28.7  & \multicolumn{1}{c|}{87.5} & 59.31 \\ 
			
			\multicolumn{1}{c|}{1}     & 91.7 & 47.1  &  80.8  &  29.5 &  27.2  &  47.2  &   81.0  &  81.5  & 58.3  &  26.3   & \multicolumn{1}{c|}{88.0} &  59.87  \\ 
			 \multicolumn{1}{c|}{2}    & 90.8 & 39.9  & 83.2  & 33.2 & 30.1  & 50.7 & 84.8  & 82.3 & 61.2  & 32.7  & \multicolumn{1}{c|}{87.4} & 61.48  \\
           \multicolumn{1}{c|}{3}    & 92.5 & 51.5  & 81.9 & 29.8 & 25.9  & 47.5 & 84.0  & 82.6 & 60.0  & 28.5  & \multicolumn{1}{c|}{87.3} & 61.05  \\
            \multicolumn{1}{c|}{4}    & 91.5 & 45.4  & 82.1  & 31.3 & 22.2  & 16.5 & 84.5  & 82.9 & 61.9  & 27.6  & \multicolumn{1}{c|}{87.3} & 60.29  \\
            \hline
	\end{tabular}}
\end{table*}

\begin{table*}[t]
	\caption{Detailed experimental results regarding category-aware IOU and mIOU on Synthia-Seq $\rightarrow$ Cityscapes-Seq.
		\label{supp_tab:table111}}
	\centering
	\renewcommand{\arraystretch}{1.47}
	\footnotesize
	\scalebox{1.05}{
		\begin{tabular}{ccccccccccccc}
			\toprule[1.0pt]
			\multicolumn{13}{c}{Synthia-Seq $\rightarrow$ Cityscapes-Seq}                                                                                                                       \\ \hline
			\multicolumn{1}{c|}{Methods}           & road & side. & buil. & pole & light & sign & vege. & sky  & pers. & rider & \multicolumn{1}{c|}{car}  & mIOU \\ \hline

        \multicolumn{1}{c|}{QuadMix (ViT)}          & 94.1 & 61.9  & 82.9  & 36.9 & 41.0 & 59.1 & 85.2  & 85.6 & 64.3  & 37.8 & \multicolumn{1}{c|}{90.3} & \textbf{67.19} \\  \hline
        
			\multicolumn{1}{c|}{ QuadMix(DL-V2)}          & 90.8 & 39.9  &  83.2   &  33.2  &  30.1   &  50.7 &  84.8  & 82.3 &  61.2   &  32.7  & \multicolumn{1}{c|}{ 87.4}  & 61.48  \\

   \multicolumn{1}{c|}{w/o Long-tail (DL-V2)}          & 90.2 & 48.3  & 82.8  & 31.5 & 27.6  & 46.6 & 83.5  & 81.7 & 60.5  & 27.2  & \multicolumn{1}{c|}{88.0} & 60.72  \\ \hline
   
   \multicolumn{1}{c|}{Table V-0*}          & 89.7 & 47.1  & 73.2  & 20.2 & 14.0  & 29.6 & 69.4  & 69.2 & 55.6  & 19.6  & \multicolumn{1}{c|}{84.5} & 52.01  \\

   \multicolumn{1}{c|}{Table V-1}          & 90.0 & 48.3  & 73.0  & 29.9 & 18.1  & 41.6 & 73.7  & 72.6 & 58.0  & 22.5  & \multicolumn{1}{c|}{85.1} & 55.71 \\ 

   \multicolumn{1}{c|}{Table V-2}          & 87.5 & 44.7  & 75.4  & 34.1 & 23.2  & 44.7 & 78.9  & 76.8 & 58.7  & 24.9  & \multicolumn{1}{c|}{80.9} & 57.25  \\ 

   \multicolumn{1}{c|}{Table V-3}          & 85.7 & 43.4  & 75.3  & 25.6 & 19.0  & 37.1 & 73.2  & 69.3 & 58.9  & 21.1  & \multicolumn{1}{c|}{85.5} & 54.01 \\ 

   \multicolumn{1}{c|}{Table V-4}          & 92.5 & 43.1  & 81.7  & 28.9 & 22.1  & 49.5 & 80.1  & 84.4 & 62.1  & 26.4  & \multicolumn{1}{c|}{85.0} & 59.62  \\ 

   \multicolumn{1}{c|}{Table V-5}          & 90.1 & 44.0  & 78.3  & 30.5 & 26.0  & 42.1 & 79.2  & 72.5 & 54.6  & 33.4  & \multicolumn{1}{c|}{86.7} & 57.95  \\ 

    \multicolumn{1}{c|}{Table V-6}          & 89.2 & 43.8  & 78.2  & 36.4 & 22.9  & 38.1 & 72.0  & 70.7 & 54.5  & 25.3  & \multicolumn{1}{c|}{84.6} & 55.97 \\ 
    
   \multicolumn{1}{c|}{Table V-7}          & 91.9 & 54.1  & 79.5  & 32.7 & 25.2  & 42.5 & 80.9  & 76.1 & 60.3  & 34.4  & \multicolumn{1}{c|}{85.3} & 60.26  \\ 

   \multicolumn{1}{c|}{Table V-8}          & 90.8 & 39.9  &  83.2   &  33.2  &  30.1   &  50.7 &  84.8  & 82.3 &  61.2   &  32.7  & \multicolumn{1}{c|}{ 87.4}  &  61.48\\  \hline

   \multicolumn{1}{c|}{Table VI-0*}          & 89.2 & 43.8  & 78.2  & 36.4 & 22.9  & 38.1 & 72.0  & 70.7 & 54.5  & 25.3  & \multicolumn{1}{c|}{84.6} & 55.97  \\ 

   \multicolumn{1}{c|}{Table VI-1}          & 89.7 & 55.0  & 80.8  & 30.1 & 25.0  & 42.2 & 80.4  & 69.6 & 57.7  & 25.9  & \multicolumn{1}{c|}{82.8} & 58.11  \\ 

   \multicolumn{1}{c|}{Table VI-2}          & 89.3 & 29.9  & 81.5  & 24.3 & 25.3  & 45.0 & 82.8  & 83.7 & 60.5  & 26.8  & \multicolumn{1}{c|}{87.2} & 57.84  \\ 

   \multicolumn{1}{c|}{Table VI-3}          & 91.2 & 47.2  & 81.3  & 27.0 & 16.6  & 40.5 & 82.9  & 83.1 & 58.4  & 30.8  & \multicolumn{1}{c|}{87.5} & 58.77  \\ 

   \multicolumn{1}{c|}{Table VI-4}          & 90.4 & 44.1  & 79.4  & 33.4 & 26.9  & 45.7 & 80.6  & 73.8 & 60.4  & 27.7  & \multicolumn{1}{c|}{86.4} & 58.98  \\ 

   \multicolumn{1}{c|}{Table VI-5}          & 92.1 & 52.5  & 78.3  & 21.2 & 16.4  & 42.8 & 78.5  & 75.9 & 56.6  & 23.1  & \multicolumn{1}{c|}{86.0} & 56.67  \\ 

   \multicolumn{1}{c|}{Table VI-6}          & 89.0 & 49.1  & 80.2  & 30.9 & 20.9  & 45.7 & 80.3  & 79.2 & 59.2  & 31.9  & \multicolumn{1}{c|}{85.2} & 59.24  \\ 

   \multicolumn{1}{c|}{Table VI-7}          & 91.3 & 53.2  & 81.5  & 28.3 & 22.3  & 42.0 & 80.3  & 74.3 & 57.8  & 25.6  & \multicolumn{1}{c|}{85.7} & 58.39  \\ 

   \multicolumn{1}{c|}{Table VI-8}          & 90.8 & 43.9  & 80.4  & 28.6 & 24.5  & 45.3 & 81.5  & 81.1 & 60.2  & 30.6  & \multicolumn{1}{c|}{88.1} & 59.55  \\ 

   \multicolumn{1}{c|}{Table VI-9}          & 91.5 & 45.4  & 81.1  & 31.3 & 22.2  & 46.5 & 84.5  & 82.9 & 61.9  & 27.6  & \multicolumn{1}{c|}{87.3} & 60.20  \\ 

   \multicolumn{1}{c|}{Table VI-10}          & 92.2 & 54.4  & 80.3  & 24.5 & 21.3  & 45.8 & 81.8  & 83.7 & 60.4  & 21.5  & \multicolumn{1}{c|}{87.4} & 59.39  \\ 

   \multicolumn{1}{c|}{Table VI-11}          & 91.9 & 54.1  & 79.5  & 32.7 & 25.2  & 42.5 & 80.9  & 76.1 & 60.3  & 35.8  & \multicolumn{1}{c|}{85.3} & 60.39  \\ 

   \multicolumn{1}{c|}{Table VI-12}           & 90.8 & 39.9  &  83.2   &  33.2  &  30.1   &  50.7 &  84.8  & 82.3 &  61.2   &  32.7  & \multicolumn{1}{c|}{87.4}  &  61.48 \\ \hline

   \multicolumn{1}{c|}{Table VII-Things}          & 90.6 & 51.6  & 75.2  & 31.2 & 32.1  & 49.9 & 75.9  & 62.4 & 61.6  & 29.2  & \multicolumn{1}{c|}{84.1} & 58.53  \\ 

   \multicolumn{1}{c|}{Table VII-Stuffs}          & 91.9 & 54.6  & 82.2  & 23.8 & 23.4  & 40.8 & 81.2  & 75.3 & 55.0  & 18.1  & \multicolumn{1}{c|}{86.2} & 57.50  \\ 

   \multicolumn{1}{c|}{Table VII-Move}          & 92.2 & 58.4  & 77.7  & 30.7 & 23.7  & 40.3 & 78.0  & 67.1 & 60.6  & 29.3  & \multicolumn{1}{c|}{83.2} & 58.29  \\ 

   \multicolumn{1}{c|}{Table VII-Station}          & 90.5 & 53.7  & 71.0  & 30.1 & 29.3  & 52.0 & 71.7  & 38.4 & 61.1  & 27.3  & \multicolumn{1}{c|}{85.9} & 55.55 \\ 
   
   \multicolumn{1}{c|}{Table VII-All}        & 90.8 & 39.9  &  83.2   &  33.2  &  30.1   &  50.7 &  84.8  & 82.3 &  61.2   &  32.7  & \multicolumn{1}{c|}{ 87.4}  &  61.48 \\ \hline

   \multicolumn{1}{c|}{Table VIII-w/o $\mathcal{A}$}          & 89.9 & 38.0  & 82.2  & 25.0 & 28.0  & 50.5 & 83.6  & 83.5 & 61.1  & 33.0  & \multicolumn{1}{c|}{87.7} & 60.23  \\ 

   \multicolumn{1}{c|}{Table VIII-global}          & 91.0 & 50.9  & 81.0  & 23.9 & 24.7  & 43.1 & 81.1  & 76.6 & 57.2  & 24.5  & \multicolumn{1}{c|}{86.7} & 58.24  \\  \hline

   \multicolumn{1}{c|}{Table X-ResNet-Origin}          & 77.8 & 24.4  & 70.2  & 4.7 & 0.0  & 0.3 & 50.6  & 36.1 & 30.6  & 0.5  & \multicolumn{1}{c|}{57.0} & 32.01  \\  

   \multicolumn{1}{c|}{Table X-ResNet}          & 90.6 & 41.2  &  81.1  &  29.3 &  23.1  &  47.5  &   82.7  &  83.8  & 61.1  &  28.4   & \multicolumn{1}{c|}{87.0} &  59.62    \\  

   \multicolumn{1}{c|}{Table X-DA-VSN}          & 89.0 & 25.8  & 81.6  & 34.2 & 23.2  & 50.3 & 82.1  & 83.9 & 62.4  & 30.8  & \multicolumn{1}{c|}{86.5} & 59.07  \\

   \multicolumn{1}{c|}{Table X-CMOM}          & 90.7 & 42.6  & 83.2  & 29.3 & 27.2  & 48.8 & 84.4  & 81.2 & 59.4  & 30.2  & \multicolumn{1}{c|}{86.7} & 60.33  \\

   \multicolumn{1}{c|}{Table X-TPS}           & 90.8 & 39.9  &  83.2   &  33.2  &  30.1   &  50.7 &  84.8  & 82.3 &  61.2   &  32.7  & \multicolumn{1}{c|}{87.4}  &  61.48 \\

     \toprule[1.0pt]
	\end{tabular}}
\end{table*}

\begin{table*}[t]
	\caption{Detailed experimental results regarding category-aware IOU and mIOU on VIPER $\rightarrow$ Cityscapes-Seq.
		\label{supp_tab:table112}}
	\centering
	\renewcommand{\arraystretch}{1.6}
	
	\small
	\scalebox{0.9}{
		\begin{tabular}{ccccccccccccccccc}
  \toprule[1.0pt]
			\multicolumn{17}{c}{VIPER $\rightarrow$ Cityscapes-Seq}                                                                                                                                                              \\ \hline
			\multicolumn{1}{c|}{Methods}      &    road & side. & buil. & fenc. & light & sign & vege. & terr. & sky  & pers. & car & truc. & bus  & mot. & \multicolumn{1}{c|}{bike}  & mIOU \\ \hline
			
			\multicolumn{1}{c|}{QuadMix (ViT)}  & 87.3 & 43.8  &  87.3  & 25.2  & 40.0 & 36.9 &  86.7   & 20.8  &  90.3  & 65.8  &  86.8  &  48.6  &  65.6  & 37.6 & \multicolumn{1}{c|}{49.7} & {\textbf{58.16}} \\ \hline
            
            \multicolumn{1}{c|}{QuadMix (DL-V2)}    &  91.6  & 51.4  &  87.0   & 24.1  & 32.3  & 37.2 & 84.1  & 28.4  & 84.8 & 64.4  &  85.7  &  41.4  & 46.5 & 34.0 & \multicolumn{1}{c|}{49.6} & 56.17 \\ 

             \multicolumn{1}{c|}{w/o Long-tail (DL-V2)}    &                  91.9 & 51.8  & 85.7  & 25.2  & 26.4  & 36.8 &83.1  &28.9  & 86.2 & 63.3  & 84.6 & 45.4  & 45.7 & 32.7 & \multicolumn{1}{c|}{ 41.8}  &  55.31  \\ \hline

              \multicolumn{1}{c|}{Table VIII-w/o $\mathcal{A}$}    &                  91.9 & 50.2  & 86.5  & 26.1  & 27.3  & 35.6 &82.4  &27.9  & 85.8 & 64.2  & 83.9 & 46.1  & 42.3 & 33.4 & \multicolumn{1}{c|}{ 45.2}  &  55.25  \\

               \multicolumn{1}{c|}{Table VIII-global}    &                  91.2 & 43.4  & 85.4  & 19.6  & 29.5  & 35.9 &83.5  &28.4  & 85.2 & 65.6  & 85.6 & 38.5  & 35.1 & 33.5 & \multicolumn{1}{c|}{ 43.9}  &  53.62  \\  \hline

                \multicolumn{1}{c|}{Table IX-1}    &                  91.2 & 40.0  & 85.8  & 22.3  & 25.7  & 33.9 &83.8  &28.7  & 86.8 & 62.6  & 85.4 & 42.1  & 49.7 & 34.6 & \multicolumn{1}{c|}{ 45.9}  &  54.56  \\

                 \multicolumn{1}{c|}{Table IX-2}                &  91.6  & 51.4  &  87.0   & 24.1  & 32.3  & 37.2 & 84.1  & 28.4  & 84.8 & 64.4  &  85.7  &  41.4  & 46.5 & 34.0 & \multicolumn{1}{c|}{49.6} & 56.16\\ 

                  \multicolumn{1}{c|}{Table IX-3}    &                  90.5 & 49.2  & 83.7  & 21.5  & 22.3  & 30.3 &81.8  &28.4  & 87.0 & 58.7  & 86.1 & 47.6  & 51.3 & 33.2 & \multicolumn{1}{c|}{ 45.7}  &  54.49  \\

                   \multicolumn{1}{c|}{Table IX-4}    &                  90.8 & 49.8  & 86.2  & 25.5  & 26.6  & 33.3 &83.6  &29.0  & 85.4 & 63.5  & 84.5 & 37.4  & 46.0 & 34.2 & \multicolumn{1}{c|}{ 42.5}  &  54.56  \\

                    \multicolumn{1}{c|}{Table IX-5}    &                  92.4 & 51.6  & 86.8  & 25.2  & 29.4  & 29.4 &84.7  &32.5  & 86.7 & 63.9  & 83.1 & 29.8  & 43.6 & 33.0 & \multicolumn{1}{c|}{ 46.6}  &  54.58  \\

                     \multicolumn{1}{c|}{Table IX-6}    &                  92.2 & 54.8  & 86.1  & 29.2  & 27.0  & 35.0 &85.0  &32.8  & 85.6 & 63.6  & 82.0 & 25.0  & 48.1 & 34.0 & \multicolumn{1}{c|}{ 47.7}  &  55.21  \\ \hline

                      \multicolumn{1}{c|}{Table X-ResNet-Origin}    &                  75.1 & 6.5  & 72.8  & 0.0  & 0.0  & 0.0 &71.1  &1.1  & 58.4 & 11.2  & 63.0 & 9.8  & 0.0 & 0.0& \multicolumn{1}{c|}{0.0}  &  24.60 \\

                       \multicolumn{1}{c|}{Table X-ResNet}     &                  90.8 & 39.4  & 84.6  & 27.2  & 23.9  & 34.2 &82.4  &29.2  & 85.8 & 64.3  & 84.5 & 43.6  & 50.6 & 34.7 & \multicolumn{1}{c|}{ 48.8}  &   54.94  \\

                        \multicolumn{1}{c|}{Table X-DA-VSN}    &                  90.9 & 52.1  & 85.9  & 24.2  & 26.6  & 37.2 &82.5  &27.6  & 85.6 & 60.1  & 86.4 & 46.4  & 45.3 & 33.6 & \multicolumn{1}{c|}{ 45.3}  &  55.31\\
                        
                         \multicolumn{1}{c|}{Table X-CMOM}    &                  91.6 & 53.6  & 86.1  & 17.3  & 28.0  & 42.1 &86.4  &38.8  & 88.0 & 64.7  & 83.6 & 41.3  & 47.9 & 30.8 & \multicolumn{1}{c|}{ 41.4}  &  56.10  \\

                \multicolumn{1}{c|}{Table X-TPS}             &  91.6  & 51.4  &  87.0   & 24.1  & 32.3  & 37.2 & 84.1  & 28.4  & 84.8 & 64.4  &  85.7  &  41.4  & 46.5 & 34.0 & \multicolumn{1}{c|}{ 49.6} & 56.16  \\

     \toprule[1.0pt]
			
	\end{tabular}
 }
\end{table*}

\FloatBarrier
\clearpage

\bibliographystyle{IEEEtran}

\bibliography{Figures/refer}

\begin{IEEEbiography}[{\includegraphics[width=1in,height=1.25in,clip,keepaspectratio]{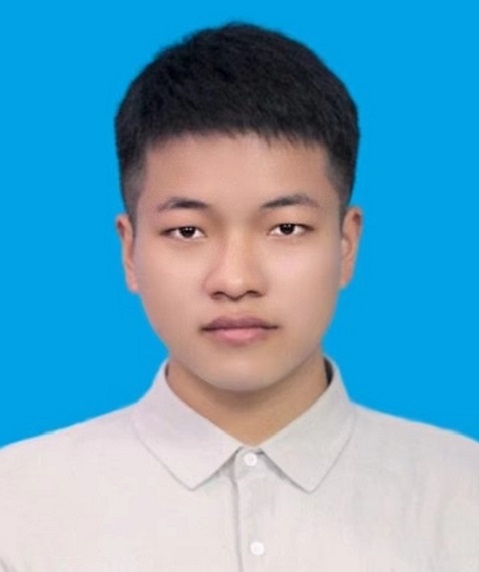}}]{Zhe Zhang}
	received the BS degree in the College of Information Science and Engineering, Northeastern University, China, in 2021. He is currently working toward a Ph.D. degree in the State Key Laboratory of Synthetical Automation for Process Industries, Northeastern University, Shenyang, China. His current research interests include computer vision, deep learning, video representation learning, multi-modal learning, and their applications in industrial fields.
\end{IEEEbiography}

\begin{IEEEbiography}[{\includegraphics[width=1in,height=1.25in,clip,keepaspectratio]{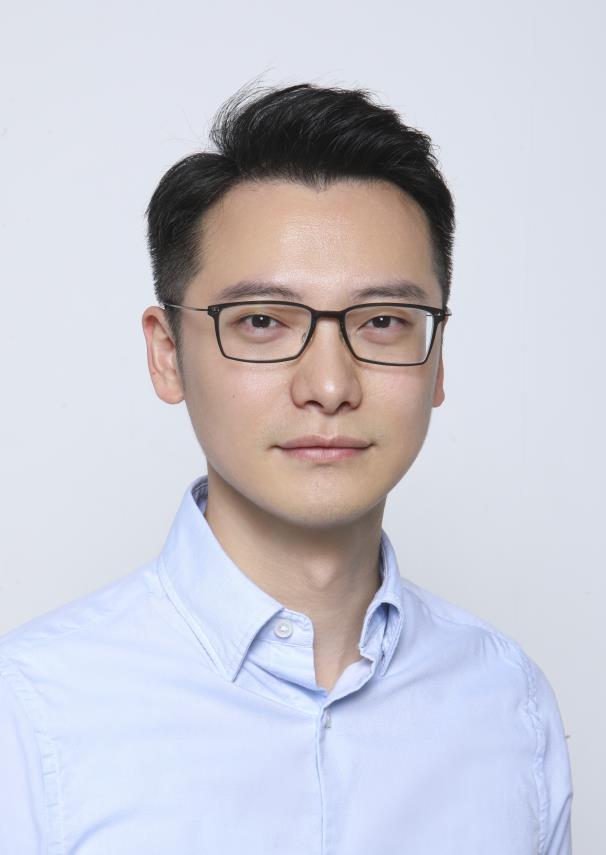}}]{Gaochang Wu}
	received the BE and MS degrees in mechanical engineering in Northeastern University, Shenyang, China, in 2013 and 2015, respectively, and Ph.D. degree in control theory and control engineering in Northeastern University, Shenyang, China in 2020. He is currently an associate professor in the State Key Laboratory of Synthetical Automation for Process Industries, Northeastern University. He was selected for the 2022-2024 Youth Talent Support Program of the Chinese Association of Automation. His current research interests include multimodal perception and recognition, as well as light field processing.
\end{IEEEbiography}

\begin{IEEEbiography}[{\includegraphics[width=1in,height=1.25in,clip,keepaspectratio]{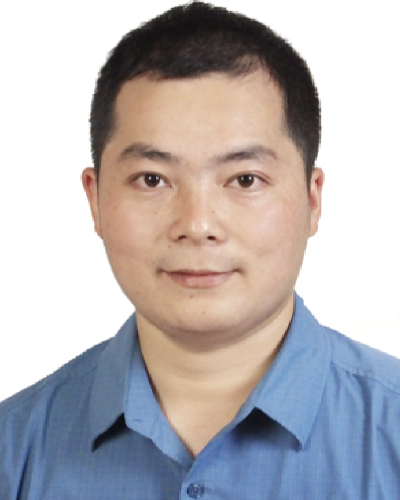}}]{Jing Zhang} (Senior Member, IEEE) is currently a professor at the School of Computer Science, Wuhan University, Wuhan, China. He previously served as a Research Fellow at the School of Computer Science, The University of Sydney. He has published over 100 papers in leading venues such as CVPR, ICCV, NeurIPS, IEEE TPAMI, and IJCV, with research focused on computer vision and deep learning. He is an Area Chair for NeurIPS and ICPR, a Senior Program Committee member for AAAI and IJCAI, and a guest editor for IEEE TBD, while also regularly reviewing for top-tier journals and conferences.
\end{IEEEbiography}

\begin{IEEEbiography}[{\includegraphics[width=1in,height=1.25in,clip,keepaspectratio]{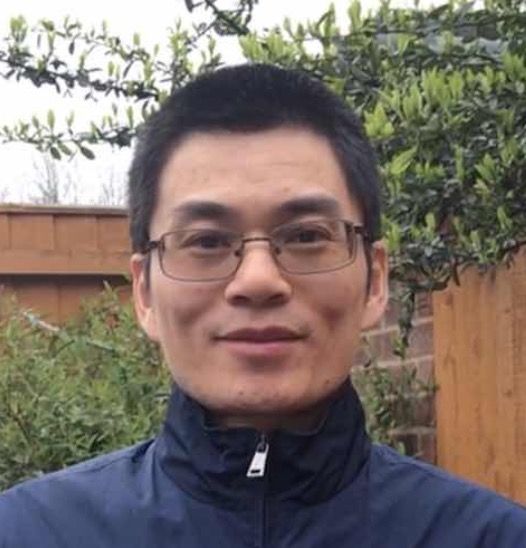}}]{Xiatian Zhu} received the PhD degree from the Queen Mary University of London. He is a senior lecturer with Surrey Institute for People-Centred Artificial Intelligence, and Centre for Vision, Speech and Signal Processing (CVSSP), University of Surrey, Guildford, U.K. He won the Sullivan Doctoral Thesis Prize 2016. He was a research scientist with Samsung AI Centre, Cambridge, U.K. His research interests include computer vision and machine learning.
\end{IEEEbiography}

\begin{IEEEbiography}[{\includegraphics[width=1in,height=1.25in,clip,keepaspectratio]{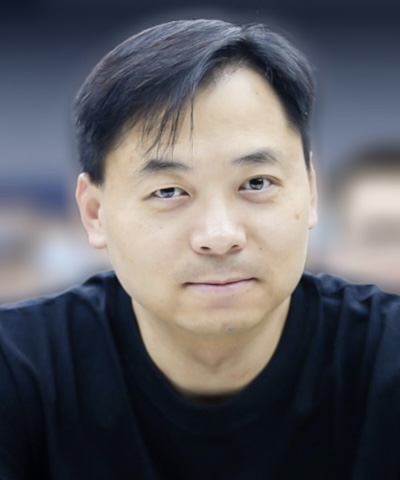}}]{Dacheng Tao} (Fellow, IEEE) is currently a Distinguished University Professor in the College of Computing \& Data Science at Nanyang Technological University. He mainly applies statistics and mathematics to artificial intelligence and data science, and his research is detailed in one monograph and over 200 publications in prestigious journals and proceedings at leading conferences, with best paper awards, best student paper awards, and test-of-time awards. His publications have been cited over 112K times and he has an h-index 160+ in Google Scholar. He received the 2015 and 2020 Australian Eureka Prize, the 2018 IEEE ICDM Research Contributions Award, and the 2021 IEEE Computer Society McCluskey Technical Achievement Award. He is a Fellow of the Australian Academy of Science, AAAS, ACM and IEEE.
\end{IEEEbiography}

\begin{IEEEbiography}[{\includegraphics[width=1in,height=1.25in,clip,keepaspectratio]{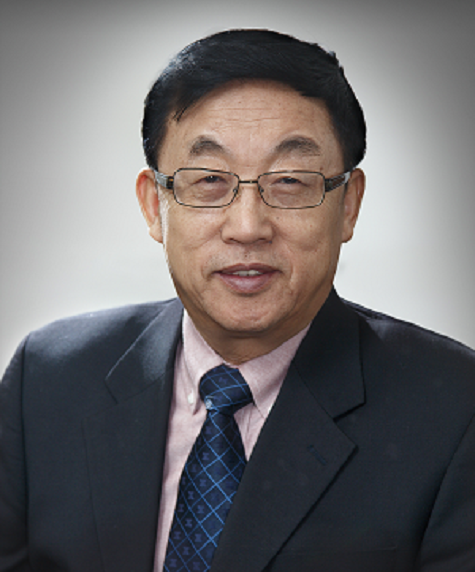}}]{Tianyou Chai}
	(Life Fellow, IEEE) received the Ph.D. degree in control theory and engineering from Northeastern University, Shenyang, China, in 1985. He became a Professor at Northeastern University in 1988. He is the Founder and the Director of the Center of Automation, Northeastern University, which became the National Engineering and Technology Research Center and the State Key Laboratory. He was the Director of the Department of Information Science, National Natural Science Foundation of China, from 2010 to 2018. He has developed control technologies with applications to various industrial processes. He has published more than 320 peer-reviewed international journal articles. His current research interests include modeling, control, optimization, and integrated automation of complex industrial processes.
	
	Dr. Chai is a member of the Chinese Academy of Engineering and a Fellow of International Federation for Automatic Control (IFAC). His paper titled “Hybrid intelligent control for optimal operation of shaft furnace roasting process” was selected as one of the three best papers for the Control Engineering Practice Paper Prize for the term 2011–2013. For his contributions, he has won five prestigious awards of the National Natural Science, the National Science and Technology Progress, and the National Technological Innovation, the 2007 Industry Award for Excellence in Transitional Control Research from IEEE Multi-Conference on Systems and Control, and the 2017 Wook Hyun Kwon Education Award from the Asian Control Association. 
\end{IEEEbiography}

\end{document}